\documentclass[journal]{IEEEtran}
\usepackage{ifpdf}

\usepackage{cite}

\ifCLASSINFOpdf
   \usepackage[pdftex]{graphicx}
\else
   \usepackage[dvips]{graphicx}
\fi
\usepackage{amsmath}
\usepackage{algorithmic}

\usepackage{array}

\ifCLASSOPTIONcompsoc
 \usepackage[caption=false,font=normalsize,labelfont=sf,textfont=sf]{subfig}
\else
 \usepackage[caption=false,font=footnotesize]{subfig}
\fi

\usepackage{stfloats}
\usepackage{url}

\usepackage{amssymb}
\usepackage{array}
\usepackage{tabularx}
\usepackage{multirow,bigstrut}
\usepackage{booktabs}
\usepackage{mathtools}
\usepackage{algorithm}
\usepackage{listings}
\usepackage{xcolor}
\usepackage{amsthm}
\usepackage{textcomp}
\usepackage{multirow}

\theoremstyle{definition}

\newtheorem{theorem}{Theorem}

\newcommand{\etal}{\textit{et al.}}
\newcommand{\ie}{\textit{i.e.}}
\newcommand{\viz}{\textit{viz.}}
\newcommand{\fig}[1]{Fig. \ref{#1}}

\newcommand{\sect}[1]{Section \ref{#1}}

\newcommand{\tab}[1]{Table \ref{#1}}
\newcommand{\alg}[1]{Algorithm~\ref{#1}}
\newcommand{\thm}[1]{Theorem~\ref{#1}}
\newcommand{\lin}[1]{Line~\ref{#1}}
\newcommand{\nCr}[2]{\,^{#1}C_{#2}} 

\DeclareMathOperator*{\argmin}{arg\,min}

\DeclareGraphicsExtensions{.pdf, .JPG}
\graphicspath{{./figures/}}

\makeatletter
\newcommand{\pushright}[1]{\ifmeasuring@#1\else\omit\hfill$\displaystyle#1$\fi\ignorespaces}
\newcommand{\pushleft}[1]{\ifmeasuring@#1\else\omit$\displaystyle#1$\hfill\fi\ignorespaces}
\makeatother

\begin{document}
%
\title{Stochastic Graphlet Embedding}
%
%
%
%

\author{Anjan Dutta,~\IEEEmembership{Member,~IEEE} and Hichem Sahbi,~\IEEEmembership{Member,~IEEE}
\thanks{Anjan Dutta is with the Computer Vision Center, Computer Science Department, Autonomous University of Barcelona, Edifici O, Campus UAB, 08193 Bellaterra, Barcelona, Spain. He was a postdoctoral researcher at the T\'{e}l\'{e}com ParisTech, Paris, France, when most of the work was done (under the MLVIS project) and part of the paper was written. (E-mail: adutta@cvc.uab.es)}
\thanks{Hichem Sahbi is with the CNRS, UPMC, Sorbonne University, Paris, France. (E-mail: hichem.sahbi@lip6.fr)}}
\maketitle


\begin{abstract}
Graph-based methods are known to be successful in many machine learning and pattern classification tasks. These methods consider semi-structured data as graphs where nodes correspond to primitives (parts, interest points, segments, etc.) and edges characterize the relationships between these primitives. However, these non-vectorial graph data cannot be straightforwardly plugged into off-the-shelf machine learning algorithms without a preliminary step of -- explicit/implicit -- graph vectorization and embedding. This embedding process should be resilient to intra-class graph variations while being highly discriminant. In this paper, we propose a novel high-order stochastic graphlet embedding (SGE) that maps graphs into vector spaces. Our main contribution includes a new stochastic search procedure that efficiently parses a given graph and extracts/samples unlimitedly high-order graphlets. We consider these graphlets, with increasing orders, to model local primitives as well as their increasingly complex interactions. In order to build our graph representation, we measure the distribution of these graphlets into a given graph, using particular hash functions that efficiently assign sampled graphlets into isomorphic sets with a very low probability of collision. When combined with maximum margin classifiers, these graphlet-based representations have positive impact on the performance of pattern comparison and recognition as corroborated through extensive experiments using standard benchmark databases.
\end{abstract}


%

\IEEEpeerreviewmaketitle

\section{Introduction}
\label{sec:intro}

In this paper, we consider the problem of graph-based classification: given a pattern (image, shape, handwritten character, document etc.) modeled with a graph, the goal is to predict the class that best describes the visual and the semantic content of that pattern, which essentially turns into a \emph{graph classification/recognition} problem. Most of the early pattern classification methods were designed using numerical feature vectors resulting from statistical analysis~\cite{Csurka2004,Ling2007a}. Other more successful extensions of these methods also integrate structural information~(see for instance~\cite{Lazebnik2006}). These extensions were built upon the assumption that parts, in patterns, do not appear independently and structural relationships among these parts are crucial in order to achieve effective description and classification~\cite{Harchaoui2007}.

Among existing pattern description and classification solutions, those based on graphs are particularly successful~\cite{Conte2004,Duchenne2011,Foggia2014}. In these methods, patterns are first modeled with graphs (where nodes correspond to local primitives and edges describe their spatial and geometric relationships), then graph matching techniques are used for recognition. This framework has been successfully applied to many pattern recognition problems~\cite{Cho2010,Sharma2011,Duchenne2011,Zhou2013,Wu2014}. This success is mainly due to the ability to encode interactions between different inter/intra class object entities and the relatively efficient design of some graph-based matching algorithms. 

The main disadvantage of graphs, compared to the usual vector-based representations, is the significant increase of complexity in graph-based algorithms. For instance, the complexity of feature vector comparison is linear (w.r.t vector dimension) while the complexity of general graph comparison is currently known to be GI-complete~\cite{Kobler1994} for graph isomorphism and NP-complete for subgraph isomorphism. Another serious limitation, in the use of graphs for pattern recognition tasks, is the incompatibility of most of the mathematical operations in graph domain. For example, computing pairwise sums or products (which are elementary operations in many classification and clustering algorithms) is not defined in a standardized way in graph domain. However, these elementary operations should be defined in a particular way in different machine learning algorithms. \textcolor{black}{Considering $\mathbb{G}$ as an arbitrary set of graphs, a possible way to address this issue is either to define an \emph{explicit embedding} function $\varphi:\mathbb{G}\rightarrow \mathbb{R}^n$ to a real vector space or to define an \textit{implicit embedding} function $\varphi:\mathbb{G}\rightarrow \mathcal{H}$ to a high dimensional Hilbert space $\mathcal{H}$ where a dot product defines similarity between two graphs $K(G,G')=\langle \varphi(G),\varphi(G') \rangle$, $G,G'\in\mathbb{G}$.} In graph domain, this implicit inner product is termed as \emph{graph kernel} that basically defines similarity between two graphs which is usually coupled with machine learning and inference techniques such as support vector machine (SVM) in order to achieve classification. Graph kernels are usually designed in two ways: (i) by approximate graph matching, \ie, by defining similarity between two graphs proportionally to the number of aligned sub-patterns, such as, nodes, edges, random walks~\cite{Gartner2003}, shortest paths~\cite{Dupe2010}, cycles~\cite{Horvath2004}, subtrees~\cite{Shervashidze2009a}, etc. or (ii) by considering similarity as a decreasing function of a distance between first or high order statistics of their common substructures, such as, graphlets~\cite{Shervashidze2009,Saund2013} or graph edit distances w.r.t a predefined set of prototype graphs~\cite{Bunke2010}. Thus, the second family of methods first defines an explicit graph embedding and then compute similarities in the embedding vector space. Nevertheless, these methods are usually memory and time demanding as sub-patterns are usually taken from large dictionaries and searched by handling the laborious subgraph isomorphism problem~\cite{Mehlhorn1984} which is again known to be NP-complete for general and unconstrained graph structures.

\begin{figure*}[!t]
\begin{center}
\fbox{\includegraphics[width=\textwidth]{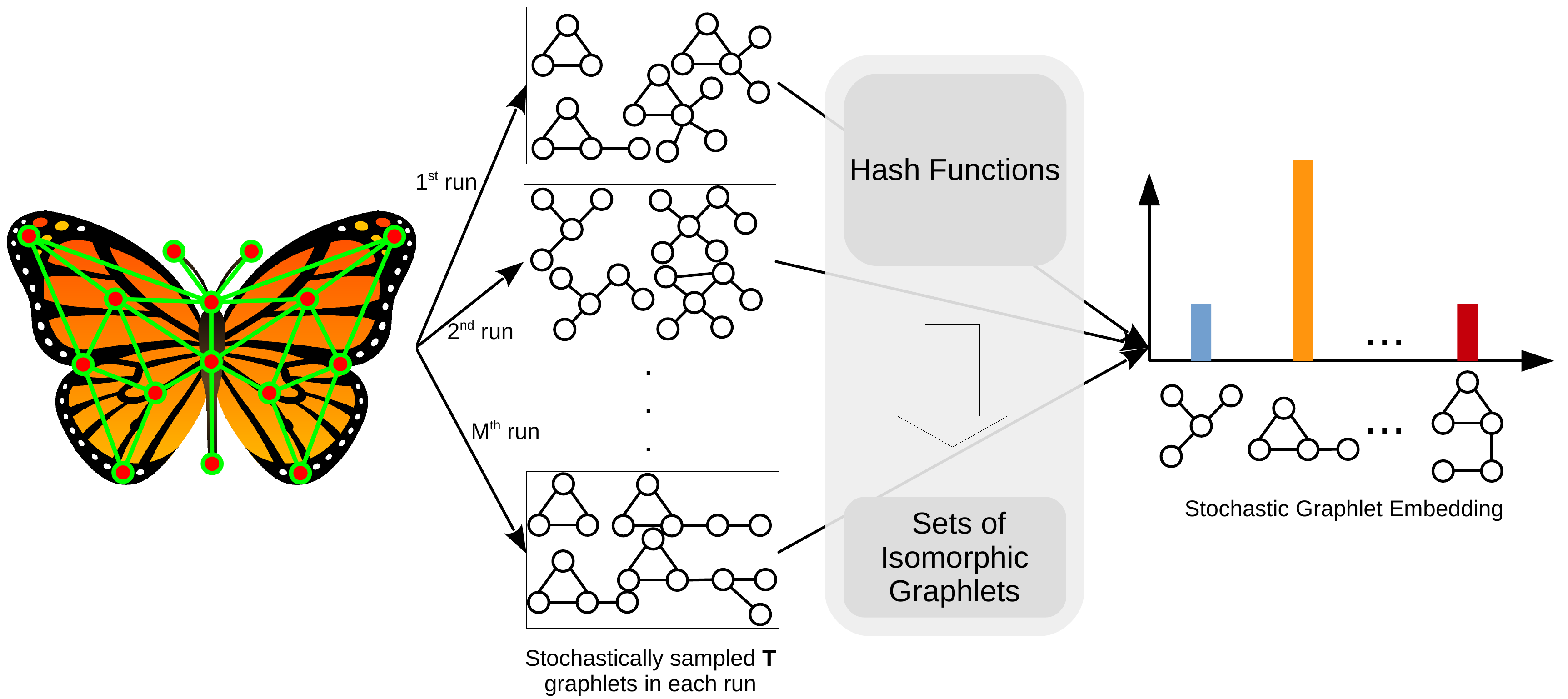}}
\caption{Overview of our stochastic graphlet embedding (SGE). Given a graph of a pattern (hand-crafted graph on the butterfly) and denoted as $G$, our stochastic search algorithm is able to sample graphlets of increasing size. Controlled by two parameters $M$ (number of graphlets to be sampled) and $T$ (maximum size of graphlets in terms of number of edges), our method extracts in total $M\times T$ graphlets. These graphlets are encoded and partitioned into isomorphic graphlets using our well designed hash functions with a low probability of collision. A distribution of different graphlets is obtained by counting the number of graphlets in each of these partitions. This procedure results in a vectorial representation of the graph $G$ referred to as stochastic graphlet embedding.}
\label{fig:schematic-diag}
\end{center}
\end{figure*}

In this paper, we propose a high-order \emph{stochastic graphlet embedding} method that models the distribution of (unlimitedly) high-order\footnote{In general, the order of a graph is defined as the total number of its vertices. In this paper, we use a dual definition of the term ``order'' to indicate the number of its edges.} connected graphlets (subgraphs) of a given graph. The proposed method gathers the advantages of the two aforementioned families of graph kernels while discarding their limitations. Indeed, our technique does not maintain predefined dictionaries of graphlets, and does not perform laborious exact search of these graphlets using subgraph isomorphism. \textcolor{black}{In contrast, the proposed algorithm samples high-order graphlets in a stochastic way, and allows us to obtain a distribution asymptotically close to the actual distribution.} Furthermore, graphlets -- as complex structures -- are much more discriminating compared to simple walks or tree patterns. Following these objectives, the whole proposed procedure is achieved by: 
\begin{itemize}
\item Significantly restricting graphlets to include only subgraphs belonging to training and test data.
\item Parsing this restricted subset of \textcolor{black}{graphlets}, using an efficient stochastic depth-first-search procedure that extracts statistically meaningful distributions of graphlets.
\item Indexing these graphlets using hash functions, with low probability of collision, that capture isomorphic relationships between graphlets quite accurately.
\end{itemize}

Our technique randomly samples high-order graphlets in a given graph, splits them into subsets and obtains the cardinality and {\it thereby} the distribution of these graphlets efficiently. This is obtained thanks to our search strategy that parses and hashes graphlets into subsets of similar and topologically isomorphic graphlets. More precisely, we employ effective graph hashing functions, such as \emph{degree of nodes} and \emph{betweenness centrality}; while it is always guaranteed that isomorphic graphlets will obtain identical hash codes with these hash functions, it is not always guaranteed that non-isomorphic graphlets will always avoid collisions (\ie, obtain different hash codes)\footnote{though this collision happens with a very low probability.}, and this is in accordance with the GI-completeness of graph-isomorphism. In summary, with this parsing strategy, we obtain resilient and efficient graph representations (compared to many related techniques including subgraph isomorphism as also shown in experiments) to the detriment of a negligible increase of the probability of collision in the obtained distributions. Put differently, the proposed procedure is very effective and can fetch the distribution of unlimited order graphlets with a controlled complexity. These graphlets, with relatively high orders, have positive and more influencing impact on the performance of pattern classification, as supported through extensive experiments which also show that our proposed method is highly effective for structurally informative graphs with possibly attributed nodes and edges. Considering these issues, the main contributions of our work include:
\begin{enumerate}
\item A new stochastic depth-first-search strategy that parses any given graph in order to extract increasingly complex graphlets with a large bound on the number of their edges. 
\item Efficient and also effective hash functions, that index and partition graphlets into isomorphic sets with a low probability of collision. 
\item Last but not least, a comprehensive experimental setting that shows the resilience of our graph representation method against intra-class graph variations and its efficiency as well as its comparison against related methods. 
\end{enumerate}

\fig{fig:schematic-diag} illustrates the key idea and the flowchart of our proposed stochastic graphlet embedding algorithm; as shown in this example, we consider the butterfly image as a pattern endowed with a hand-crafted input graph. We sample $M\times T$ connected graphlets of increasing orders with the proposed stochastic depth-first-search procedure (in \sect{sec:graphlets}). We also consider well-crafted graph hash functions with low probability of collision (in \sect{sec:hash-fns}). After sampling the graphlets, we partition them into disjoint isomorphic subsets using these hash functions. The cardinality of each subsets allows us to estimate the empirical distribution of isomorphic graphlets present in the input graph. This distribution is referred to as \emph{stochastic graphlet embedding} (SGE).

At the best of our knowledge, no existing work in pattern analysis has achieved this particularly effective, efficient and resilient graph embedding scheme, \ie, being able to extract graphlet patterns using a stochastic search procedure and assign them to topologically isomorphic sets of similar graphlets using efficient and accurate hash functions with a low probability of collision. In this context, the two most closely related works were proposed by Shervashidze~\etal~\cite{Shervashidze2009} and Saund~\cite{Saund2013}. In Shervashidze~\etal~\cite{Shervashidze2009}, authors consider a fixed dictionary of subgraphs (with a bound on their degree set to $5$). They provide two schemes in order to enumerate graphlets; one based on sampling and the other one specifically designed for bounded degree graphs. Compared to this work, the enumeration of larger graphlets in our method carries out more relevant information, which has been revealed in our experiment.\\ 
\noindent In Saund~\cite{Saund2013}, authors provide a set of primitive nodes, create a graph lattice in a bottom-up way, which is used to enumerate the subgraphs while parsing a given graph. However, the way of considering limited number of primitives has made their method application specific. In addition, increment of the average degrees of node in a dataset would result in a very big graph lattice, which will increase the time complexity when parsing graphs. In contrast, our proposed method in this paper does not require a fixed vocabulary of graphlets. The candidate graphlets to be considered for enumeration are entirely determined by training and test data. Furthermore, our method is not dependent on any specific application and is versatile. This fact has been proven by experiments on different type of datasets, \viz, protein structures, chemical compound, form documents, graph representation of digits, shape, etc.

The rest of this paper is organized as follows: \sect{sec:relwrks} reviews the related work on graph-based kernels and explicit graph embedding methods. \sect{sec:graphlets} introduces our efficient stochastic graphlet parsing algorithm, and \sect{sec:hash-fns} describes hashing techniques in order to build our stochastic graphlet embedding. \sect{sec:comp-anal} discusses the computational complexity of our proposed method and~\sect{sec:results} presents a detailed experimental validation of the proposed method showing the positive impact of high-order graphlets on the performance of graph classification. Finally, \sect{sec:concl} concludes the paper while briefly providing possible extensions for a future work. 

\section{Related Work}
\label{sec:relwrks}

In what follows, we review the related work on explicit and implicit graph embedding. The former seeks to generate explicit vector representations suitable for learning and classification while the latter endows graphs with inner products involving maps in high dimensional Hilbert spaces; these maps are implicitly obtained using graph kernels. 
 
\subsection{Graph Kernel Embedding}

Kernel methods have been popular during the last two decades mainly because of their ability to extend, in a unified manner, the existing machine learning algorithms to non-linear data. The basic idea, known as the kernel trick~\cite{Vapnik1998}, consists in using positive semi-definite kernels in order to implicitly map non-linearly separable data from an original space to a high dimensional Hilbert space without knowing these maps explicitly; only kernels are known. Another major strength of kernel methods resides in their ability to handle non-vectorial data (such as graphs, string or trees) by designing appropriate kernels on these data while still using off-the-shelf learning algorithms.

\subsubsection{Diffusion Kernels}
Given a collection of graphs $\mathbb{G} = \lbrace G_1, G_2, \dots, G_N\rbrace$, a decay factor $0 < \lambda < 1$, and a similarity function $s: \mathbb{G}\times\mathbb{G} \rightarrow \mathbb{R}$, a diffusion kernel \cite{Lafferty2005} is defined as 
\begin{equation*}
\mathbf{K} = \sum_{k=0}^{\infty} \frac{1}{k!}\lambda^k\mathbf{S}^k = \exp(\lambda\mathbf{S}),
\end{equation*}
here $\mathbf{S}=(s_{ij})_{N\times N}$ is a matrix of pairwise similarities; when $\mathbf{S}$ is symmetric, $\mathbf{K}$ becomes positive definite \cite{Smola2003}. An alternative, known as the \emph{von Neumann diffusion kernel}~\cite{Kandola2002}, is also defined as $\mathbf{K} = \sum_{k=0}^{\infty} \lambda^k\mathbf{S}^k$. In these diffusion kernels, the decay factor $\lambda$ should be sufficiently small in order to ensure that the weighting factor $\lambda^k$ will be negligible for sufficiently large $k$. Therefore, only a finite number of addends are evaluated in practice. 

\subsubsection{Convolution Kernels}
The general principle of convolution kernels consists in measuring the similarity of composite patterns (modeled with graphs) using the similarity of their parts (\ie~nodes)~\cite{Watkins1999}. Prior to define a convolution kernel on any two given graphs $G,G' \in \mathbb{G}$, one should consider elementary functions $\{\kappa_\ell\}_{\ell=1}^d$ that measure the pairwise similarities between nodes $\lbrace v_i\rbrace_{i}$, $\lbrace v_j'\rbrace_{j}$ in $G$, $G'$ respectively. Hence, the convolution kernel can be written as~\cite{Neuhaus2007}:
\begin{equation*}
\kappa(G,G')=\sum_{i}\sum_{j}\prod_{\ell=1}^d \kappa_\ell(v_i,v_j').
\end{equation*}
This graph kernel derives the similarity between two graphs $G$, $G'$ from the sum, over all decompositions, of the similarity products of the parts of $G$ and $G'$~\cite{Neuhaus2007}. Recently, Kondor and Pan~\cite{Kondor2016} proposed multi-scale Laplacian graph kernel having the property of lifting a base kernel defined on the vertices of two graphs to a kernel between graphs.

\subsubsection{Substructure Kernels}

A third class of graph kernels is based on the analysis of common substructures, including random walks~\cite{Vishwanathan2010}, backtrackless walks~\cite{Aziz2013}, shortest paths~\cite{Borgwardt2005}, subtrees~\cite{Shervashidze2009a}, graphlets~\cite{Shervashidze2009}, edit distance graphlets~\cite{Lugo-Martinez2014}, etc. These kernels measure the similarity of two graphs by counting the frequency of their substructures that have all (or some of) the labels in common~\cite{Borgwardt2005}. Among the above mentioned graph kernels, the random walk kernel has received a lot of attention~\cite{Gartner2003,Vishwanathan2010}; in~\cite{Gartner2003}, G\"{a}rtner~\etal~showed that the number of matching walks in two graphs $G$ and $G'$ can be computed by means of the direct product graph, without explicitly enumerating the walks and matching them. This makes it possible to consider random walks of unlimited length.
 
\subsection{Explicit Graph Embedding}
Explicit graph embedding is another family of representation techniques that aims to map graphs to vector spaces prior to apply usual kernels (on top of these graph representations) and off-the-shelf learning algorithms. In this family of graph representation techniques, three different classes of methods exist in the literature; the first one, known as \emph{graph probing}~\cite{Luqman2013}, seeks to measure the frequency of specific substructures (that capture content and topology) into graphs. For instance, the method in~\cite{Shervashidze2009a} estimates the number of non-isomorphic graphlets while the approach in Gibert~\etal~\cite{Gibert2012} is based on node label and edge relation statistics. Authors in Luqman~\etal~\cite{Luqman2013} consider graph information at different topological levels (structures and attributes) while authors in~\cite{Saund2013} introduce a bottom-up graph lattice in order to estimate the distribution of graphlets into document graphs; this distribution is afterwards used as an index for document retrieval.

The second class of graph embedding methods is based on \emph{spectral graph theory}~\cite{Caelli2004,Wilson2005,RoblesKelly2007,Jouili2010}. The latter aims to analyze the structural properties of graphs using eigenvectors/eigenvalues of adjacency or Laplacian matrices~\cite{Wilson2005}. In spite of their relative success in graph representation and embedding, spectral methods are not fully able to handle noisy graphs. Indeed, this limitation stems from the fact that eigen-decompositions are sensitive to structural errors such as missing nodes/edges and short cuts. Moreover, spectral methods are applicable to unlabeled graphs or labeled graphs with small alphabets, although recent extensions tried to overcome this limitation~\cite{Lee2009}. \\
\indent The third class of methods is inspired by \emph{dissimilarity representations} proposed in~\cite{Pekalska2005}; in this context, Bunke and Riesen present the vectorial description of a given graph by its distances to a number of pre-selected prototype graphs~\cite{Riesen2007a,Riesen2009a,Bunke2010,Borzeshi2013}. Finally, and besides these three categories of explicit graph embedding, Mousavi~\etal~\cite{Mousavi2017} recently proposed a generic framework based on graph pyramids which hierarchically embeds any given graph to a vector space (that models both local and global graph information).
 
\section{High Order Stochastic Graphlets}
\label{sec:graphlets}

Our main goal is to design a novel explicit graph embedding technique that combines the representational power and the robustness of high-order graphlets as well as the efficiency of graph hashing. As shown subsequently, patterns represented with graphs are described with distributions of high-order graphlets, where the latter are extracted using an efficient stochastic depth-first-search strategy and partitioned into isomorphic sets of graphlets using well defined hashing functions.

\subsection{Graphs and Graphlets}
\label{ssec:graph-based-image-reprn}
Let us consider a finite collection of $m$ patterns $\mathcal{S} = \lbrace\mathcal{P}_1,...,\mathcal{P}_m\rbrace$. A given pattern $\mathcal{P}\in\mathcal{S}$ is described with an \emph{attributed graph} which is basically a $4$-tuple $G=(V,E,\phi,\psi)$; here $V$ is a node set and $E\subseteq V\times V$ is an edge set. The two mappings $\phi:V\rightarrow\mathbb{R}^m$ and $\psi:E\rightarrow\mathbb{R}^n$ respectively assign attributes to nodes and edges of $G$. An attributed graph $G'=(V',E',\phi',\psi')$ is a \emph{subgraph} of $G$ (denoted by $G'\subseteq G$) if the following conditions are satisfied:
\begin{itemize}
\item $V'\subseteq V$
\item $E'=E\cap V'\times V'$
\item $\phi'(u)=\phi(u), \forall u\in V'$
\item $\psi'(e)=\psi(e), \forall e\in E'$
\end{itemize}

A graphlet refers to any subgraph $g$ of $G$ that may also inherit the topological and the attribute properties of $G$; in this paper, we only consider ``connected graphlets'' and, for short, we omit the terminology ``connected'' when referring to graphlets. We use these graphlets to characterize the distribution of local pattern parts as well as their spatial relationships. As will be shown, and in contrast to the mainstream work, our method neither requires a preliminary tedious step of specifying large dictionaries of graphlets {\it nor} checking for the existence of these large dictionaries (in the input graphs) using subgraph isomorphism which is again intractable.

\begin{algorithm}
\caption{\textsc{Stochastic-Graphlet-Parsing}($G$): Create a set of graphlets $\mathbb{S}$ by traversing $G$.}
\label{alg:extractgrahletsdfs}
\begin{algorithmic}[1]
\REQUIRE $G=(V,E)$, $\mathit{M}$, $\mathit{T}$
\ENSURE $\mathbb{S}$
\STATE $\mathbb{S}\leftarrow \emptyset$
\FOR {$i=1$ to $\mathit{M}$} \label{alg:extractgrahletsdfs:it1}
\STATE $u \leftarrow \textsc{SelectRandomNode}(V)$ \label{alg:extractgrahletsdfs:ransel1}
\STATE $\mathit{U_0}\leftarrow u$, $\mathit{A_0} \gets \emptyset$ \label{alg:extractgrahletsdfs:init}
\FOR {$t=1$ to $\mathit{T}$} \label{alg:extractgrahletsdfs:it2}
\STATE $u \leftarrow \textsc{SelectRandomNode}(\mathit{U_{t-1}})$ \label{alg:extractgrahletsdfs:ransel2}
\STATE $v \leftarrow \textsc{SelectRandomNode}(V): (u,v)\in E\setminus\mathit{A_{t-1}}$ \label{alg:extractgrahletsdfs:ransel3}
\STATE $U_t \leftarrow U_{t-1} \cup \{v\}$ , \ \ \ $A_t \leftarrow A_{t-1} \cup \{(u,v)\}$ \label{alg:extractgrahletsdfs:updateVN}
\STATE $\mathbb{S} \leftarrow \mathbb{S} \cup\lbrace( U_t, A_t)\rbrace$
\ENDFOR
\ENDFOR
\end{algorithmic}
\end{algorithm}

\subsection{Stochastic Graphlet Parsing}
\label{ssec:higher-order-spatial-feat}

Considering an input graph $G=(V,E,\phi,\psi)$ corresponding to a pattern $\mathcal{P}\in\mathcal{S}$, our goal is to obtain the distribution of graphlets in $G$, without considering a predefined dictionary and without explicitly tackling the subgraph isomorphism problem. The way we acquire graphlets is stochastic and we {consider both the low and high-order graphlets} without constraining their topological or structural properties (max degree, max number of nodes, etc.).

Our graphlet extraction procedure is based on a random walk process that efficiently parses and extracts subgraphs from $G$ with increasing complexities measured by the number of edges. This graphlet extraction process, outlined in \alg{alg:extractgrahletsdfs}, is iterative and regulated by two parameters $M$ and $T$, where $M$ denotes the number of runs (related to the number of distinct connected graphlets to extract) and $T$ refers to a bound on the number of edges in graphlets. In practice, $M$ is set to relatively large values in order to make graphlet generation statistically meaningful (see \lin{alg:extractgrahletsdfs:it1}). Our stochastic graphlet parsing algorithm iteratively visits the connected nodes and edges in $G$ and extracts (samples) different graphlets with an increasing number of edges denoted as $t \leq T$ (see~\lin{alg:extractgrahletsdfs:it2}), following a $T$-step random walk process with restart. Considering $U_t$, $A_t$ respectively as the aggregated sets of visited nodes and edges till step $t$, we initialize, $A_0=\emptyset$ and $U_0$ with a randomly selected node $u$ which is uniformly sampled from $V$ (see~\lin{alg:extractgrahletsdfs:ransel1} and~\lin{alg:extractgrahletsdfs:init}). For $t \geq 1$, the process continues by sampling a subsequent node $v \in V$, according to the following distribution 
\begin{equation*}
P_{t}(v|u) = \alpha \ P_{t,w}(v|u) + (1-\alpha) \ P_{t,r}(v), 
\end{equation*}
here $P_{t,w}(v|u)$ corresponds to the conditional probability of a random walk from node $u$ to its neighbor $v$ {set to uniform (if graphs are label/attribute-free) or set proportional to the label/attribute similarity between nodes $u$, $v$ otherwise}, and $P_{t,r}(v)$ is the probability to restart the random walk from an already visited node $v \in U_{t-1}$, defined as $P_{t,r}(v) = 1_{\{v \in U_{t-1} \}} \ . \ \frac{1}{|U_{t-1}|}$, with $1_{\{\}}$ being the indicator function. In the definition of $P_{t}(v|u)$, the coefficient $\alpha \in [0,1]$ controls the trade-off between random walks and restarts, and it is set to $\frac{1}{2}$ in practice. This choice of $\alpha$ provides an equilibrium between two processes (either ``continue the random walk'' from the last visited node or ``restart this random walk'' from another node); when $\alpha \gg \frac{1}{2}$ the algorithm gives preference to "continue" and this may statistically bias the sampling by giving preference to ``chain-like'' graphlet structures (that favor the increase of their depth/diameter) while $\alpha \ll \frac{1}{2}$ results into ``tree-like'' graphlet structures (that favor the increase of their width). Considering this model, graphlet sampling is achieved following two steps:
\begin{itemize}
\item Random walks: in order to expand a currently generated graphlet with a neighbor $v$ of the (last) node $u$ visited in that graphlet which possibly has similar visual features/attributes.
\item Restarts: in order to continue the expansion of the currently generated graphlet using other nodes if the set of edges connected to $u$ is fully exhausted.
\end{itemize}
Finally, if $(u,v) \in E$ and $(u,v) \notin A_{t-1}$, then the aggregated sets of nodes and edges at step $t$ are updated as:
 \begin{equation*}
U_t \leftarrow U_{t-1} \cup \{v\}
\end{equation*}
\begin{equation*}
 A_t \leftarrow A_{t-1} \cup \{(u,v)\},
\end{equation*}
\noindent which is also shown in \lin{alg:extractgrahletsdfs:updateVN} of Algorithm~1.\\
\noindent This algorithm iterates $M$ times and, at each iteration, it generates $T$ graphlets including $1,\ldots,T$ edges; in total, it generates $M\times T$ graphlets. Note that \alg{alg:extractgrahletsdfs} is already efficient on single CPU configurations -- and also highly parallelizable on multiple CPUs -- so it is suitable to parse and extract huge collections of graphlets from graphs.

This proposed graphlet parsing algorithm, by its design, allows us to uniformly sample subgraphs (graphlets) from a given graph $G$ and assign them to isomorphic sets in order to measure the distribution of graphlets into $G$. By the law of large numbers, this sampling guarantees that the empirical distribution of graphlets is asymptotically close to the actual distribution. In the non-asymptotic regime (\ie,~$M \ll \infty$), the actual number of samples needed to achieve a given confidence with a small probability of error is called the \emph{sample complexity} (see for instance the related work in bioinformatics~\cite{Przulj2007}, \cite{Shervashidze2009} and also Weissman~\etal~\cite{Weissman2003} who provide a distribution dependent bound on sample complexity, for the $L_1$ deviation, between the true and the empirical distributions). Similarly to ~\cite{Shervashidze2009}, we adapt a strong sample complexity bound $M$ as shown subsequently.
\begin{theorem}
Let $D$ be a probability distribution on a finite set of cardinality $\mathit{a}$ and let $\lbrace X_j\rbrace_{j=1}^M$ be $M$ samples identically distributed from $D$. For a given error $\epsilon>0$ and confidence $(1-\delta) \in [0,1]$, 
\begin{equation*}
M=\Bigg\lceil \frac{2\Big(\mathit{a}\ln2+\ln(\frac{1}{\delta})\Big)}{\epsilon^2}\Bigg\rceil
\end{equation*}
samples suffice to ensure that $P\Big\lbrace ||D-\hat{D}^M||_1\leq\epsilon\Big\rbrace\geq1-\delta$, with $\hat{D}^M$ being the empirical estimate of $D$ from the $M$ samples $\lbrace X_j\rbrace_{j=1}^M$. 
\label{thm:sample-compl}
\end{theorem}
 
\begin{table}[!htbp]
\begin{center}
\caption{Sample complexity bounds according to~\thm{thm:sample-compl} for graphlets with orders ranging from $1$ to $10$ and for different settings of $\epsilon$ and $\delta$.}
\label{tab:sample-no}
\resizebox{\columnwidth}{!}{
\begin{tabular}{|c|c|c|c|c|c|}
\hline
Orders & Number & $M$ & $M$ & $M$ & $M$ \\
of & of possible & ($\epsilon=0.1,$ & ($\epsilon=0.1,$ & ($\epsilon=0.05,$ & ($\epsilon=0.05,$\\
graphs & graphs ($a$) & $\delta=0.1$) & $\delta=0.05$) & $\delta=0.1$) & $\delta=0.05$) \\\hline
$1$ & $1$ & $600$ & $738$ & $2397$ & $2952$\\
$2$ & $1$ & $600$ & $738$ & $2397$ & $2952$\\
$3$ & $3$ & $877$ & $1016$ & $3506$ & $4061$\\
$4$ & $5$ & $1154$ & $1293$ & $4615$ & $5170$\\
$5$ & $12$ & $2125$ & $2263$ & $8497$ & $9051$\\
$6$ & $30$ & $4620$ & $4759$ & $18478$ & $19033$\\
$7$ & $79$ & $11413$ & $11551$ & $45649$ & $46204$\\
$8$ & $227$ & $31930$ & $32069$ & $127718$ & $128273$\\
$9$ & $710$ & $98888$ & $99027$ & $395550$ & $396105$\\
$10$ & $2322$ & $322359$ & $322497$ & $1289433$ & $1289987$\\
\hline
\end{tabular}}
\end{center}
\end{table}
\noindent The proof of the above theorem is out of the main scope of this paper and related background can be found in~\cite{Weissman2003,Shervashidze2009}. In order to highlight the benefit of this theorem, we show in \tab{tab:sample-no} different estimates of $M$ w.r.t $\delta$, $\epsilon$ and increasing graph orders. For instance, with $4$ edges, only $5$ categories of non-isomorphic graphlets\footnote{Refer to the article A002905 (\url{http://oeis.org/A002905}) of OEIS (Online Encyclopedia of Integer Sequence) to know more about the number of graphs with a specific number of edges.} exist in a given graph $G$; for this setting, when $\epsilon=0.1$ and $\delta=0.1$, the overestimated value of $M$ is set to $1154$. For $(\epsilon=0.1,\delta=0.05)$, $(\epsilon=0.05,\delta=0.1)$ and $(\epsilon=0.05,\delta=0.05)$, $M$ is set to $1293$, $4615$ and $5170$ respectively.

\section{Graphlet Hashing}
\label{sec:hash-fns}

In order to obtain the distribution of sampled graphlets in a given graph $G$, one may consider subgraph isomorphism (which is again NP-complete for general graphs~\cite{Mehlhorn1984}) or alternatively partition the set of sampled graphlets into isomorphic subsets using graph isomorphism; yet, this is also computationally intractable\footnote{We tested such isomorphism-based graphlet partitioning strategy and compared it against our hashing-based partitioning and we found that the latter is at least 2 orders of magnitude faster (see~\tab{speedups}).} and known to be GI-complete~\cite{Kobler1994}, so no polynomial solution is known for general graphs. In what follows, we approach the problem differently using graph hashing. The latter generates compact and also effective hash codes for graphlets based on their local as well as holistic topological characteristics and allows one to group generated isomorphic graphlets while colliding non-isomorphic ones with a very low probability.

\begin{table*}[!t]
\begin{center}
\caption{Probability of collision $E(f)$ of different hash functions \viz~\emph{betweenness centrality}, \emph{core numbers}, \emph{degree of nodes} and \emph{clustering coefficients}. These values are enumerated on graphlets with number of edges $t=1,\ldots,10$; some examples of these graphlets are shown in Fig~\ref{fig:example1-graphlets}.}
\label{tab:hash-fn}
\resizebox{\textwidth}{!}{
\begin{tabular}{|c|c|c||c|c||c|c||c|c||c|c||}
\hline
 & & & \multicolumn{2}{c||}{betweenness centrality} & \multicolumn{2}{c||}{core numbers} & \multicolumn{2}{c||}{degree} & \multicolumn{2}{c||}{clustering coefficients} \\ \cline{4-11}
Order & Number & Number of compar- & Number of & Probability & Number of & Probability & Number of & Probability & Number of & Probability \\
of & of possible & isons for checking & collision & of & collision & of & collision & of & collision & of \\
graphlets ($t$) & graphlets ($a$) & collisions ($\nCr{a}{2}$) & occurs & collision & occurs & collision & occurs & collision & occurs & collision \\ \hline
$1$ & $1$ & $-$ & $0$ & $0.00000$ & $0$ & $0.0000$ & $0$ & $0.0000$ & $0$ & $0.0000$ \\
$2$ & $1$ & $-$ & $0$ & $0.00000$ & $0$ & $0.0000$ & $0$ & $0.0000$ & $0$ & $0.0000$ \\
$3$ & $3$ & $3$ & $0$ & $0.00000$ & $1$ & $0.3333$ & $0$ & $0.0000$ & $1$ & $0.3333$ \\
$4$ & $5$ & $10$ & $0$ & $0.00000$ & $2$ & $0.2000$ & $0$ & $0.0000$ & $3$ & $0.3000$\\
$5$ & $12$ & $66$ & $0$ & $0.00000$ & $7$ & $0.1061$ & $2$ & $0.0303$ & $7$ & $0.1061$ \\
$6$ & $30$ & $435$ & $0$ & $0.00000$ & $22$ & $0.0506$ & $11$ & $0.0253$ & $18$ & $0.0414$ \\
$7$ & $79$ & $3081$ & $1$ & $0.00032$ & $68$ & $0.0221$ & $44$ & $0.0143$ & $50$ & $0.0162$ \\
$8$ & $227$ & $25651$ & $5$ & $0.00019$ & $211$ & $0.0082$ & $167$ & $0.0065$ & $157$ & $0.0061$ \\
$9$ & $710$ & $251695$ & $27$ & $0.00011$ & $687$ & $0.0027$ & $604$ & $0.0024$ & $537$ & $0.0021$ \\
$10$ & $2322$ & $2694681$ & $108$ & $0.00004$ & $2290$ & $0.0008$ & $2145$ & $0.0008$ & $1907$ & $0.0007$ \\
\hline
\end{tabular}}
\end{center}
\end{table*}

\begin{figure*}[!htbp]
\centering
\resizebox{\textwidth}{!}{
\begin{tabular}{cccccccccc}
\subfloat{\includegraphics[width=0.1\textwidth]{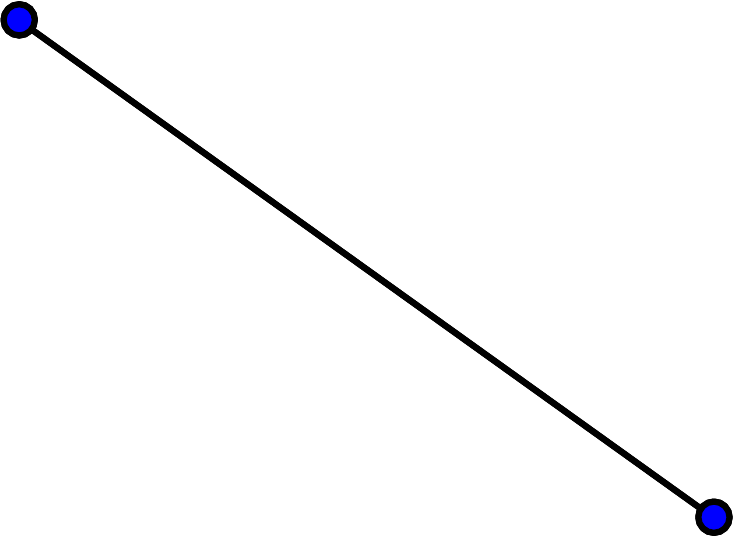}}&
\subfloat{\includegraphics[width=0.1\textwidth]{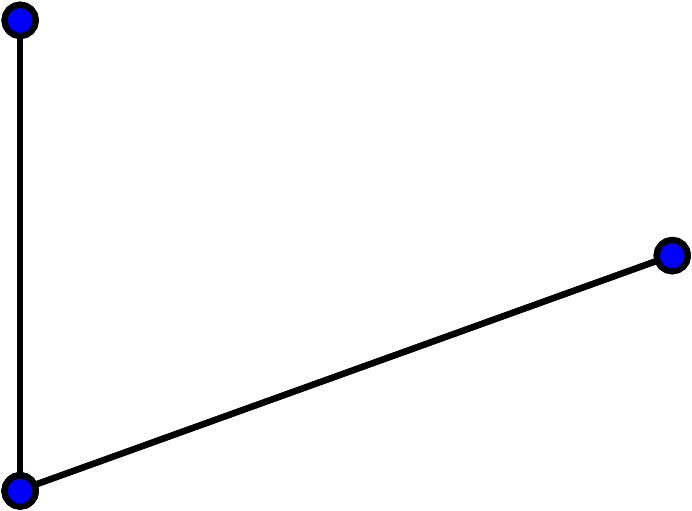}}&
\subfloat{\includegraphics[width=0.1\textwidth]{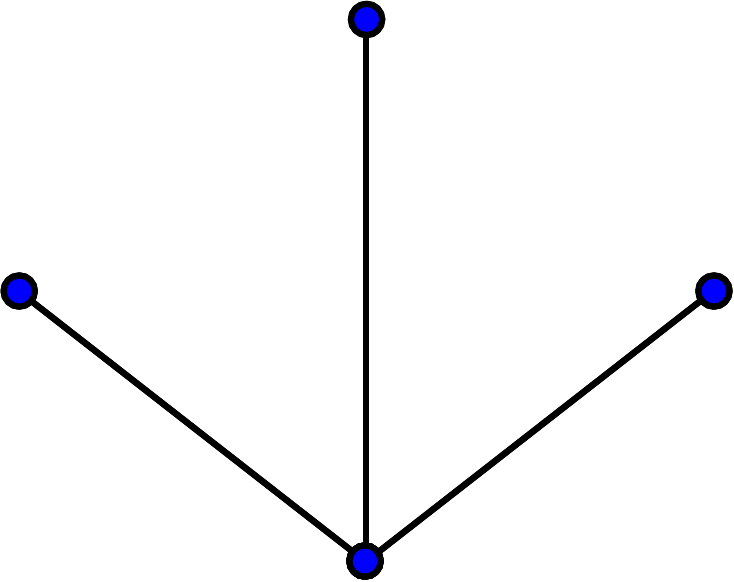}}&
\subfloat{\includegraphics[width=0.1\textwidth]{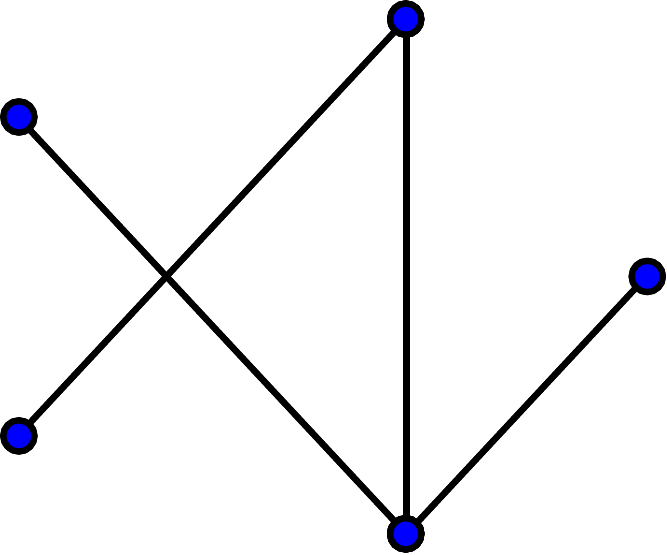}}&
\subfloat{\includegraphics[width=0.1\textwidth]{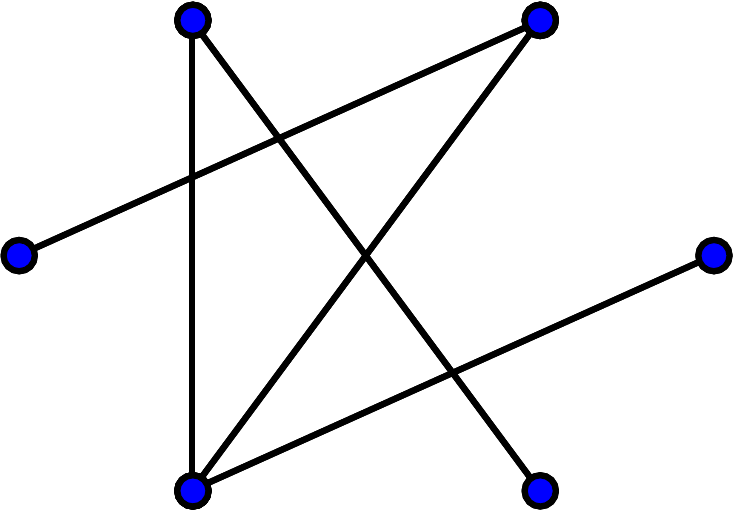}}&
\subfloat{\includegraphics[width=0.1\textwidth]{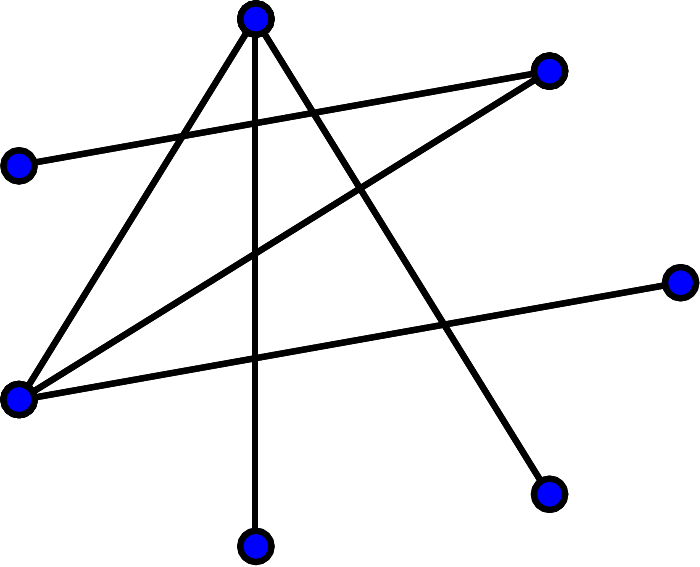}}&
\subfloat{\includegraphics[width=0.1\textwidth]{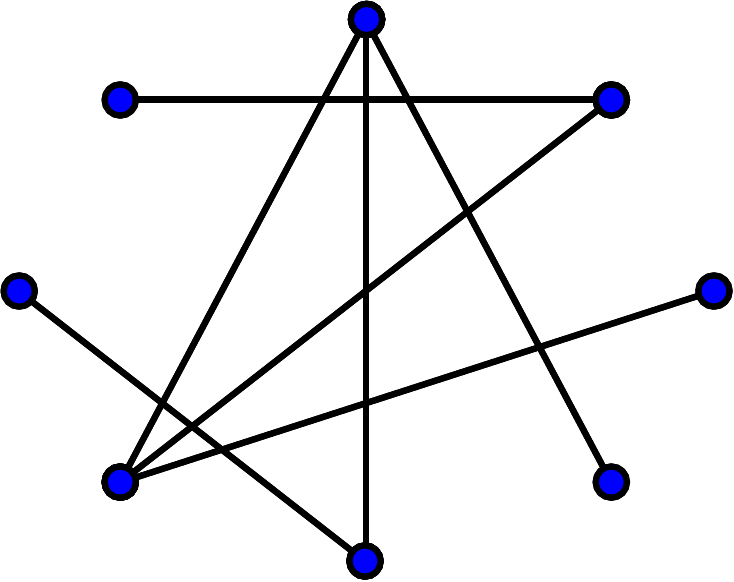}}&
\subfloat{\includegraphics[width=0.1\textwidth]{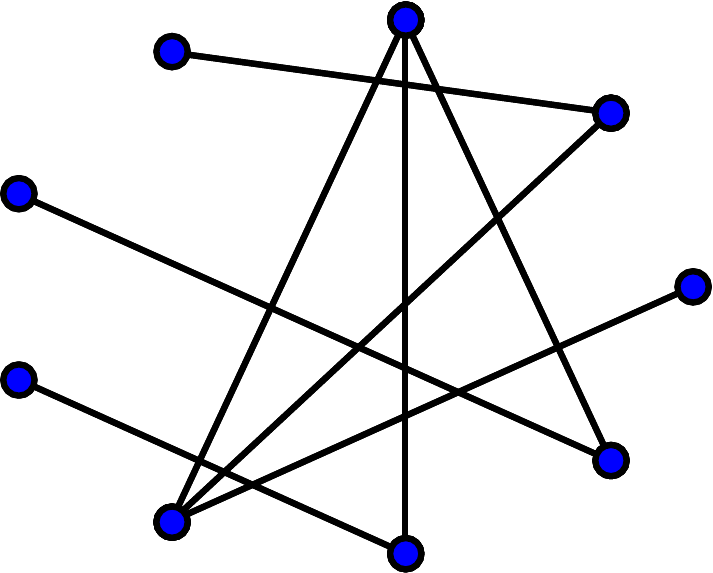}}&
\subfloat{\includegraphics[width=0.1\textwidth]{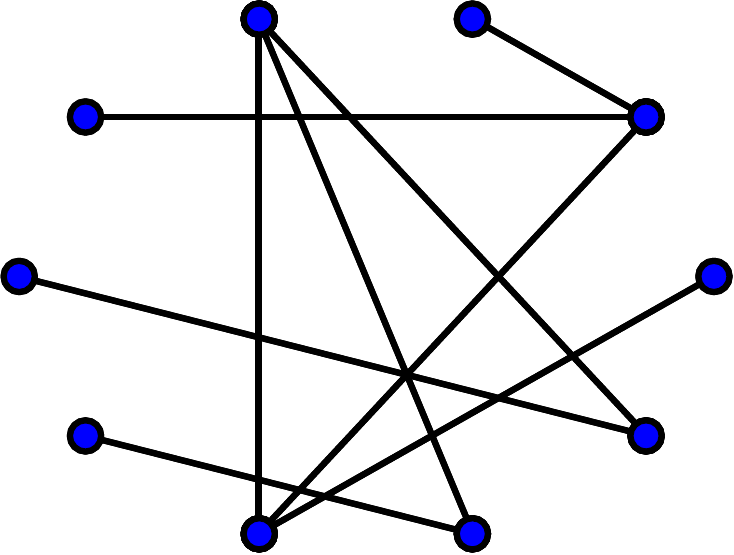}}&
\subfloat{\includegraphics[width=0.1\textwidth]{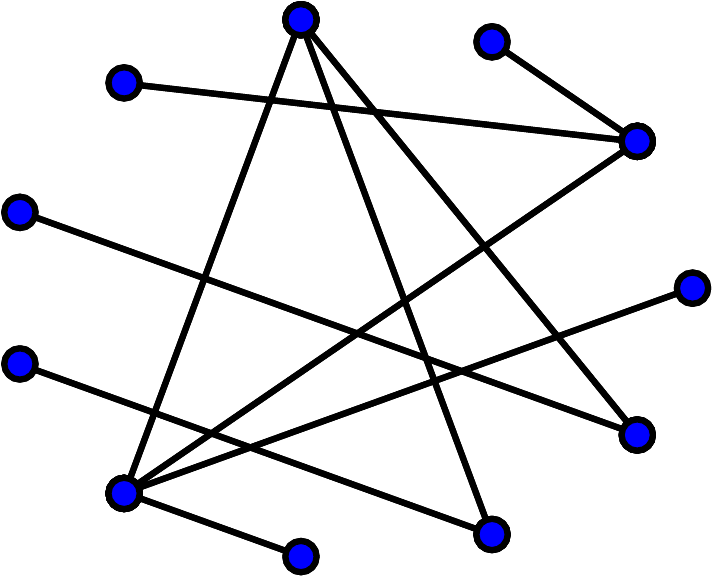}}\\
\subfloat{\includegraphics[width=0.1\textwidth]{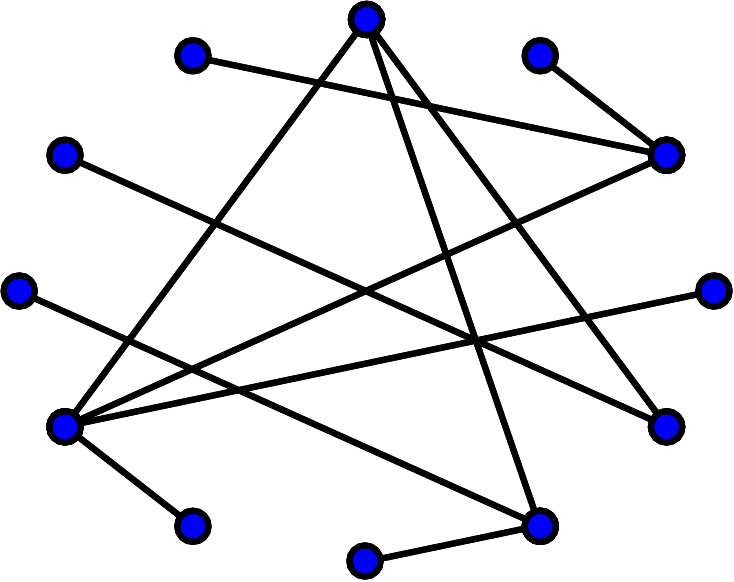}}&
\subfloat{\includegraphics[width=0.1\textwidth]{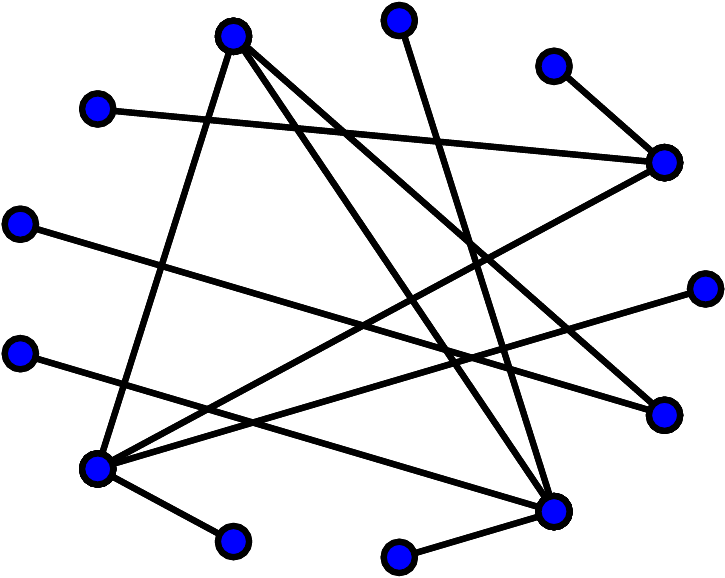}}&
\subfloat{\includegraphics[width=0.1\textwidth]{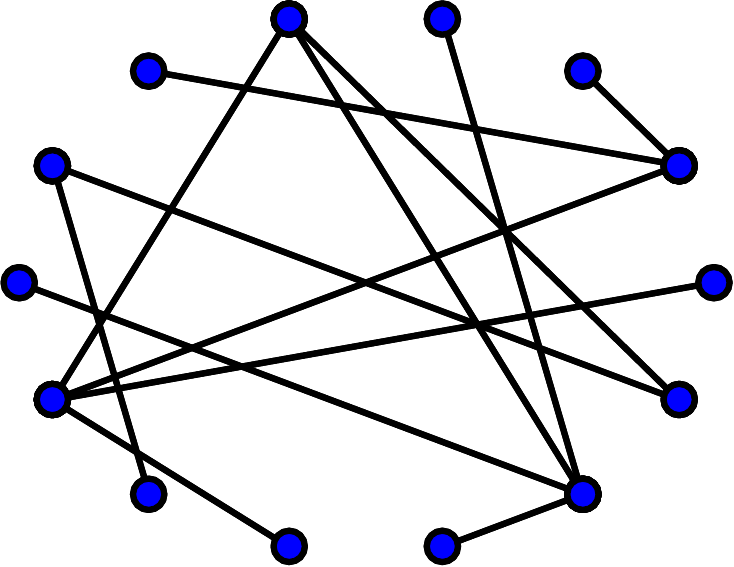}}&
\subfloat{\includegraphics[width=0.1\textwidth]{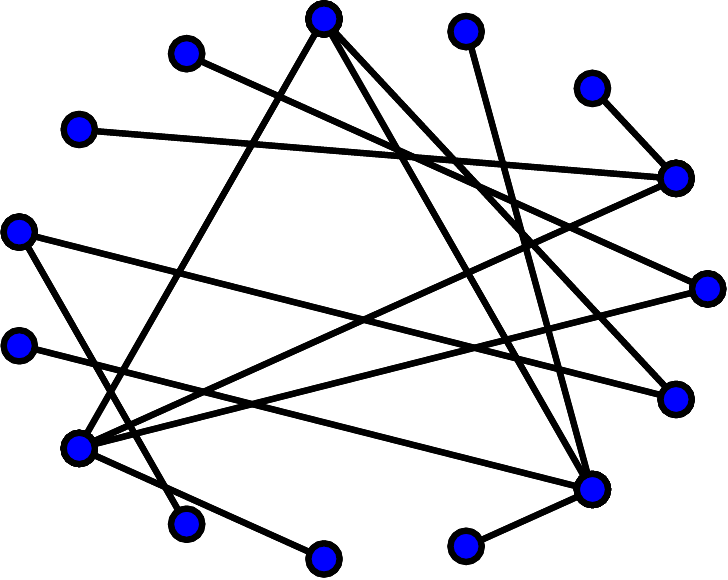}}&
\subfloat{\includegraphics[width=0.1\textwidth]{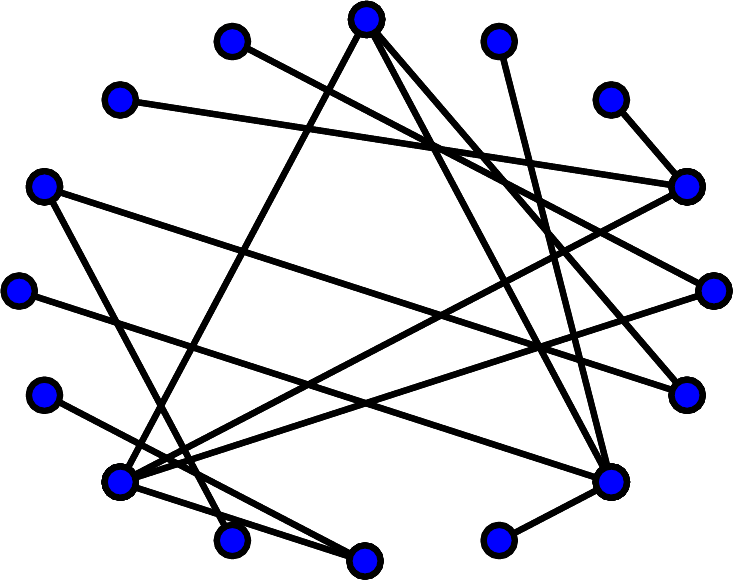}}&
\subfloat{\includegraphics[width=0.1\textwidth]{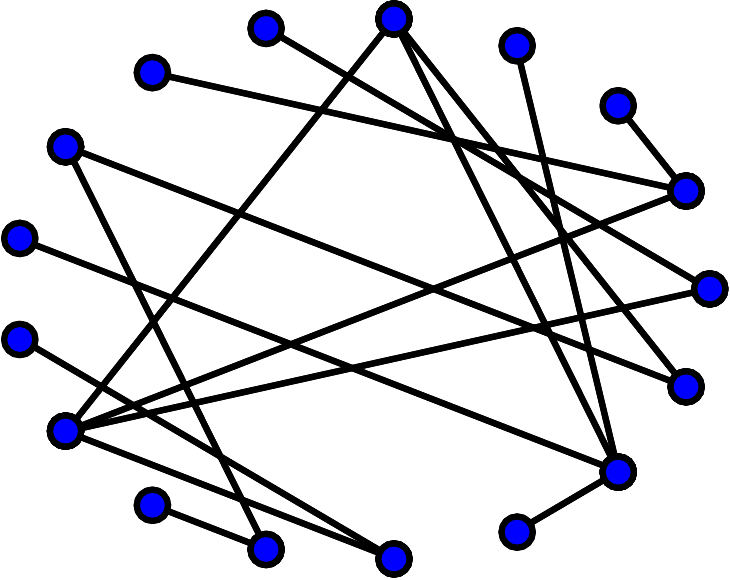}}&
\subfloat{\includegraphics[width=0.1\textwidth]{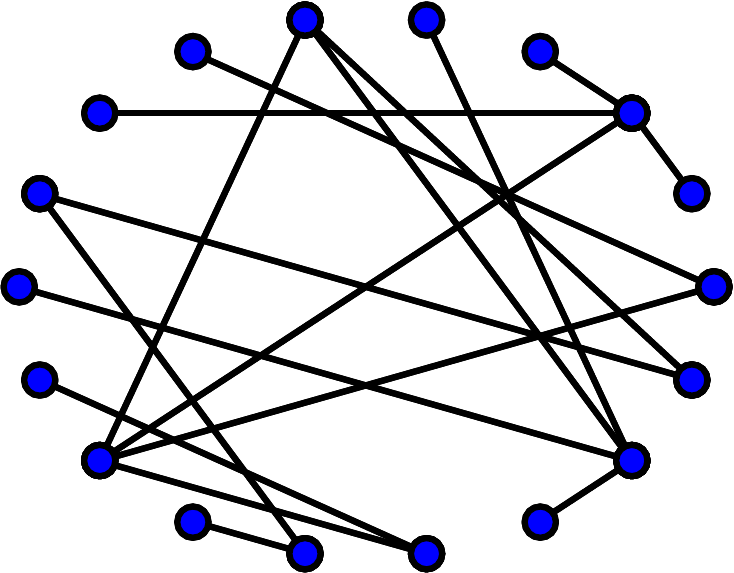}}&
\subfloat{\includegraphics[width=0.1\textwidth]{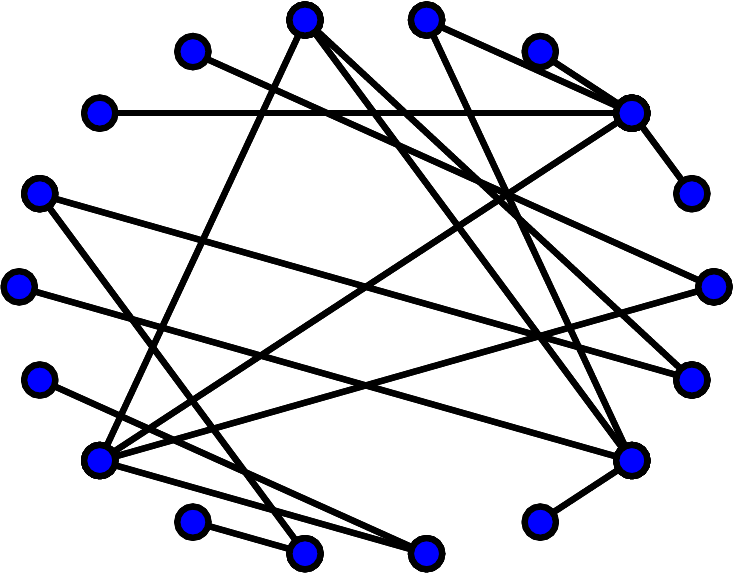}}&
\subfloat{\includegraphics[width=0.1\textwidth]{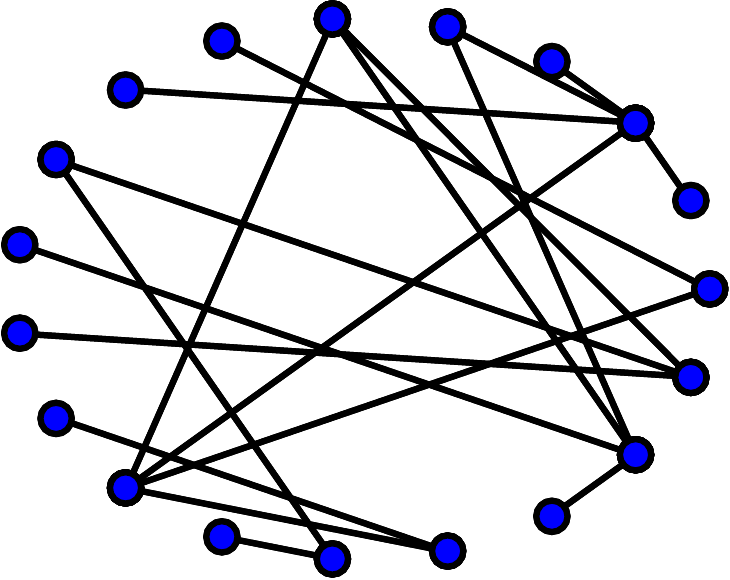}}&
\subfloat{\includegraphics[width=0.1\textwidth]{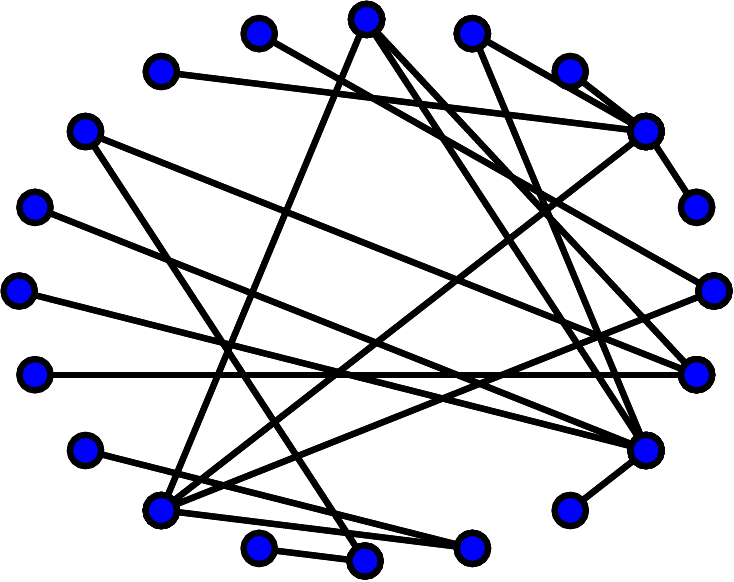}}\\
\subfloat{\includegraphics[width=0.1\textwidth]{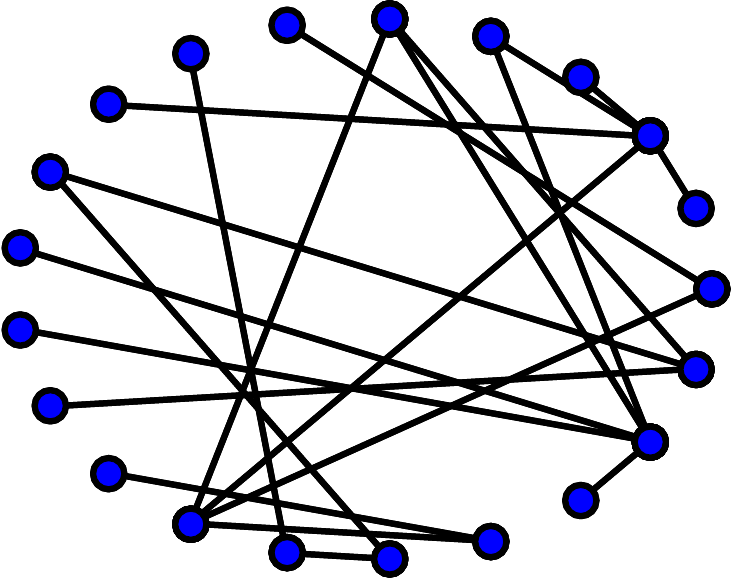}}&
\subfloat{\includegraphics[width=0.1\textwidth]{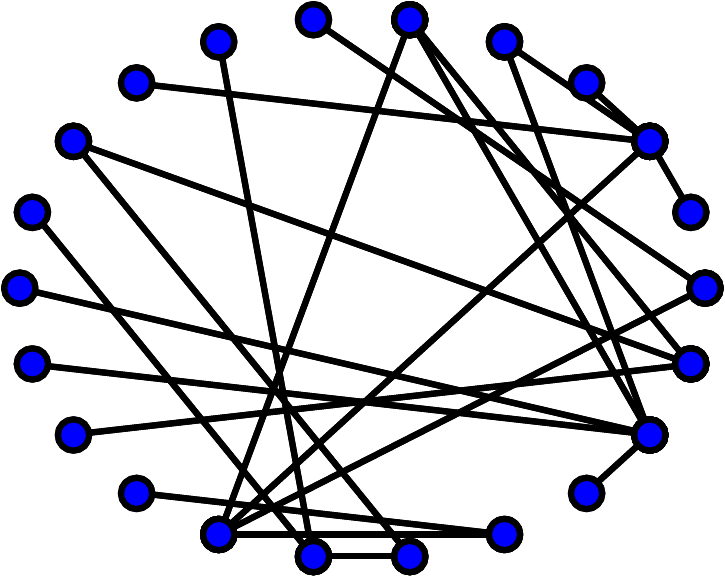}}&
\subfloat{\includegraphics[width=0.1\textwidth]{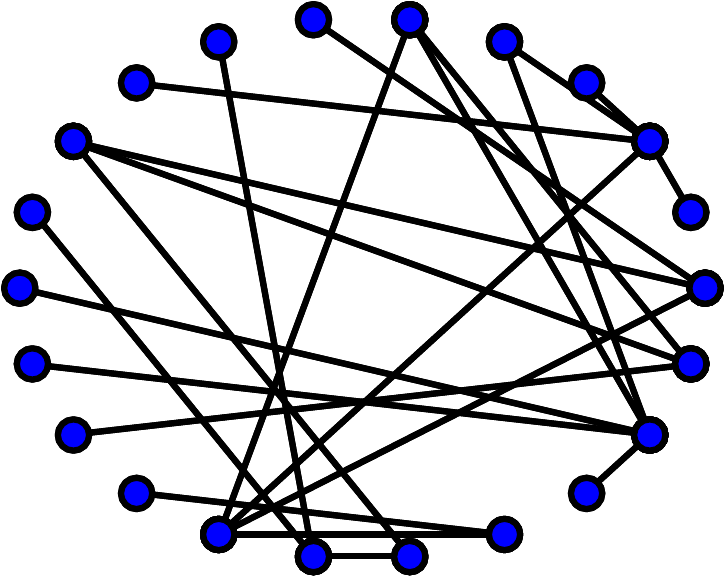}}&
\subfloat{\includegraphics[width=0.1\textwidth]{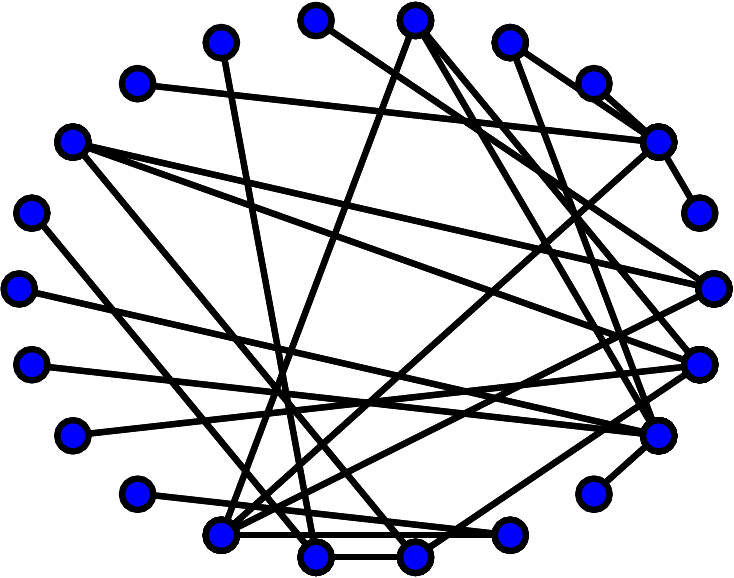}}&
\subfloat{\includegraphics[width=0.1\textwidth]{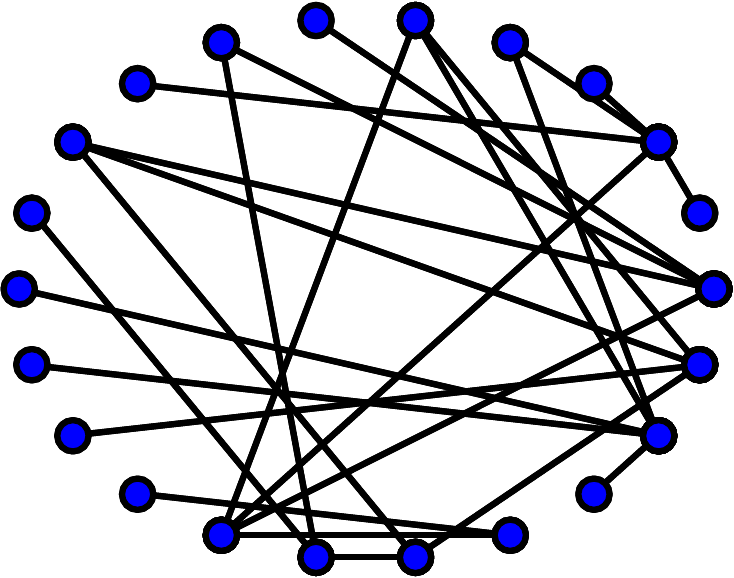}}&
\subfloat{\includegraphics[width=0.1\textwidth]{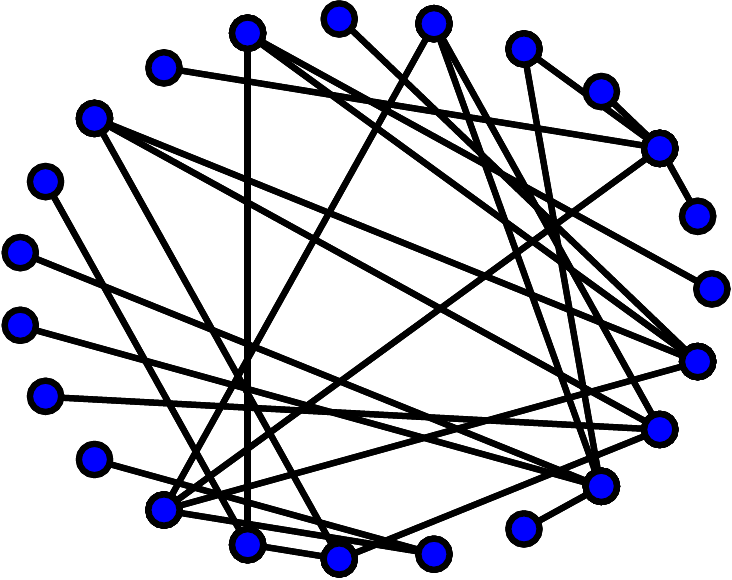}}&
\subfloat{\includegraphics[width=0.1\textwidth]{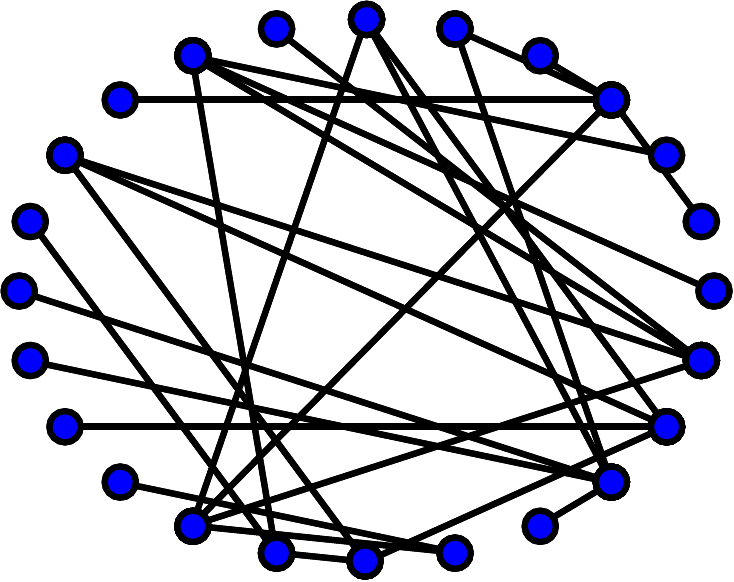}}&
\subfloat{\includegraphics[width=0.1\textwidth]{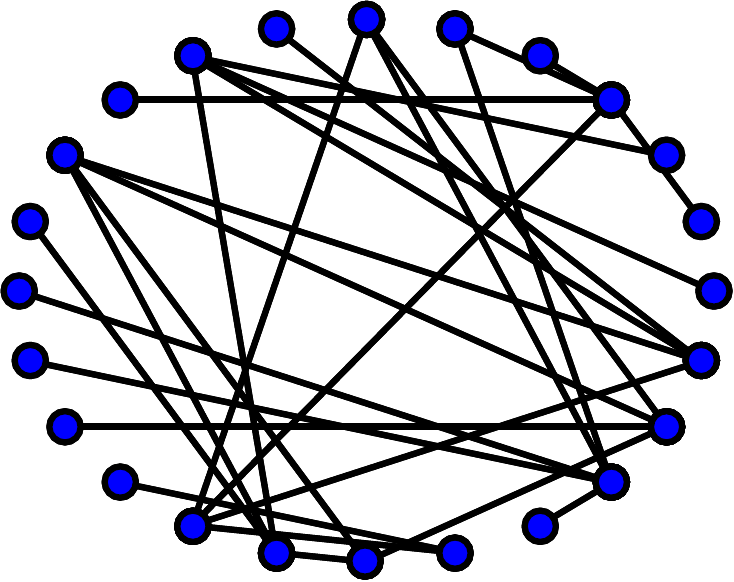}}&
\subfloat{\includegraphics[width=0.1\textwidth]{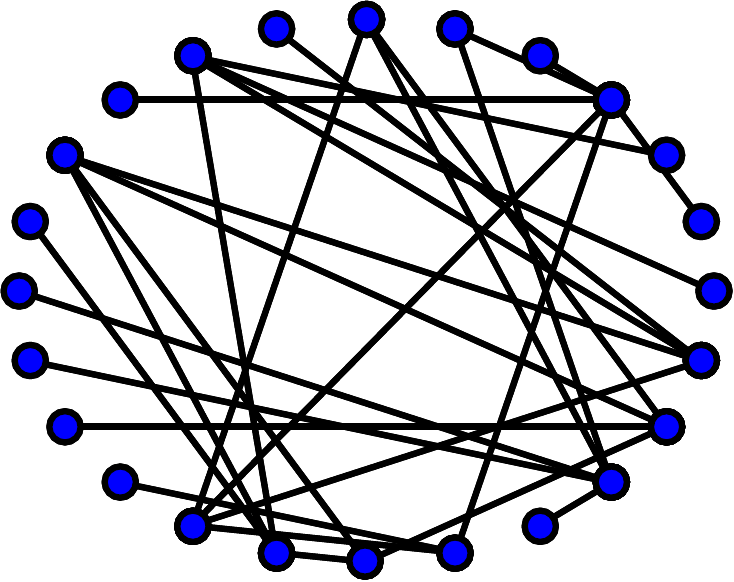}}&
\subfloat{\includegraphics[width=0.1\textwidth]{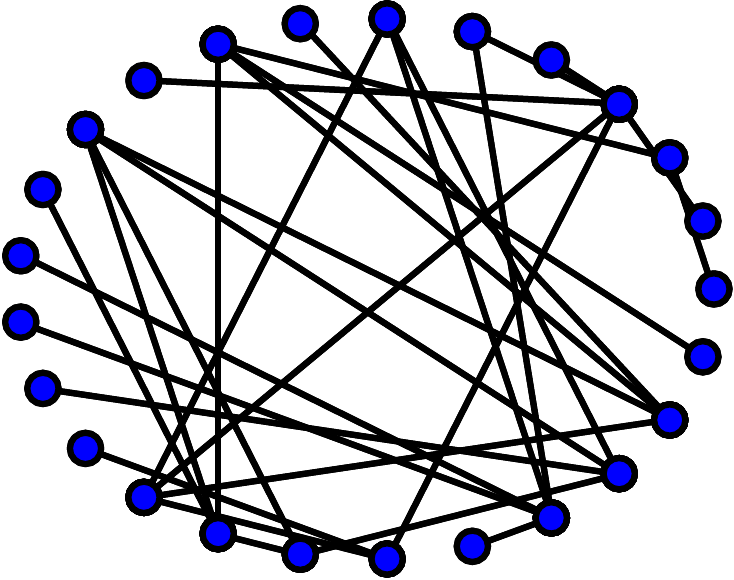}}\\
\subfloat{\includegraphics[width=0.1\textwidth]{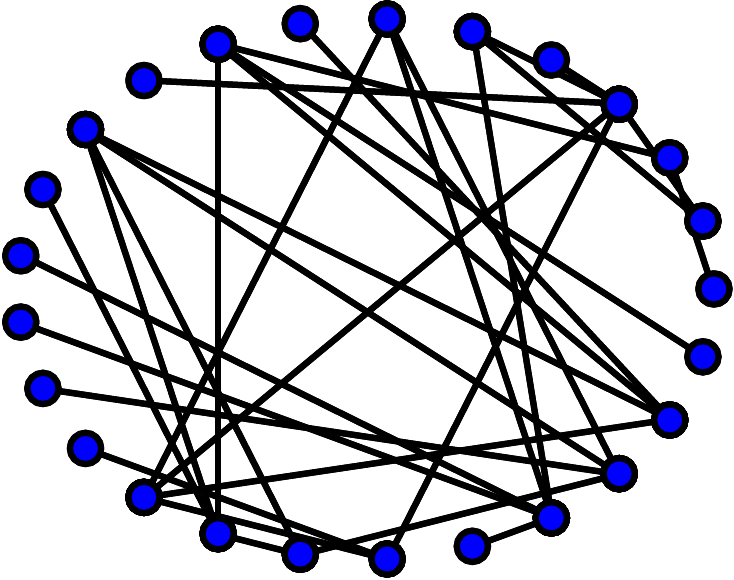}}&
\subfloat{\includegraphics[width=0.1\textwidth]{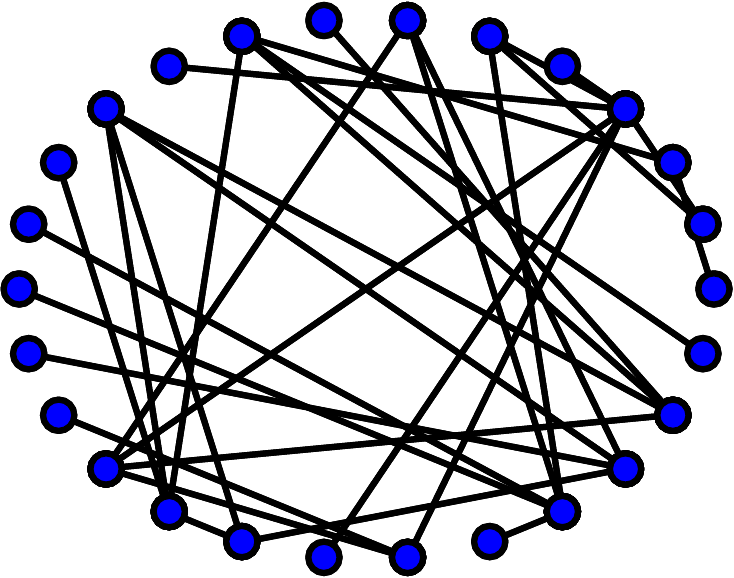}}&
\subfloat{\includegraphics[width=0.1\textwidth]{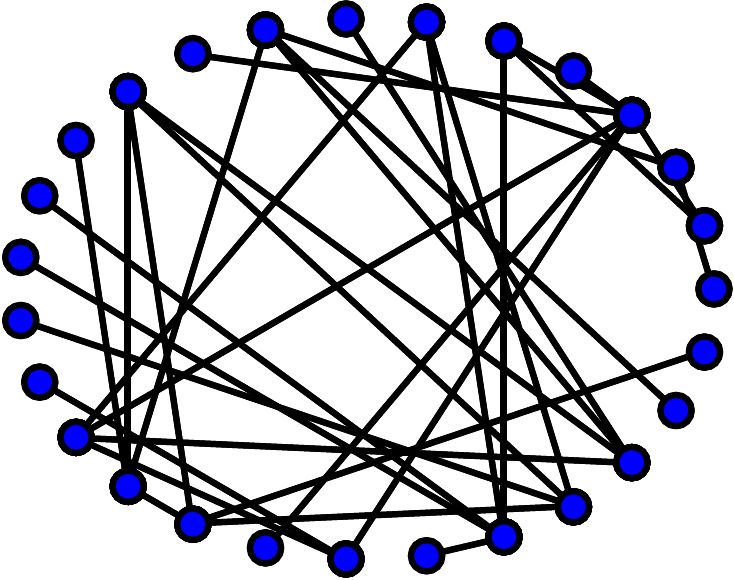}}&
\subfloat{\includegraphics[width=0.1\textwidth]{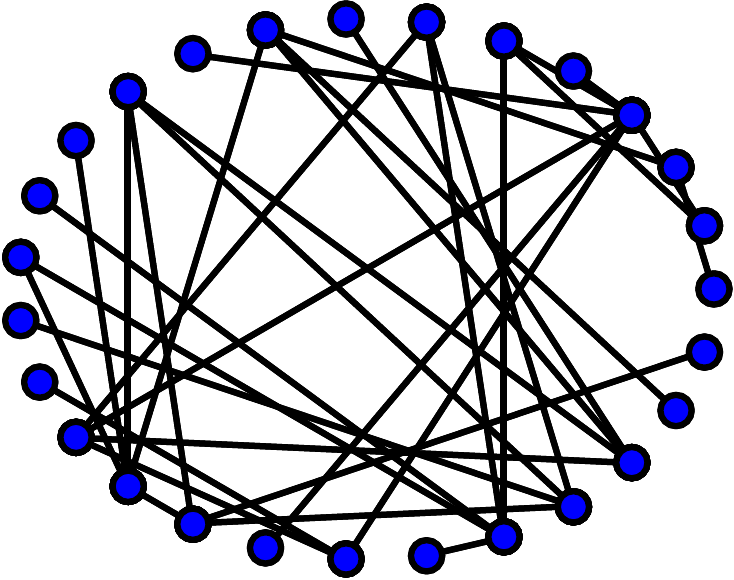}}&
\subfloat{\includegraphics[width=0.1\textwidth]{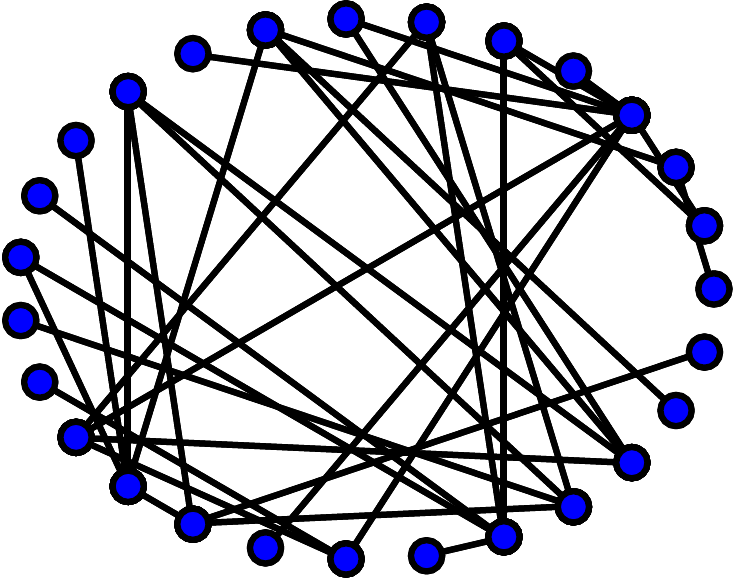}}&
\subfloat{\includegraphics[width=0.1\textwidth]{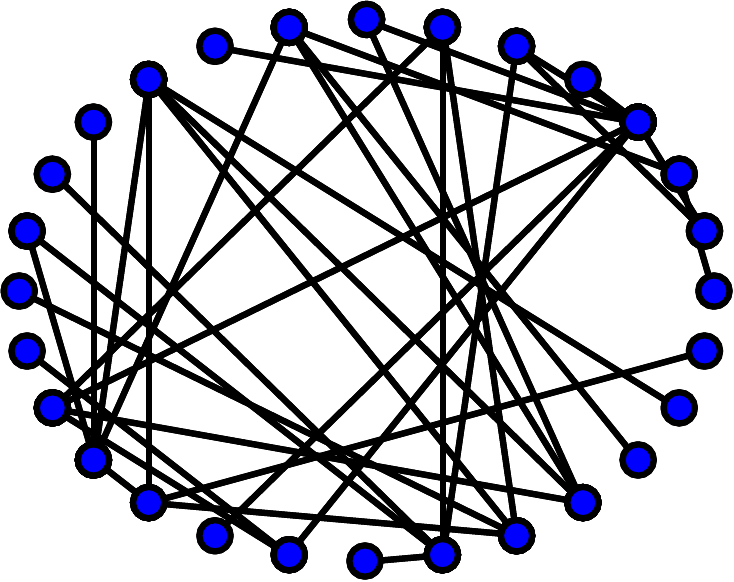}}&
\subfloat{\includegraphics[width=0.1\textwidth]{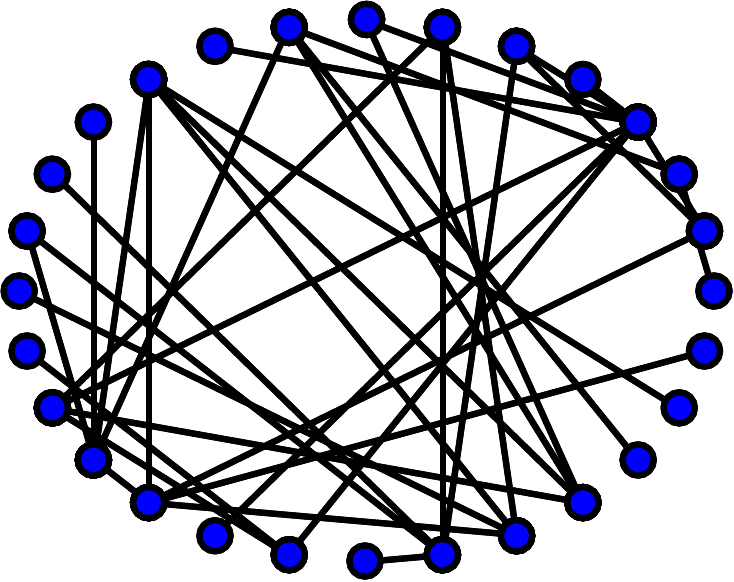}}&
\subfloat{\includegraphics[width=0.1\textwidth]{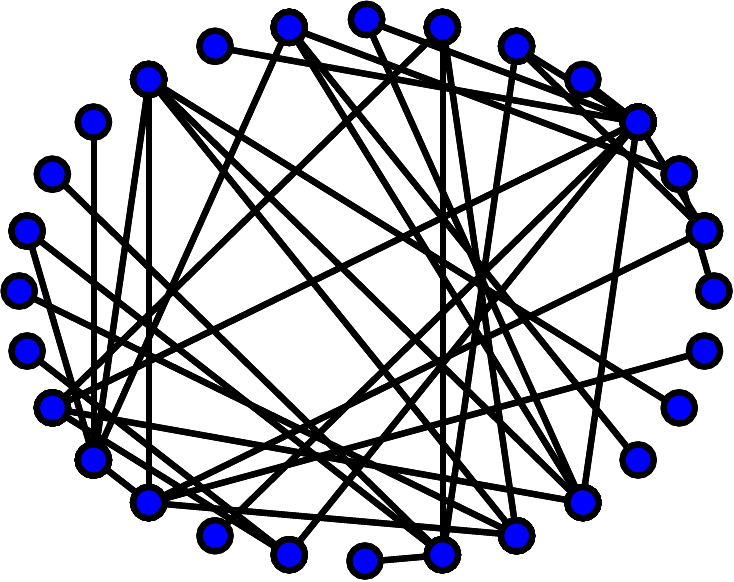}}&
\subfloat{\includegraphics[width=0.1\textwidth]{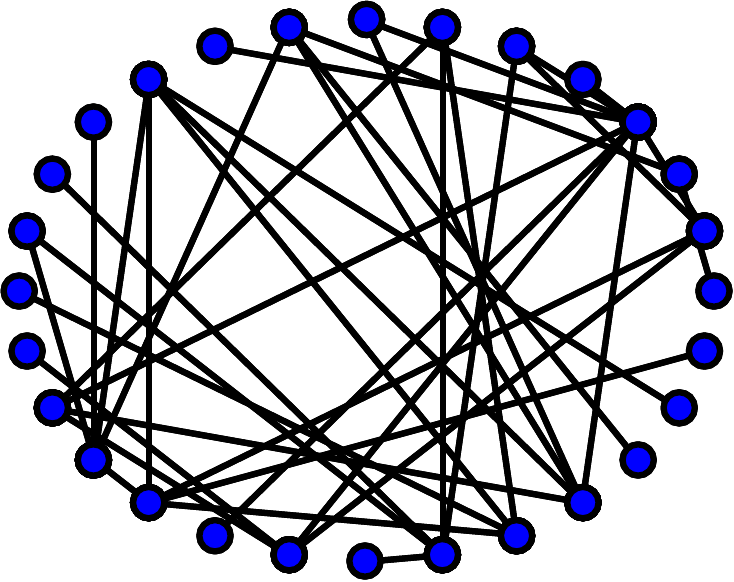}}&
\subfloat{\includegraphics[width=0.1\textwidth]{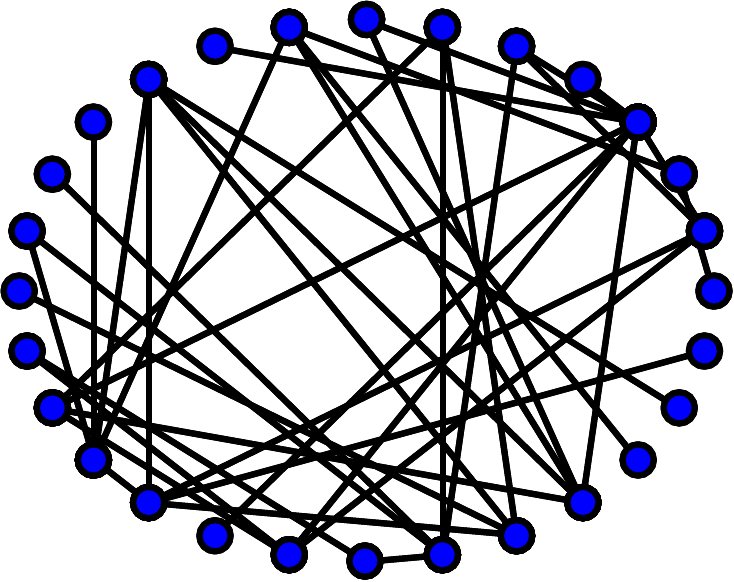}}
\end{tabular}
}
\caption{Example of graphlets with an increasing number of edges, for generating these particular examples we have used $T=40$. This shows that our stochastic search algorithm is not restricted to small orders.}
\label{fig:example1-graphlets}
\end{figure*}

\begin{figure*}[!t]
\begin{center}
\resizebox{\textwidth}{!}{
\begin{tabular}{|cc|cc|cc|cc|}
\hline
\includegraphics[width=0.2\textwidth]{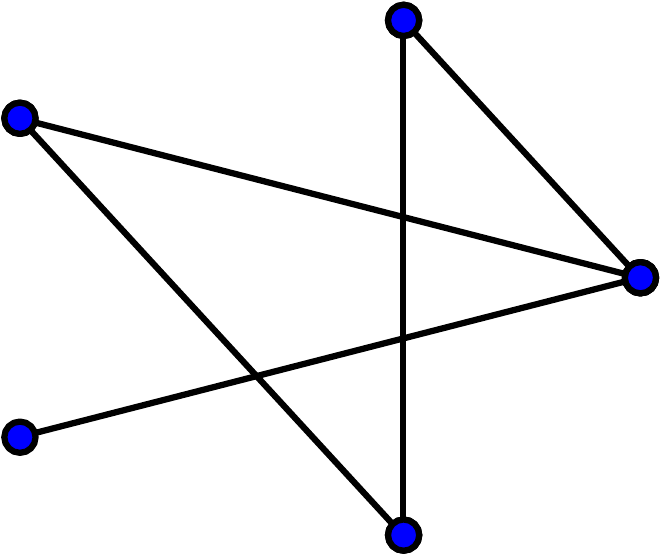} & \includegraphics[width=0.2\textwidth]{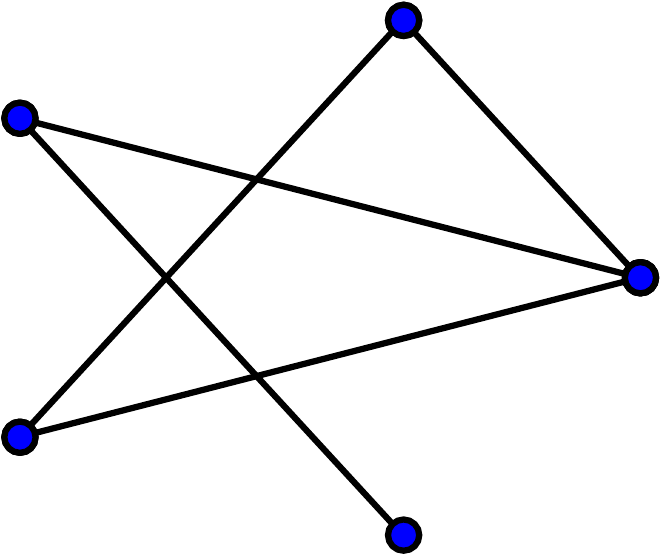} & \includegraphics[width=0.2\textwidth]{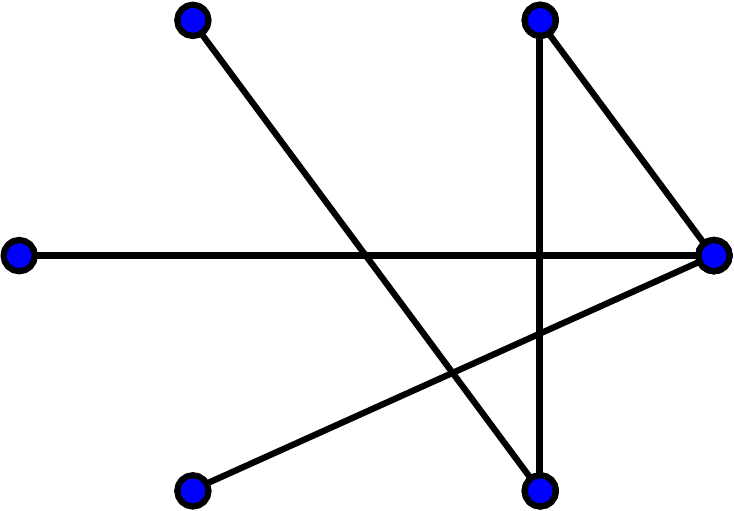} & \includegraphics[width=0.2\textwidth]{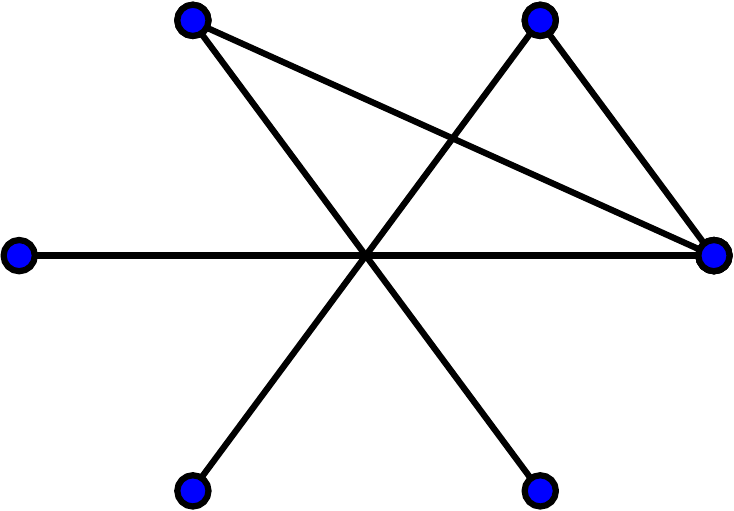} & \includegraphics[width=0.2\textwidth]{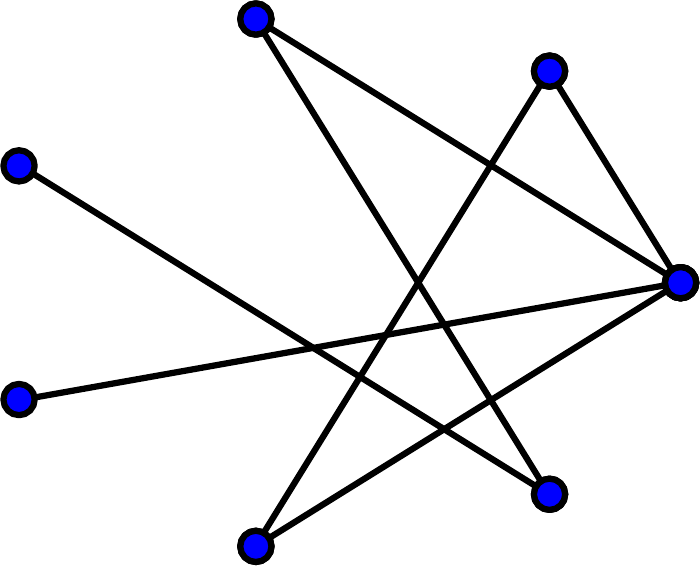} & \includegraphics[width=0.2\textwidth]{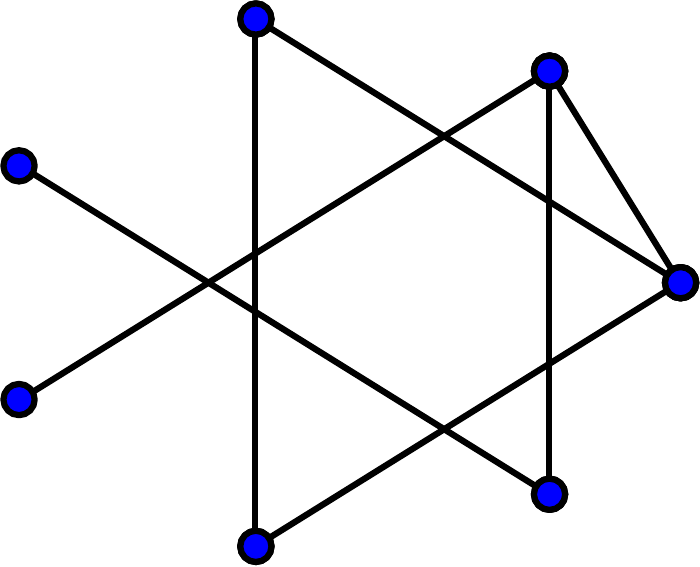} & \includegraphics[width=0.2\textwidth]{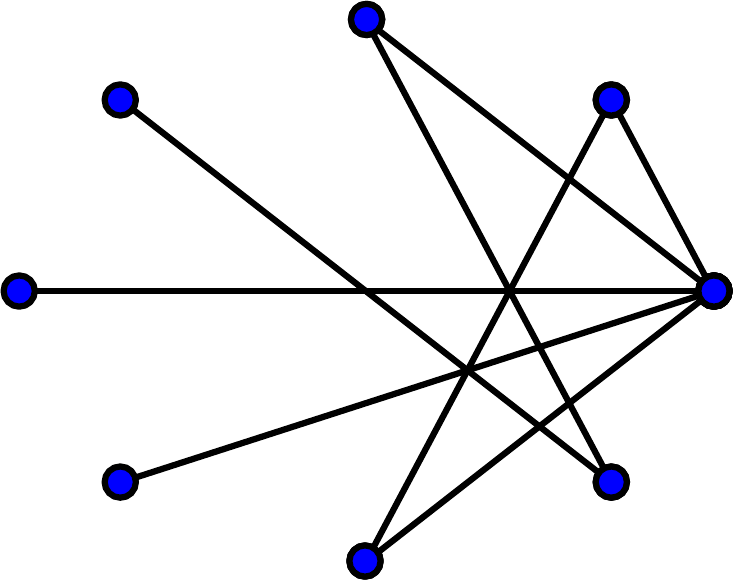} & \includegraphics[width=0.2\textwidth]{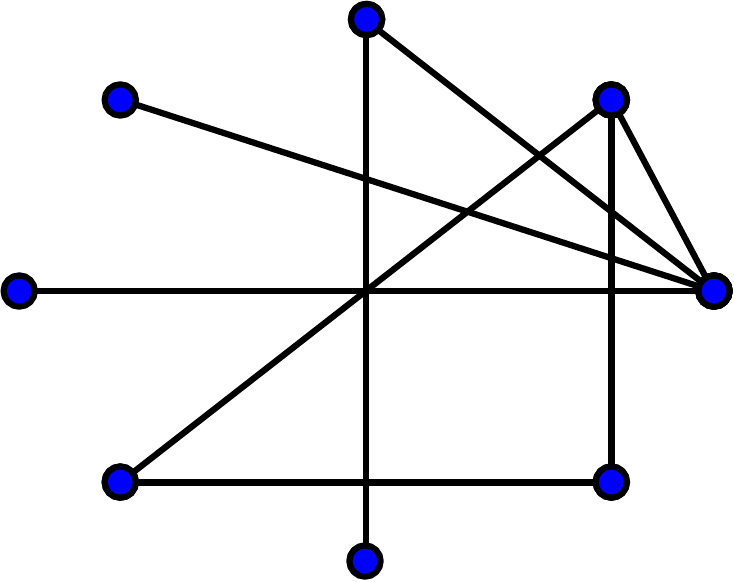} \\ \hline
$[1,2,2,2,3]$ & $[1,2,2,2,3]$ & $[1,1,1,2,2,3]$ & $[1,1,1,2,2,3]$ & $[0,0,0,0,10,16,22]$ & $[0,0,0,0,10,16,22]$ & $[0,0,0,0,0,12,20,34]$ & $[0,0,0,0,0,12,20,34]$ \\ \hline
\multicolumn{8}{c}{}\\
\multicolumn{2}{c}{\Large(a)} & \multicolumn{2}{c}{\Large(b)} & \multicolumn{2}{c}{\Large(c)} & \multicolumn{2}{c}{\Large(d)} \\
\multicolumn{8}{c}{}\\ \hline
\includegraphics[width=0.2\textwidth]{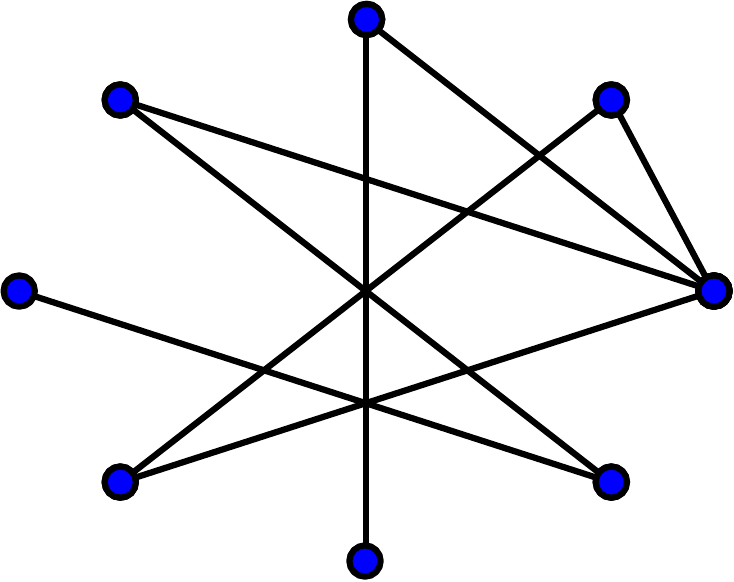} & \includegraphics[width=0.2\textwidth]{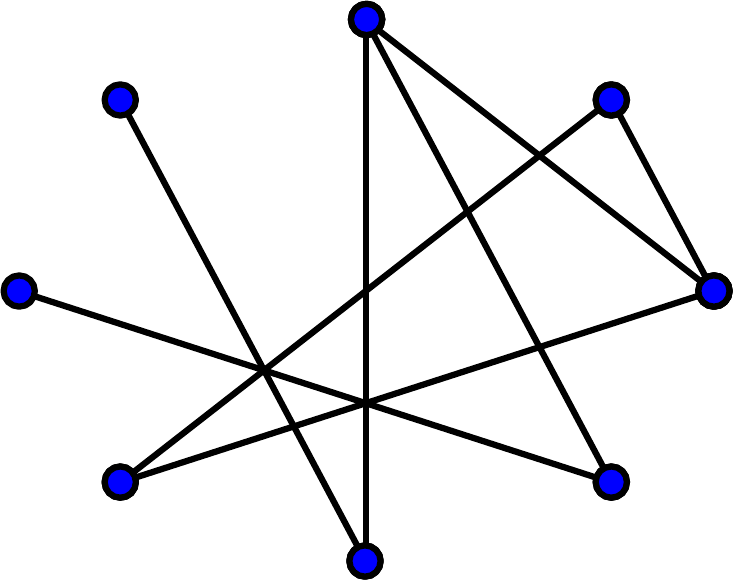} & \includegraphics[width=0.2\textwidth]{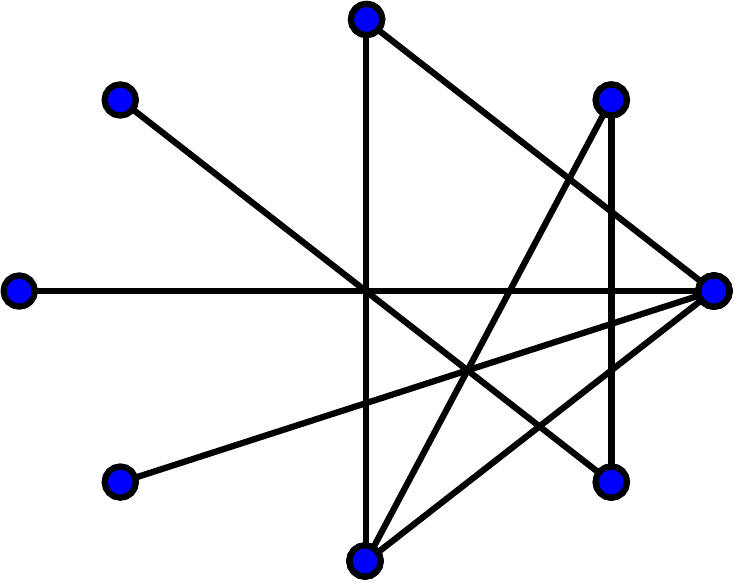} & \includegraphics[width=0.2\textwidth]{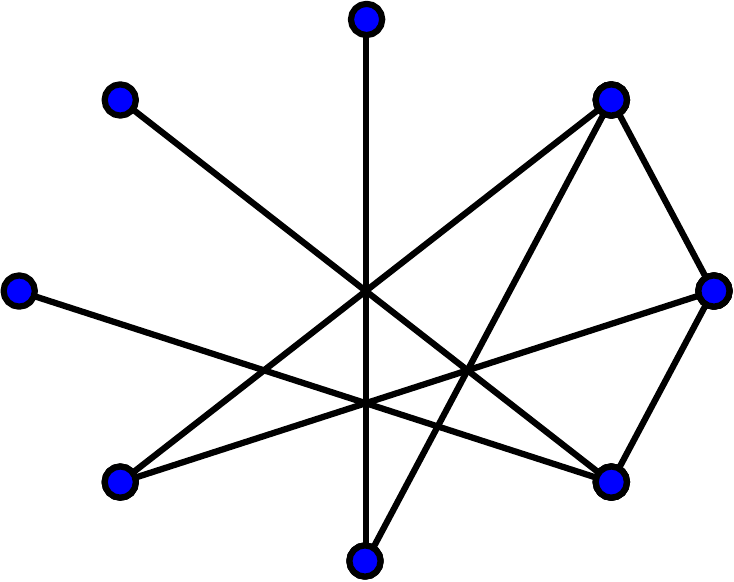} & \includegraphics[width=0.2\textwidth]{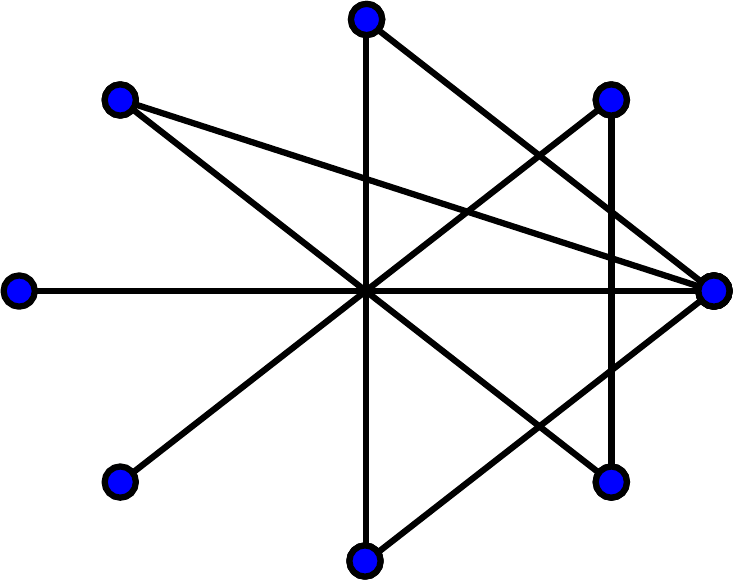} & \includegraphics[width=0.2\textwidth]{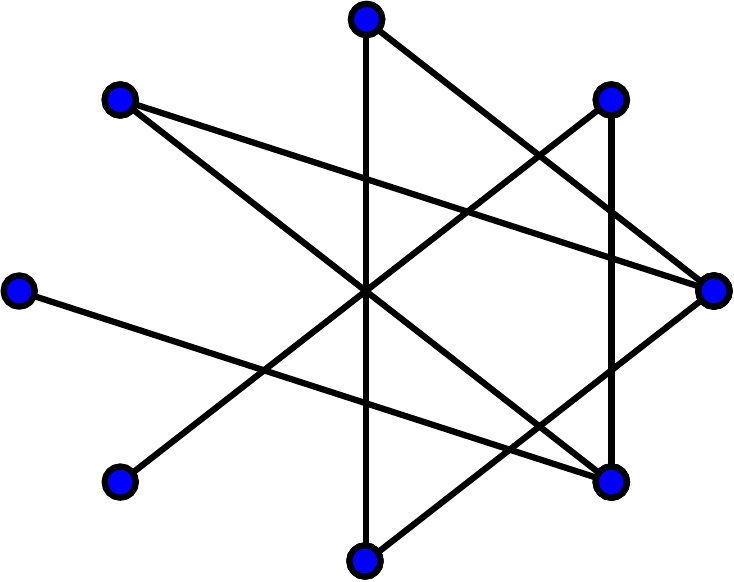} & \includegraphics[width=0.2\textwidth]{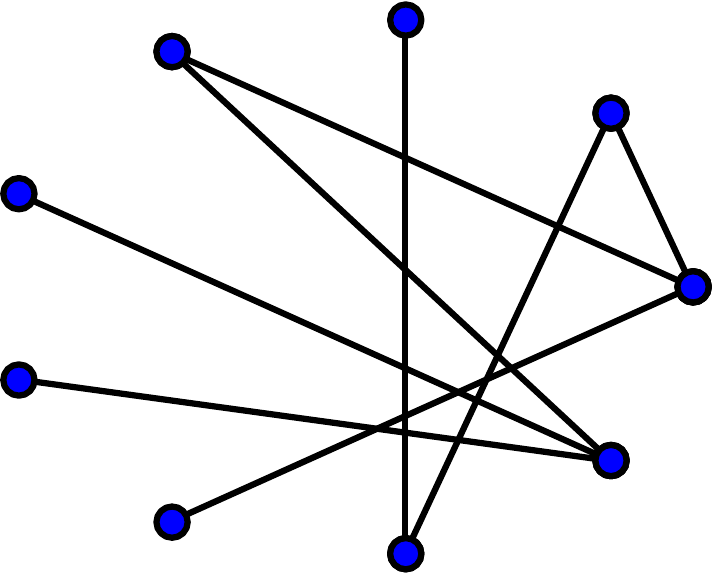} & \includegraphics[width=0.2\textwidth]{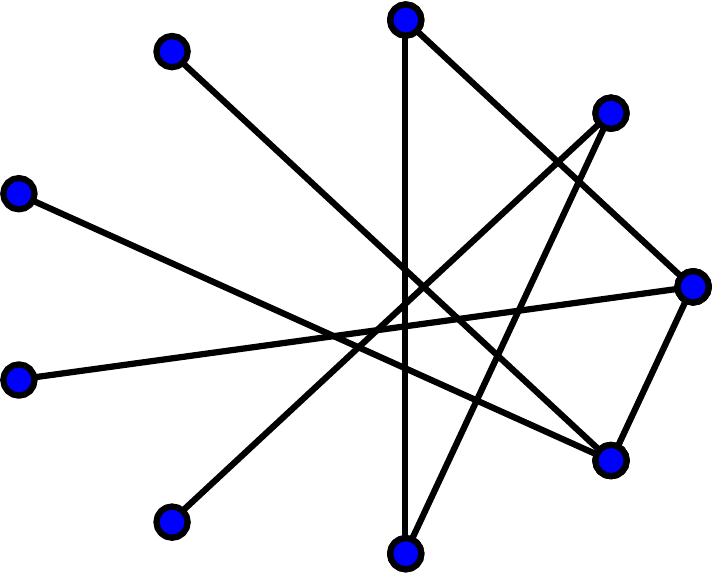} \\ \hline
$[0,0,0,0,12,12,20,32]$ & $[0,0,0,0,12,12,20,32]$ & $[0,0,0,0,12,20,22,24]$ & $[0,0,0,0,12,20,22,24]$ & $[0,0,0,0,12,20,24,28]$ & $[0,0,0,0,12,20,24,28]$ & $[0,0,0,0,14,24,26,30,38]$ & $[0,0,0,0,14,24,26,30,38]$ \\ \hline
\multicolumn{8}{c}{}\\
\multicolumn{2}{c}{\Large(e)} & \multicolumn{2}{c}{\Large(f)} & \multicolumn{2}{c}{\Large(g)} & \multicolumn{2}{c}{\Large(h)} \\
\end{tabular}}
\end{center}
\caption{Examples of non-isomorphic graphlets with the same hash codes (shown just below the respective graphlets) for different hash functions: (a)-(b) Two pairs of non-isomorphic graphlets (with $t=5$) that have the same degree values, (c) A pair of non-isomorphic graphlets (with $t=7$) that have the same betweenness centrality values, (d)-(h) Five pairs of non-isomorphic graphlets (with $t=8$) that have the same betweenness centrality values.}
\label{fig:example-non-iso-same-hash-code}
\end{figure*}

The goal of our graphlet hashing is to assign and count the frequency of graphlets (in $G$) whose hash codes fall into the {\it bins} of a global hash table (referred to as $\mathbf{HashTable}$ in \alg{alg:graphlethistogram}); each bin in this table is associated with a subset of isomorphic graphlets (see \alg{alg:graphlethistogram} and \lin{alg:graphlethistogram:hist}). These hash codes are related to the topological properties of graphlets which should ideally be identical for isomorphic graphlets and different for non-isomorphic ones (see~\cite{Dahm2013} for a detailed discussion about these topological properties). When using appropriate hash functions (see~\sect{ssec:seslct-hash-fns}), this algorithm, even though not tackling the subgraph isomorphism, is able to {\it count} the number of isomorphic subgraphs in a given graph with a controlled (polynomial) complexity.
 
\begin{algorithm}
\caption{\textsc{Hashed-Graphlets-Statistics}($G$): Create a histogram $\mathbf{H}$ of graphlet distribution for a graph $G$.}
\label{alg:graphlethistogram}
\begin{algorithmic}[1]
\small
\REQUIRE $G$, $\mathbf{HashTable}$
\ENSURE $\mathbf{H}$
\STATE $\mathbb{S} \leftarrow \textsc{Stochastic-Graphlet-Parsing}(G)$
\STATE $\mathbf{H}_i\leftarrow 0$, $i=1,\dots,|\mathbb{S}|$
\FORALL {$g \in \mathbb{S}$}
\STATE $\mathbf{hashcode} \leftarrow \textsc{HashFunction}(g)$\label{alg:graphlethistogram:hashcode}
\IF {$\mathbf{hashcode}\notin\mathbf{HashTable}$}
\STATE $\mathbf{HashTable}\leftarrow \mathbf{HashTable} \cup \mathbf{\{hashcode\}} $
\ENDIF
\STATE $i \leftarrow \textsc{GetIndex-In-HashTable}(\mathbf{hashcode})$\label{alg:graphlethistogram:idx}
\STATE $\mathbf{H}_i \leftarrow \mathbf{H}_i + 1$\label{alg:graphlethistogram:hist}
\ENDFOR
\end{algorithmic}
\end{algorithm}

\begin{figure*}[!htbp]
\centering
\resizebox{\textwidth}{!}{
\begin{tabular}{ccccccc}
\subfloat[$1$]{\includegraphics[width=0.1\textwidth]{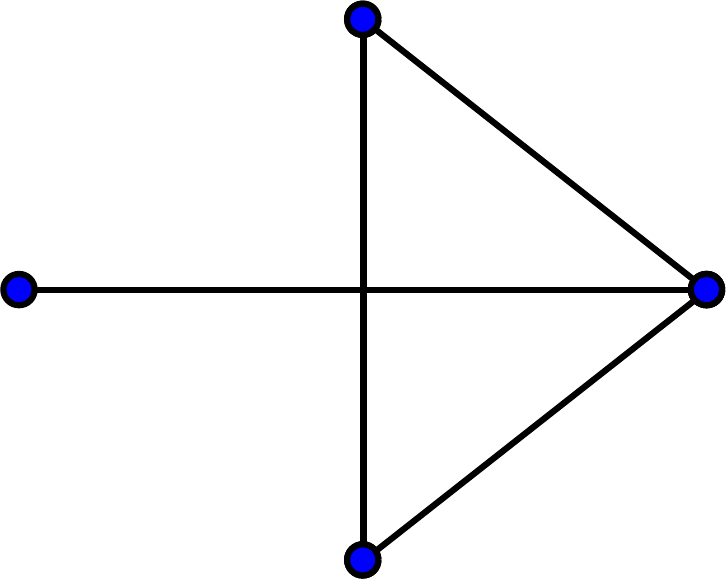}} & \subfloat[$2$]{\includegraphics[width=0.1\textwidth]{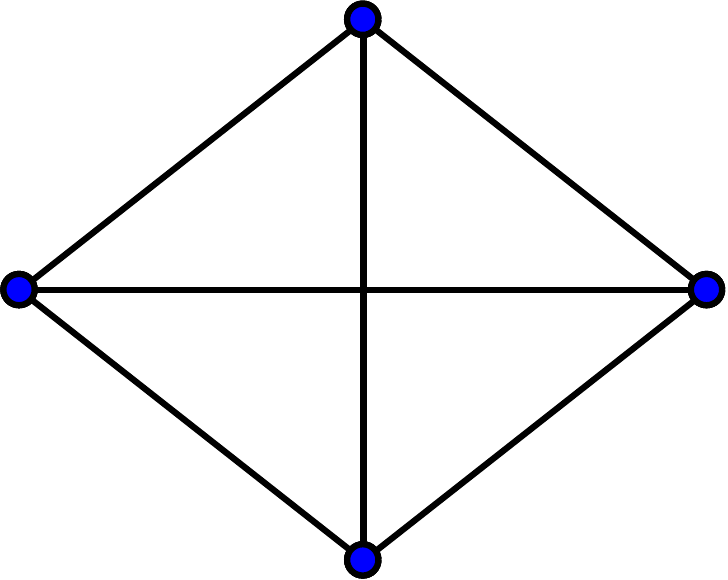}} & \subfloat[$3$]{\includegraphics[width=0.1\textwidth]{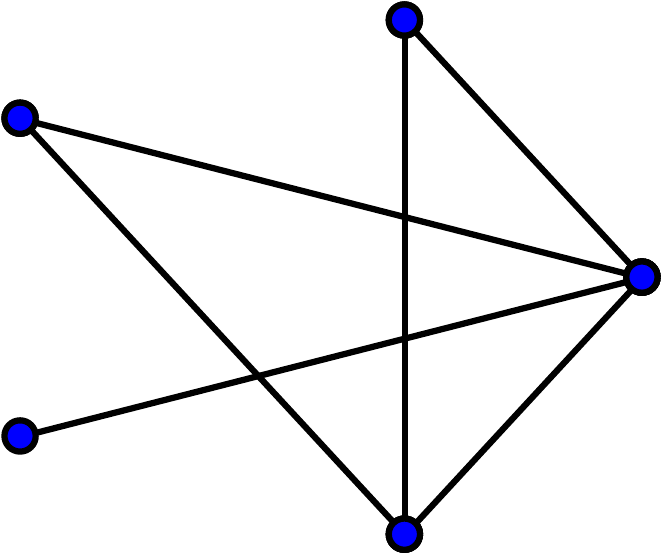}} & \subfloat[$4$]{\includegraphics[width=0.1\textwidth]{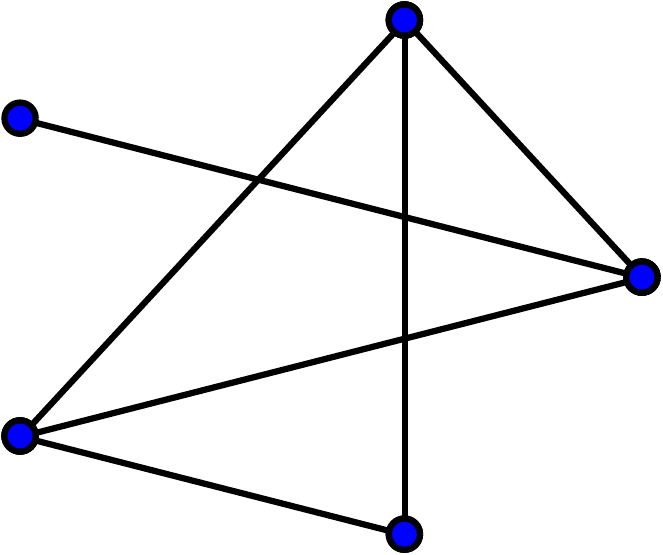}} & \subfloat[$5$]{\includegraphics[width=0.1\textwidth]{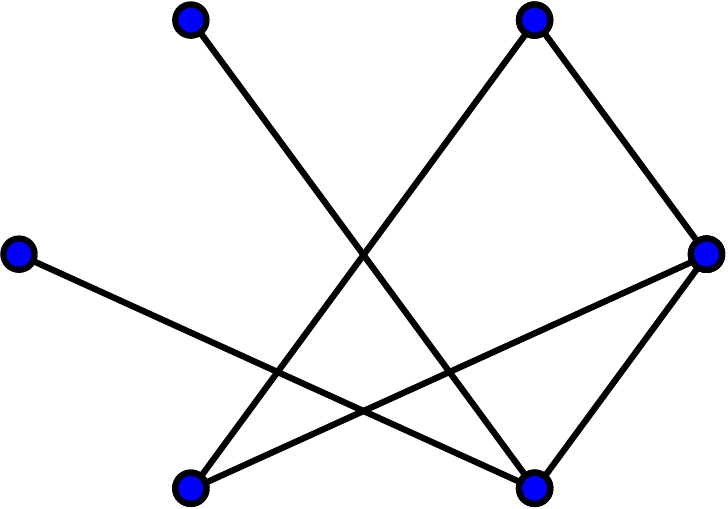}} & \subfloat[$6$]{\includegraphics[width=0.1\textwidth]{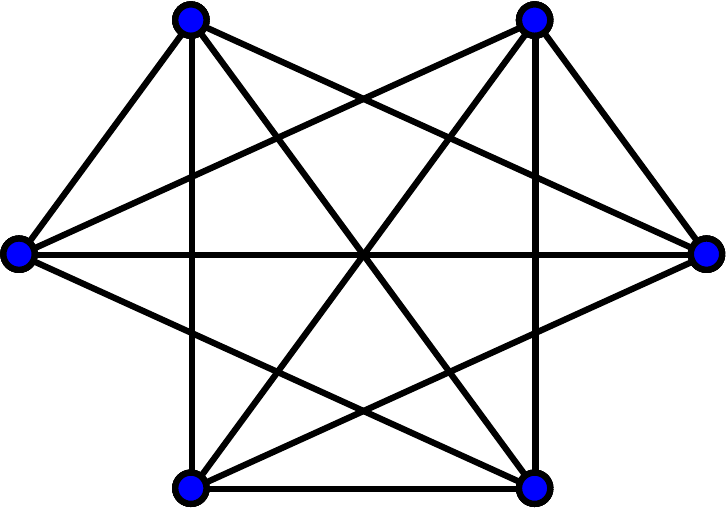}} & \multirow{2}[2]{*}[1.5cm]{\subfloat[]{\includegraphics[width=0.3\textwidth]{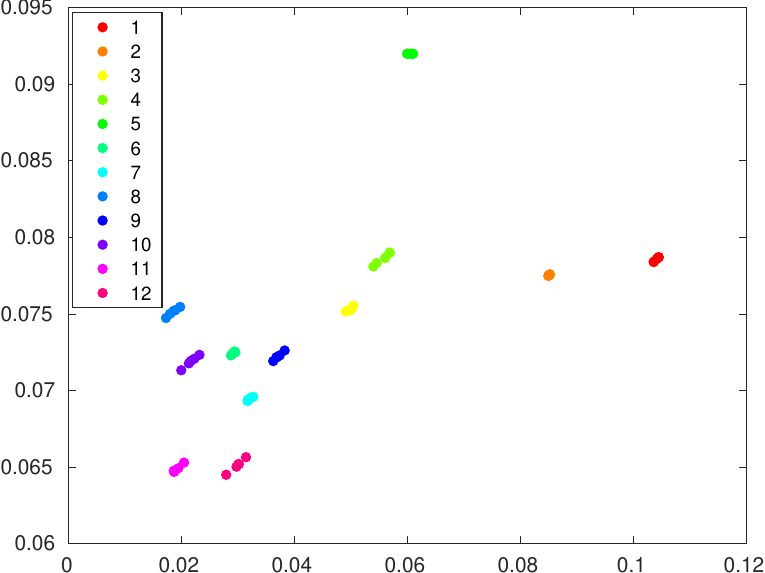}\label{sfig:plot_sge}}}\\
\subfloat[$7$]{\includegraphics[width=0.1\textwidth]{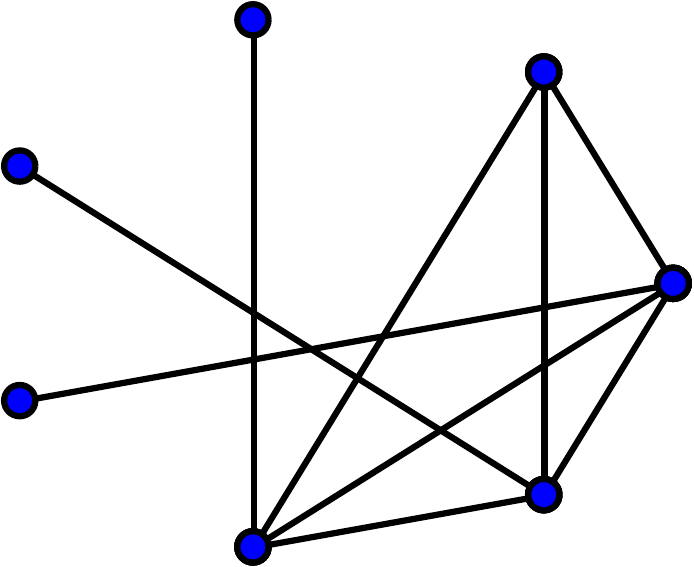}} & \subfloat[$8$]{\includegraphics[width=0.1\textwidth]{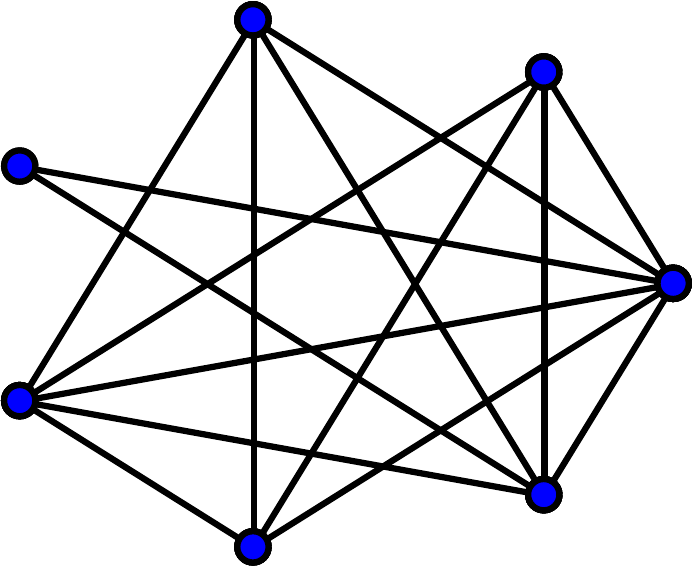}} & \subfloat[$9$]{\includegraphics[width=0.1\textwidth]{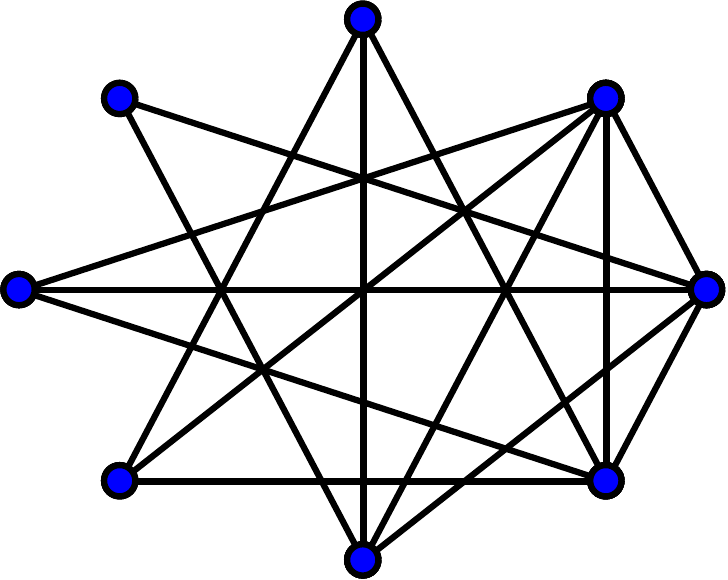}} & \subfloat[$10$]{\includegraphics[width=0.1\textwidth]{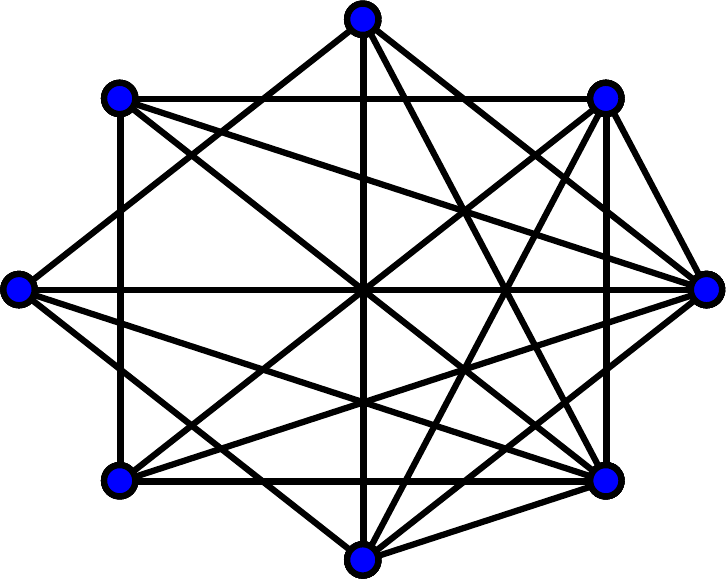}} & \subfloat[$11$]{\includegraphics[width=0.1\textwidth]{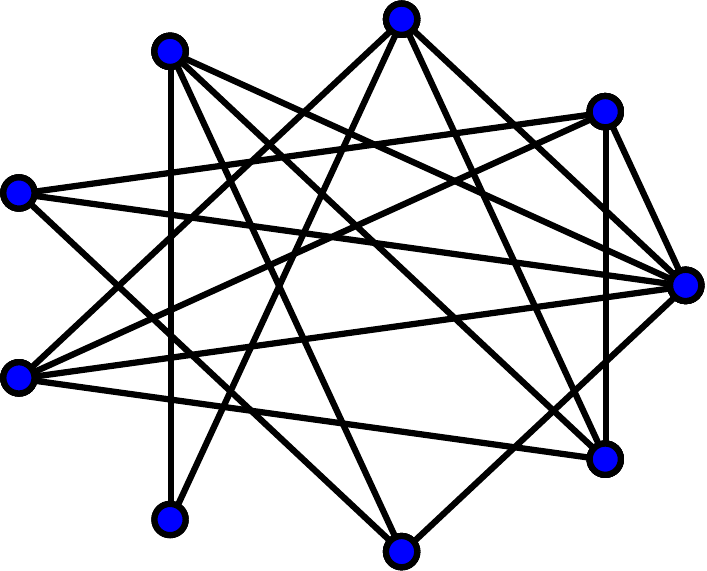}} & \subfloat[$12$]{\includegraphics[width=0.1\textwidth]{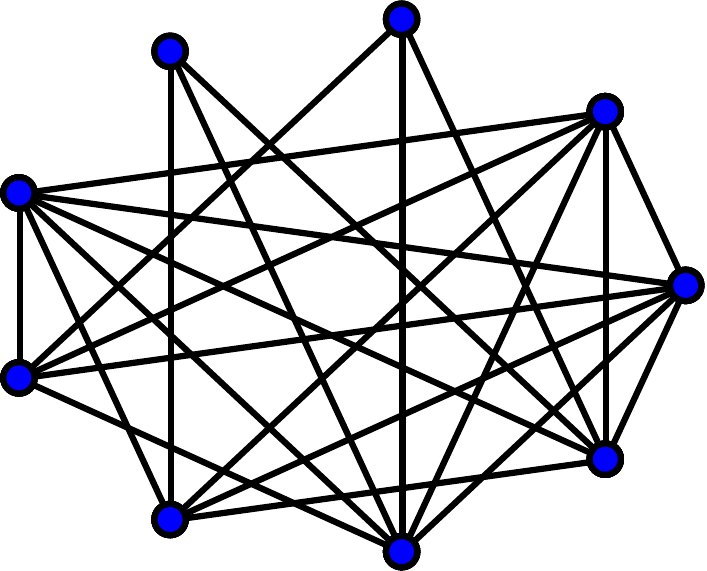}} & \\
\end{tabular}
}
\caption{{(a)--(m) An example of twelve graphs which are mutually non-isomorphic; these graphs are representatives of twelve groups with each one including a subset of (5+1) isomorphic graphs (only the twelve representatives of these groups are shown in this figure). (g) In this 2D plot, points with different colors stand for non-isomorphic graph groups (whose representatives are shown in (a)--(m)) while points with the same colors stand for isomorphic graphs. (Best viewed in pdf)}}
\label{fig:plot_sge}
\end{figure*}

Two types of hash functions exist in the literature: \emph{local} and \emph{holistic}. Holistic functions are computed globally on a given graphlet and include number of nodes/edges, sum/product of node labels, and frequency distribution of node labels, while local functions are computed at the node level; among these functions
\begin{itemize}
\item \emph{Local clustering coefficient} of a node $u$ in a graph is the ratio between the number of triangles connected to $u$ and the number of triples centered around $u$. The local clustering coefficient of a node in a graph quantifies how close its neighbors are for being a clique.
\item \emph{Betweenness centrality} of a node $u$ is the number of shortest paths from all nodes to all others that pass through the node $u$. In a generic graph, betweenness centrality of a node provides a measurement about the centrality of that node with respect to the entire graph.
\item \emph{Core number} of a node $u$ is the largest integer $c$ such that the node $u$ has degree greater than zero when all the nodes of degree less than $c$ are removed.
\item \emph{Degree} of a node $u$ is the number of edges connected to the node $u$.
\end{itemize}
As these local measures are sensitive to the ordering of nodes in graphlets, we sort and concatenate them in order to obtain global permutation invariant hash codes.
\begin{table}[!ht]
    \centering
    \textcolor{black}{
    \caption{\textcolor{black}{Examples of speedup factors (with different settings of $t$, $\epsilon$ and $\delta$) of our hashing-based method vs. graph isomorphism, on the MUTAG database (see details about MUTAG later in experiments). }}\label{speedups}
    \resizebox{\columnwidth}{!}{
    \begin{tabular}{|l|c||l|c|}
        \hline
       Setting & Speedup & Setting & Speedup \\
        \hline
         ($t=3, \epsilon = 0.1, \delta = 0.1$) & $121\times$ & ($t=5, \epsilon = 0.1, \delta = 0.1$) & $239\times$ \\
          ($t=3, \epsilon = 0.1, \delta = 0.05$) & $124\times$ & ($t=5, \epsilon = 0.1, \delta = 0.05$) & $252\times$ \\
         ($t=3, \epsilon = 0.05, \delta = 0.1$) & $163\times$ & ($t=5, \epsilon = 0.05, \delta = 0.1$) & $297\times$ \\
          ($t=3, \epsilon = 0.05, \delta = 0.05$) & $173\times$ & ($t=5, \epsilon = 0.05, \delta = 0.05$) & $318\times$ \\
        \hline
         ($t=4, \epsilon = 0.1, \delta = 0.1$) & $154\times$ & ($t=6, \epsilon = 0.1, \delta = 0.1$) & $303\times$ \\
          ($t=4, \epsilon = 0.1, \delta = 0.05$) & $161\times$ & ($t=6, \epsilon = 0.1, \delta = 0.05$) & $319\times$ \\
          ($t=4, \epsilon = 0.05, \delta = 0.1$) & $214\times$ & ($t=6, \epsilon = 0.05, \delta = 0.1$) & $356\times$ \\
          ($t=4, \epsilon = 0.05, \delta = 0.05$) & $242\times$ & ($t=6, \epsilon = 0.05, \delta = 0.05$) & $371\times$ \\
        \hline
    \end{tabular}}
    }
\end{table}

\def\AA{{\cal I}}

\subsection{Hash Function Selection}
\label{ssec:seslct-hash-fns}
Ideally, a \emph{reliable} hash function is expected to provide identical hash codes for two isomorphic graphlets and two different hash codes for two non-isomorphic ones. While it is easy to design hash functions that provide identical hash codes for isomorphic graphlets, it is very challenging to guarantee that non-isomorphic graphlets could never be mapped to the same hash code. This is also in accordance with the fact that graph isomorphism detection is GI-complete and no polynomial algorithm is known to solve it. The possibility of mapping two non-isomorphic graphlets to the same hash code is termed as \emph{a collision}. Let $f$ be a function that returns a hash code for a given graphlet, then the probability of collision of that function is defined as
\begin{equation*}
E(f) = P\big((g,g') \in \AA_0 \ | \ f(g)= f(g') \big),
\end{equation*}
here $g$, $g'$ denote two graphlets, and the probability is with respect to $\AA_0$ which stands for pairs of non-isomorphic graphlets; equivalently, we can define $\AA_1$ as the pairs of isomorphic graphlets. Since the cardinality of $\AA_0$ is really huge for graphlets with large number of edges, \ie, $|\AA_1|\ll |\AA_0|$, one may instead consider
\begin{equation*}
E(f) = 1 - P\big((g,g') \in \AA_1 \ | \ f(g)= f(g') \big),
\end{equation*}
which also results from the fact that our hash functions produce same codes for isomorphic graphlets. For bounded $t$ ($t\leq T$), the evaluation of $E(f)$ becomes tractable and reduces to
\begin{equation*}
E(f) = 1 - \frac{\sum_{g,g'} 1_{ \{(g,g') \in \AA_1 \}}}{\sum_{g,g'} 1_{ \{f(g)=f(g')\}}}.
\end{equation*}
Considering a collection of hash functions $\{f_c\}_c$, the best one is chosen as 
\begin{equation*}
f^* = \argmin_{f_c}\allowbreak E(f_c)
\end{equation*}
\tab{tab:hash-fn} shows the values of $E(f)$ for different hash functions including \emph{betweenness centrality}, \emph{core numbers}, \emph{degree} and \emph{clustering coefficients}, and for different graphlet orders (number of edges) ranging from $1$ to $10$. In order to build this table, we enumerate all the non-isomorphic graphs {~\cite{McKay2014}} with a number of edges bounded\footnote{{More details can be found at: \url{http://users.cecs.anu.edu.au/~bdm/data/graphs.html}}} by $10$ and compute the hash codes with the above mentioned hash functions to quantify the probability of collisions. First, we observe that $E(f)$ is close to $0$ as $t$ reaches large values for all the hash functions. Moreover, the hash function \emph{degree} of nodes has probability of collision equal to $0$ for graphlets with $t\leq 4$ but this probability increases for larger values of $t$, while \emph{betweenness centrality} has the lowest probability of collision for all $t$; the number of non-isomorphic graphs with the same \emph{betweenness centrality} is very small for low order graphs and increases slowly as the order increases (see for instance \fig{fig:example-non-iso-same-hash-code}) and this is in accordance with facts known in network analysis community. Indeed, two graphs with the same \emph{betweenness centrality} would indeed be isomorphic with a high probability~\cite{Newman2005,Comellas2008}; see also our MATLAB library\footnote{Available at \url{https://github.com/AnjanDutta/StochasticGraphletEmbedding/tree/master/HashFunctionGraphlets}} that reproduces the results shown in~\tab{tab:hash-fn}.

The proposed algorithm involves random sampling of graphlets and partitioning them with well designed hash functions having very low probability of collisions. This technique fetches accurate distribution of those sampled high order graphlets in a given graph and maps the isomorphic graphs to similar points and non-isomorphic ones to different points. \fig{fig:plot_sge} shows this principle for different and increasing graph orders; from this figure, it is clear that all the non-isomorphic graphs are mapped to very distinct points while isomorphic graphs are mapped to very similar points. Hence, the randomness (in graphlet parsing) does not introduce any arbitrary behavior in the graph embedding and the SGE of isomorphic graphlets converge to very similar points in spite of being {\it seeded} differently\footnote{In practice, we found that random uniform node sampling (with different seeds) is the best strategy among others including sampling nodes with \emph{highest betweenness centrality}, \emph{highest degree} and \emph{random seeds}, etc (see Table $III$ of~\cite{Dutta2018SGEsupp}).}.

\section{Computational Complexity}
\label{sec:comp-anal}
The computational complexity of our method is $O(MT)$ for \alg{alg:extractgrahletsdfs} and $O(MTC)$ for \alg{alg:graphlethistogram}, here $M$ is again the number of runs, $T$ is an upper bound on the number of edges in graphlets and $C$ is the computational complexity of the used hash function; for ``degree'' and ``betweenness centrality'' this complexity is respectively $O(|V|)$ and $O(|V||E|)$, where $|V|$ (resp. $|E|$) stands for the cardinality of node (resp. edge) set in the sampled graphlets. Hence, it is clear that the complexity of these two algorithms is not dependent on the size of the input graph $G$, but only on the parameters $M$, $T$ and the used hash functions. 

As graphlets are sampled independently, both algorithms mentioned above are trivially parallelizable. \tab{tab:comp-time} shows examples of processing time (in s) for different settings of $M$, $T$ and for single and multiple parallel CPU workers; with $M=11413$, $T=7$, our method takes $6.13$s on average (on a single CPU) in order to parse a graph and to generate the stochastic graphlets, compute their hash codes and find their respective histogram bins while it takes only $3.14$s (with 4 workers). With $M=46204$, $T=7$ this processing time reduces from $22.57$s to $5.62$s (with 4 workers) while it reduces from $1.13$s to $1.01$s when $M=4061$, $T=3$. From all these results, the parallelized setting is clearly interesting especially when $M$ and $T$ are large as the overhead time due to "task distribution" (through workers) and "result collection" (from workers) becomes negligible.
 
\begin{table}[!htbp]
\begin{center}
\caption{Computation time for different values of $M$ and $T$ both in serialized and parallel (with $4$ workers) settings.}
\label{tab:comp-time}
\resizebox{0.6\columnwidth}{!}{
\begin{tabular}{|c|c|c|c|}
\hline
\multirow{2}{*}{$M$} & \multirow{2}{*}{$T$} & \multicolumn{2}{c|}{Time in secs.} \\
\cline{3-4}
 & & Serialized & Parallel (4 workers)\\
\hline
$877$   & $3$ & $0.23$  & $0.27$\\
$4061$  & $3$ & $1.13$  & $1.01$\\
$2125$  & $5$ & $3.18$  & $2.42$\\
$9051$  & $5$ & $10.76$ & $2.83$\\
$11413$ & $7$ & $6.13$  & $3.14$\\
$46204$ & $7$ & $22.57$ & $5.62$\\
\hline
\end{tabular}}
\end{center}
\end{table}

\section{Experimental Validation}
\label{sec:results}
In order to evaluate the impact of our proposed stochastic graphlet embedding, we consider four different experiments described below. We consider graphlets (with different fixed orders) taken {\it separately} and {\it combined}; as shown subsequently, the combined setting brings a substantial gain in performances. All these experiments are shown in the remainder of this section and {\it also} in a supplemental material~\cite{Dutta2018SGEsupp}\footnote{Due to the limited number of pages in the paper, we added more extensive experiments in~\cite{Dutta2018SGEsupp}}. A Matlab library is also available in \url{https://github.com/AnjanDutta/StochasticGraphletEmbedding}.

\begin{table}[!htbp]
\begin{center}
\caption{Some details on MUTAG, PTC, ENZYMES, D\&D, NCI1 and NCI109 graph datasets.}
\label{tab:det-graph-datasets1}
\resizebox{\columnwidth}{!}{
\begin{tabular}{|l|r|l|r|r|}
\hline
Datasets & \#Graphs & Classes & Avg. \#nodes & Avg. \#edges\\\hline
MUTAG & $188$ & $2$ ($125$ vs. $63$) & $17.7$ & $38.9$\\
PTC & $344$ & $2$ ($192$ vs. $152$) & $26.7$ & $50.7$\\
ENZYMES & $600$ & $6$ ($100$ each) & $32.6$ & $124.3$\\
D\&D & $1178$ & $2$ ($691$ vs. $487$) & $284.4$ & $1921.6$\\
NCI1 & $4110$ & $2$ ($2057$ vs. $2053$) & $29.9$ & $64.6$\\
NCI109 & $4127$ & $2$ ($2079$ vs. $2048$) & $29.7$ & $64.3$\\\hline
\end{tabular}}
\end{center}
\end{table}

\begin{table*}[!htbp]
\begin{center}
\caption{Classification accuracies (in \%) on MUTAG, PTC, ENZYMES, D\&D, {NCI1 and NCI109} datasets. RW corresponds to the random walk kernel~\cite{Vishwanathan2010}, SP stands for shortest path kernel~\cite{Borgwardt2005}, GK corresponds to the standard graphlet kernel~\cite{Shervashidze2009}, {MLG stands for multiscale Laplacian graph kernel~\cite{Kondor2016}}, and SGE refers to our proposed stochastic graphlet embedding. The average processing time for generating the stochastic graphlet embedding of a given graph is indicated within the parenthesis after each accuracy value. In these results, ``$> 1$ day'' means that results are not available for the state-of-the-art method \ie~computation did not finish within 24 hours.}
\label{tab:expt-graph-class1}
\resizebox{\textwidth}{!}{
\begin{tabular}{|l|c|c|c|c|c|c|}
\hline
Kernel & \ \ \ \ MUTAG \ \ \ \ & \ \ \ \ PTC \ \ \ \ & \ \ \ \ ENZYMES \ \ \ \ & \ \ \ \ D \& D \ \ \ \ & \ \ \ \ NCI1 \ \ \ \ & \ \ \ \ NCI109 \ \ \ \ \\
\hline
RW~\cite{Vishwanathan2010} & $71.89\pm 0.66$ ($0.23$) & $55.44\pm 0.15$ ($0.46$) & $14.97\pm 0.28$ ($1.08$) & $>1$ day & $>1$ day & $>1$ day \\
\hline
SP~\cite{Borgwardt2005} & $81.28\pm 0.45$ ($0.13$) & $55.44\pm 0.61$ ($0.45$) & $27.53\pm 0.29$ ($0.50$) & $75.78\pm 0.12$ ($1.55$) & $73.61\pm 0.36$ ($0.07$) & $73.23\pm 0.26$ ($0.07$)\\
\hline
GK~\cite{Shervashidze2009} & $83.50\pm 0.60$ ($2.32$) & $59.65\pm 0.31$ ($167.84$) & $30.64\pm 0.26$ ($122.61$) & $75.90\pm 0.10$ ($8.40$) & $56.56\pm 0.98$ ($0.49$) & $62.00\pm 0.87$ ($0.48$)\\
\hline
MLG~\cite{Kondor2016} & $87.94\pm 1.61$ ($1.86$) & $63.26\pm 1.48$ ($2.36$) & $35.52\pm 0.45$ ($2.56$) & $76.34\pm 0.72$ ($166.45$) & $81.75\pm 0.24$ ($2.42$) & $81.31\pm 0.22$ ($2.45$)\\
\hline
SGE ($t=3, \epsilon = 0.1, \delta = 0.1$) & $71.67\pm 0.86$ ($0.27$) & $53.53\pm 0.04$ ($0.29$) & $24.17\pm 0.54$ ($0.30$) & $60.00\pm 0.01$ ($0.29$) & $72.60\pm 0.31$ ($0.31$) & $71.66\pm 0.25$ ($0.28$) \\
SGE ($t=3, \epsilon = 0.1, \delta = 0.05$) & $75.56\pm 0.52$ ($0.39$) & $53.53\pm 0.76$ ($0.41$) & $25.33\pm 0.75$ ($0.40$) & $60.42\pm 0.23$ ($0.41$) & $74.59\pm 0.75$ ($0.39$) & $74.66\pm 0.67$ ($0.42$) \\
SGE ($t=3, \epsilon = 0.05, \delta = 0.1$) & $86.11\pm 0.00$ ($0.91$) & $54.12\pm 0.48$ ($0.89$) & $29.17\pm 0.03$ ($0.90$) & $63.39\pm 0.58$ ($0.91$) & $76.15\pm 0.72$ ($0.89$) & $74.90\pm 0.62$ ($0.91$) \\
SGE ($t=3, \epsilon = 0.05, \delta = 0.05$) & $84.44\pm 0.74$ ($1.02$) & $55.88\pm 0.67$ ($1.03$) & $29.17\pm 0.10$ ($1.02$) & $64.07\pm 0.99$ ($1.03$) & $76.15\pm 0.24$ ($1.02$) & $76.21\pm 0.82$ ($1.05$)\\
\hline
SGE ($t=4, \epsilon = 0.1, \delta = 0.1$) & $77.78\pm 0.41$ ($1.16$) & $55.59\pm 0.27$ ($1.17$) & $24.00\pm 0.92$ ($1.16$) & $59.83\pm 0.23$ ($1.18$) & $76.05\pm 0.61$ ($1.17$) & $78.05\pm 0.22$ ($1.15$) \\
SGE ($t=4, \epsilon = 0.1, \delta = 0.05$) & $78.89\pm 0.41$ ($1.24$) & $60.29\pm 0.39$ ($1.27$) & $26.00\pm 0.26$ ($1.22$) & $59.92\pm 0.88$ ($1.24$) & $75.86\pm 0.65$ ($1.25$) & $76.55\pm 0.41$ ($1.26$) \\
SGE ($t=4, \epsilon = 0.05, \delta = 0.1$) & $82.22\pm 0.31$ ($1.82$) & $61.18\pm 0.17$ ($1.85$) & $30.67\pm 0.85$ ($1.83$) & $64.41\pm 0.59$ ($1.84$) & $77.71\pm 0.91$ ($1.85$) & $78.82\pm 0.60$ ($1.86$) \\
SGE ($t=4, \epsilon = 0.05, \delta = 0.05$) & $81.67\pm 0.89$ ($1.93$) & $\mathbf{63.53\pm 0.23}$ ($1.95$) & $30.17\pm 0.72$ ($1.94$) & $64.32\pm 0.24$ ($1.96$) & $77.37\pm 0.67$ ($1.94$) & $78.48\pm 0.80$ ($1.97$) \\
\hline
SGE ($t=5, \epsilon = 0.1, \delta = 0.1$) & $86.11\pm 0.05$ ($2.39$) & $56.18\pm 0.26$ ($2.37$) & $30.50\pm 0.43$ ($2.35$) & $65.76\pm 0.60$ ($2.37$) & $78.49\pm 0.49$ ($2.35$) & $79.89\pm 0.33$ ($2.36$) \\
SGE ($t=5, \epsilon = 0.1, \delta = 0.05$) & $86.11\pm 0.05$ ($2.50$) & $54.71\pm 0.23$ ($2.49$) & $30.17\pm 0.46$ ($2.48$) & $65.68\pm 0.84$ ($2.47$) & $79.51\pm 0.67$ ($2.48$) & $79.74\pm 0.23$ ($2.50$) \\
SGE ($t=5, \epsilon = 0.05, \delta = 0.1$) & $85.56\pm 0.52$ ($2.79$) & $62.06\pm 0.90$ ($2.73$) & $32.17\pm 0.27$ ($2.75$) & $68.90\pm 0.22$ ($2.76$) & $81.26\pm 0.13$ ($2.78$) & $79.02\pm 0.80$ ($2.77$) \\
SGE ($t=5, \epsilon = 0.05, \delta = 0.05$) & $85.00\pm 0.89$ ($2.85$) & $62.06\pm 0.79$ ($2.89$) & $31.17\pm 0.85$ ($2.86$) & $68.64\pm 0.81$ ($2.88$) & $81.75\pm 0.29$ ($2.84$) & $79.89\pm 0.85$ ($2.87$) \\
\hline
SGE ($t=6, \epsilon = 0.1, \delta = 0.1$) & $87.78\pm 0.31$ ($2.68$) & $59.41\pm 0.06$ ($2.71$) & $28.67\pm 0.22$ ($2.72$) & $68.98\pm 0.90$ ($2.69$) & $81.84\pm 0.84$ ($2.70$) & $80.65\pm 0.29$ ($2.71$) \\
SGE ($t=6, \epsilon = 0.1, \delta = 0.05$) & $88.33\pm 0.15$ ($2.83$) & $61.47\pm 0.52$ ($2.84$) & $28.50\pm 0.66$ ($2.86$) & $70.08\pm 0.48$ ($2.83$) & $81.70\pm 0.94$ ($2.85$) & $80.94\pm 0.92$ ($2.87$) \\
SGE ($t=6, \epsilon = 0.05, \delta = 0.1$) & $88.89\pm 0.70$ ($3.05$) & $57.65\pm 0.58$ ($3.06$) & $36.33\pm 0.28$ ($3.07$) & $72.63\pm 0.37$ ($3.07$) & $82.40\pm 0.88$ ($3.05$) & $81.22\pm 0.54$ ($3.04$) \\
SGE ($t=6, \epsilon = 0.05, \delta = 0.05$) & $\mathbf{89.75\pm 0.24}$ ($3.29$) & $55.59\pm 0.96$ ($3.31$) & $35.17\pm 0.26$ ($3.28$) & $73.05\pm 0.64$ ($3.30$) & $82.48\pm 0.87$ ($3.30$) & $81.25\pm 0.56$ ($3.32$) \\
\hline
SGE ($t=7, \epsilon = 0.1, \delta = 0.1$) & $85.56\pm 0.68$ ($3.16$) & $58.53\pm 0.99$ ($3.15$) & $37.33\pm 0.46$ ($3.14$) & $72.54\pm 0.66$ ($3.13$) & $81.13\pm 0.74$ ($3.17$) & $81.38\pm 0.80$ ($3.15$) \\
SGE ($t=7, \epsilon = 0.1, \delta = 0.05$) & $86.11\pm 0.93$ ($3.34$) & $57.06\pm 0.82$ ($3.32$) & $36.67\pm 0.85$ ($3.33$) & $72.80\pm 0.41$ ($3.35$) & $82.03\pm 0.55$ ($3.36$) & $81.22\pm 0.15$ ($3.37$) \\
SGE ($t=7, \epsilon = 0.05, \delta = 0.1$) & $86.67\pm 0.37$ ($5.39$) & $59.12\pm 0.26$ ($5.37$) & $40.00\pm 0.50$ ($5.38$) & $76.08\pm 0.33$ ($5.37$) & $\mathbf{82.49\pm 0.91}$ ($5.35$) & $\mathbf{82.62\pm 0.42}$ ($5.36$) \\
SGE ($t=7, \epsilon = 0.05, \delta = 0.05$) & $87.22\pm 0.27$ ($5.62$) & $60.00\pm 0.99$ ($5.61$) & $\mathbf{40.67\pm 0.40}$ ($5.60$) & $\mathbf{76.58\pm 0.27}$ ($5.63$) & $82.10\pm 1.04$ ($5.62$) & $82.32\pm 0.65$ ($5.64$) \\
\hline
\end{tabular}}
\end{center}
\end{table*}

\begin{figure*}
\centering
\includegraphics[width=\textwidth]{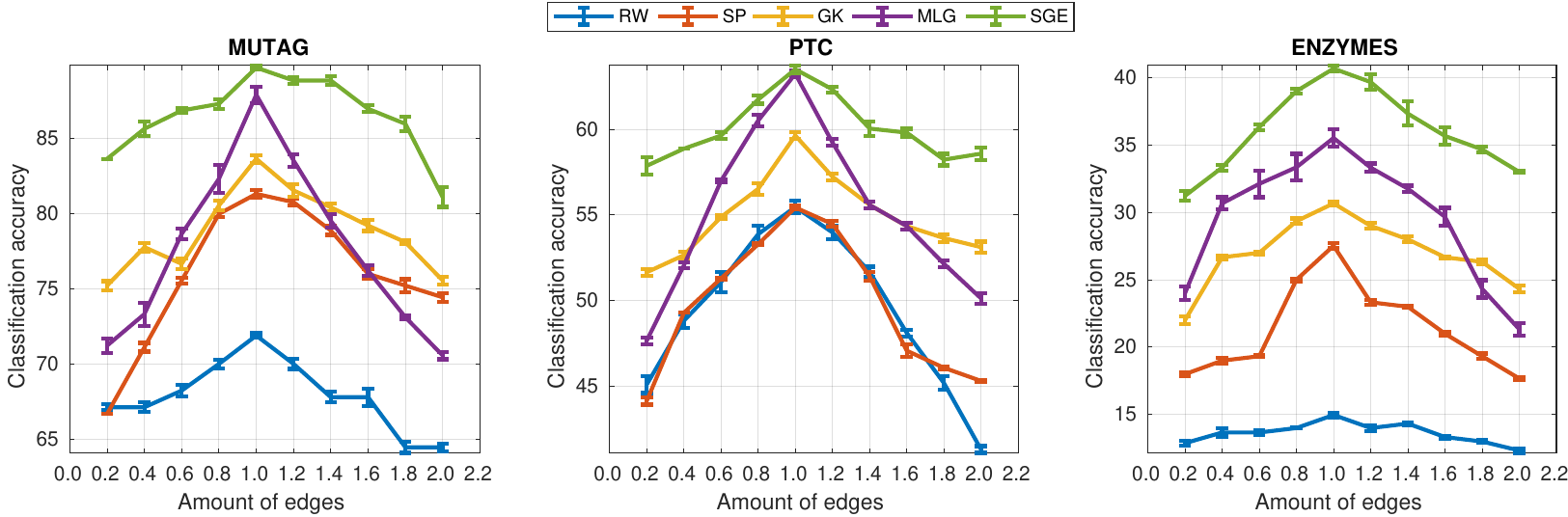}
\caption{Plot of classification accuracy versus amount of edges on MUTAG, PTC and ENZYMES datasets with our proposed stochastic graphlet embedding and other state-of-the-art methods. RW corresponds to the random walk kernel~\cite{Vishwanathan2010}, SP stands for shortest path kernel~\cite{Borgwardt2005}, GK corresponds to the standard graphlet kernel~\cite{Shervashidze2009}, MLG stands for multiscale Laplacian graph kernel~\cite{Kondor2016}, and SGE refers to our proposed stochastic graphlet embedding.}
\label{fig:expt-noisy-graphs}
\end{figure*}

\subsection{MUTAG, PTC, ENZYMES, D\&D, NCI1 and NCI109}
\label{sec:results:unlabeled}
In this section, we show the impact of our proposed stochastic graphlet embedding on the performance of graph classification using six publicly available graph databases with unlabeled nodes: \emph{MUTAG}, \emph{PTC}, \emph{ENZYMES}, \emph{D\&D}, {\emph{NCI1} and \emph{NCI109}}. The MUTAG dataset contains graphs representing $188$ chemical compounds which are either mutagenic or not. So here the task of the classifier is to predict the mutagenicity of the chemical compounds, which is a two class problem. The PTC (Predictive Toxicology Challenge) dataset consists of graphs of $344$ chemical compounds known to cause (or not) cancer in rats and mice. Hence the task of the classifier is to predict the cancerogenicity of the chemical compounds, which is also a two class problem. The ENZYMES dataset contains graphs representing protein tertiary structures consisting of $600$ enzymes from the BRENDA enzyme. Here the task is to correctly assign each enzyme to one of the $6$ EC top levels. The D\&D dataset consists of $1178$ graphs of protein structures which are either enzyme or non-enzyme. Therefore, the task of the classifier is to predict if a protein is enzyme or not, which is essentially a two class problem. {The NCI1 and NCI109 represent two balanced subsets of chemical compounds screened for activity against non-small cell lung cancer and ovarian cancer cell lines, respectively. These two datasets respectively contain $4110$ and $4127$ graphs of chemical compounds which are either active or inactive against the respective cancer cells. Hence, the goal of the classifier is to judge the activeness of the chemical compounds, which is a two class problem.} Details on the above six datasets are shown in~\tab{tab:det-graph-datasets1}.

In order to achieve graph classification, we use the \emph{histogram intersection} kernel~\cite{Barla2003} on top of our stochastic graphlet embedding, and we plug it into SVMs for training and classification. In these experiments, we report the average classification accuracies and their respective standard deviations in~\tab{tab:expt-graph-class1} using 10--fold cross validation. We also show comparison against state-of-the-art graph kernels including (i) the standard random-walk kernel (RW)~\cite{Vishwanathan2010}, that counts common random walks in two graphs, (ii) the shortest path kernel (SP)~\cite{Borgwardt2005}, that compares shortest path lengths in two graphs, (iii) the graphlet kernel (GK)~\cite{Shervashidze2009}, that compares graphlets with up to $5$ nodes, and (iv) the multiscale Laplacian graph (MLG) kernel~\cite{Kondor2016}, that takes into account the structure at different scale ranges. In these comparative methods, we use the parameters that provide overall the best performances; precisely, the discounting factor $\lambda$ of RW is set to $0.001$ and the maximum number of nodes in GK is equal to $5$ while for MLG, the underlying parameters (namely the regularization coefficient, the radius of the used neighborhood and the number of levels in MLG) are set to $0.01$, $2$ and $3$ respectively. \tab{tab:expt-graph-class1} shows the impact of our proposed stochastic graphlet embedding for different pairs of $\epsilon$ and $\delta$ with increasing order graphlets (the underlying $M$ is shown in \tab{tab:sample-no} for different pairs of $\epsilon$ and $\delta$).

Compared to all these methods, our stochastic graphlet embedding achieves the best performances on all the six datasets, and this clearly shows the positive impact of high-order graphlets w.r.t low-order ones (as also supported in~\cite{Shervashidze2009}), though a few exceptions exist; for instance, on the PTC dataset, the accuracy stabilizes and reaches its highest value with only 4 order graphlets. In all these results, we also observe that increasing the number of samples ($M$) impacts -- at some extent -- the classification accuracy; indeed, more samples make the estimated graphlet distribution close to the actual one (as also corroborated through further extensive experiments in~\cite{Dutta2018SGEsupp}, with much larger values of $M$ and $T$).

We further push experiments and study the resilience of our graph representation against inter and intra-class graph structure variations; for that purpose, we artificially disrupt graphs in MUTAG, PTC and ENZYMES datasets. This disruption process is random and consists in adding/deleting edges from each original graph $G=(V,E)$. More precisely, we derive multiples graph instances (whose edge set cardinality is equal to $\tau \lvert E\rvert$) either by deleting $(1-\tau) \lvert E\rvert$ edges from $G$ (with $\tau \in \{0.2,0.4,0.6,0.8\}$) or by adding $(\tau-1)\lvert E\rvert$ extra edges into $G$ (with $\tau \in \{1.2,1.4,1.6,1.8,2\}$). For each setting of $\tau$, we apply the proposed SGE along with the other state-of-the-art methods -- random walk kernel~\cite{Vishwanathan2010} (RW), shortest path kernel~\cite{Borgwardt2005} (SP), graphlet kernel~\cite{Shervashidze2009} (GK), and multiscale Laplacian graph kernel~\cite{Kondor2016} (MLG) -- and we plug the resulting kernels into SVM for classification. \fig{fig:expt-noisy-graphs} shows the evolution of the classification accuracy with respect to different setting of $\tau$ (also referred to as "amount of edges" in that figure). From these results, we observe that adding or deleting edges naturally harms the classification accuracies of all the methods especially MLG on MUTAG/PTC and RW on PTC and this clearly shows their high sensitivity; specifically, MLG depends on a base kernel defined on graph vertices so deleting edges (possibly along with their nodes) hampers the accuracy. As for RW, deleting (resp. adding) edges reduces (resp. increases) the number of common walks between graphs and thereby affects the relevance of their kernel similarity resulting into a drop in performances. In contrast, our SGE method and the standard graphlet kernel, are relatively more resilient to these graph structure variations.

Finally, we observe that the overall performances of all the methods (including ours) on the ENZYMES dataset are relatively low compared to the other databases. This may result from the relatively large number of classes which cannot be easily distinguished using only the structure of those graphs (without labels/attributes on their nodes, etc.). In order to better establish this fact, we will show, in section~\ref{sec:withlabels}, extra experiments while considering labeled/attributed graphs.

\subsection{COIL, GREC, AIDS, MAO and ENZYMES}\label{sec:withlabels}
We consider five different datasets (see~\tab{tab:det-graph-datasets2}) modeled with graphs whose nodes are {\it now} labeled; three of them \viz~\emph{COIL}, \emph{GREC} and \emph{AIDS} are taken from the IAM graph database repository\footnote{Available at \url{http://www.fki.inf.unibe.ch/databases/iam-graph-database}}~\cite{Riesen2008}, the fourth one \ie~\emph{MAO} is taken from the GREYC Chemistry graph dataset collection\footnote{Available at \url{https://brunl01.users.greyc.fr/CHEMISTRY/}}. The fifth one is the ENZYMES dataset mentioned earlier in~\sect{sec:results:unlabeled}, with the only difference being node and edge attributes which are now used in our experiments. The COIL database includes $3900$ graphs belonging to $100$ different classes with $39$ instances per class; each instance has a different rotation angle. The GREC dataset consists of $1100$ graphs representing $22$ different classes (characterizing architectural and electronic symbols) with $50$ instances per class; these instances have different noise levels. The AIDS database consists of $2000$ graphs representing molecular compounds which are constructed from the AIDS Antiviral Screen Database of Active Compounds\footnote{See at \url{http://dtp.nci.nih.gov/docs/aids/aids_data.html}}. This dataset consists of two classes \viz~active ($400$ elements) and inactive ($1600$ elements), which respectively represent molecules with possible activity against HIV. The MAO dataset includes $68$ graphs representing molecules that either inhibit (or not) the monoamine oxidase (an antidepressant drug with $38$ molecules). In all these datasets the task is again to infer the membership of a given test instance among two or multiple classes. 

\begin{table}[!htbp]
\begin{center}
\caption{Available details on COIL, GREC, AIDS, MAO and ENZYMES (labeled) graph datasets.}
\label{tab:det-graph-datasets2}
\resizebox{\columnwidth}{!}{
\begin{tabular}{|l|r|l|r|r|p{2cm}|p{2cm}|}
\hline
Datasets & \#Graphs & Classes & Avg. \#nodes & Avg. \#edges & Node labels & Edge labels\\\hline
COIL & $3900$ & $100$ ($39$ each) & $21.5$ & $54.2$ & NA & Valency of bonds\\
GREC & $1100$ & $22$ ($50$ each) & $11.5$ & $11.9$ & Type of joint: corner, intersection, etc. & Type of edge: line or curve.\\
AIDS & $2000$ & $2$ ($1600$ vs. $400$) & $15.7$ & $16.2$ & Label of atoms & Valency of bonds\\
MAO & $68$ & $2$ ($38$ vs. $30$) & $18.4$ & $19.6$ & Label of atoms & Valency of bonds\\
ENZYMES & $600$ & $6$ ($100$ each) & $32.6$ & $124.3$ & $-$ & $-$\\\hline
\end{tabular}}
\end{center}
\end{table}

\begin{table}[!htbp]
\begin{center}
\caption{Classification accuracies (in \%) obtained by our proposed stochastic graphlet embedding (SGE) on COIL, GREC, AIDS and MAO datasets and comparison with state-of-the-art methods \viz random walk kernel (RW)~\cite{Vishwanathan2010}, dissimilarity embedding (DE)~\cite{Bunke2012}, node attribute statistics (NAS)~\cite{Gibert2012} and multiscale Laplacian graph kernel (MLG)~\cite{Kondor2016}. {The average processing time for generating the embedding of a given graph is indicated within the parenthesis just after each accuracy result.}}
\label{tab:expt-graph-class2}
\resizebox{\columnwidth}{!}{
\begin{tabular}{|l|c|c|c|c|c|}
\hline
Method & COIL & GREC & AIDS & MAO & ENZYMES (labeled) \\
\hline
RW~\cite{Vishwanathan2010} & $94.2$ ($2.23$) & $96.2$ ($1.67$) & $98.5$ ($1.89$) & $82.4$ ($2.01$) & $28.17\pm 0.76$ ($3.14$)\\
\hline
DE~\cite{Bunke2010} & $96.8$ & $95.1$ & $98.1$ & $91.2$ & $-$ \\
\hline
NAS~\cite{Gibert2012} & $98.1$ & $99.2$ & $98.3$ & $81.7$ & $-$ \\
\hline
MLG~\cite{Kondor2016} & $97.3$ ($3.14$) & $96.3$ ($1.67$) & $94.7$ ($1.89$) & $89.2$ ($2.01$) & $61.81\pm 0.99$ ($3.16$) \\
\hline
SGE ($t=1, \epsilon = 0.1, \delta = 0.1$) & $89.60$ ($0.43$) & $98.67$ ($0.40$) & $95.45$ ($0.42$) & $82.35$ ($0.46$) & $31.67\pm 0.89$ ($0.45$) \\
SGE ($t=1, \epsilon = 0.1, \delta = 0.05$) & $90.60$ ($0.54$) & $99.05$ ($0.52$) & $94.56$ ($0.51$) & $82.35$ ($0.51$) & $33.33\pm 0.39$ ($0.53$) \\
SGE ($t=1, \epsilon = 0.05, \delta = 0.1$) & $92.40$ ($0.85$) & $99.43$ ($0.84$) & $94.54$ ($0.81$) & $85.29$ ($0.80$) & $34.00\pm 0.56$ ($0.86$) \\
SGE ($t=1, \epsilon = 0.05, \delta = 0.05$) & $93.90$ ($1.02$) & $99.43$ ($1.06$) & $95.87$ ($1.05$) & $88.24$ ($1.04$) & $35.33\pm 0.26$ ($1.05$) \\
\hline
SGE ($t=2, \epsilon = 0.1, \delta = 0.1$) & $91.50$ ($0.51$) & $99.24$ ($0.53$) & $95.54$ ($0.49$) & $85.29$ ($0.55$) & $37.00\pm 0.81$ ($0.52$) \\
SGE ($t=2, \epsilon = 0.1, \delta = 0.05$) & $92.40$ ($0.67$) & $99.24$ ($0.62$) & $96.87$ ($0.66$) & $85.29$ ($0.68$) & $38.33\pm 0.74$ ($0.69$) \\
SGE ($t=2, \epsilon = 0.05, \delta = 0.1$) & $93.90$ ($1.04$) & $99.43$ ($1.07$) & $97.76$ ($1.05$) & $85.29$ ($1.02$) & $39.67\pm 0.05$ ($1.03$) \\
SGE ($t=2, \epsilon = 0.05, \delta = 0.05$) & $94.40$ ($1.21$) & $99.43$ ($1.23$) & $97.87$ ($1.24$) & $88.24$ ($1.22$) & $38.00\pm 0.89$ ($1.22$) \\
\hline
SGE ($t=3, \epsilon = 0.1, \delta = 0.1$) & $91.80$ ($0.68$) & $99.43$ ($0.67$) & $97.51$ ($0.64$) & $88.24$ ($0.69$) & $47.33\pm 0.30$ ($0.67$) \\
SGE ($t=3, \epsilon = 0.1, \delta = 0.05$) & $93.70$ ($0.84$) & $99.24$ ($0.82$) & $98.01$ ($0.83$) & $85.29$ ($0.80$) & $45.00\pm 0.62$ ($0.82$) \\
SGE ($t=3, \epsilon = 0.05, \delta = 0.1$) & $94.70$ ($1.25$) & $99.43$ ($1.22$) & $97.98$ ($1.26$) & $85.29$ ($1.28$) & $53.33\pm 0.97$ ($1.26$) \\
SGE ($t=3, \epsilon = 0.05, \delta = 0.05$) & $95.90$ ($1.43$) & $99.43$ ($1.41$) & $97.88$ ($1.38$) & $91.18$ ($1.42$) & $51.00\pm 0.67$ ($1.45$) \\
\hline
SGE ($t=4, \epsilon = 0.1, \delta = 0.1$) & $93.50$ ($1.81$) & $99.24$ ($1.83$) & $97.98$ ($1.78$) & $88.24$ ($1.79$) & $45.33\pm 0.93$ ($1.82$) \\
SGE ($t=4, \epsilon = 0.1, \delta = 0.05$) & $94.70$ ($1.98$) & $99.43$ ($1.97$) & $98.18$ ($1.93$) & $91.18$ ($1.96$) & $45.00\pm 0.62$ ($2.02$) \\
SGE ($t=4, \epsilon = 0.05, \delta = 0.1$) & $95.80$ ($2.24$) & $99.43$ ($2.26$) & $98.32$ ($2.22$) & $91.18$ ($2.20$) & $56.00\pm 0.40$ ($2.25$) \\
SGE ($t=4, \epsilon = 0.05, \delta = 0.05$) & $96.50$ ($2.42$) & $99.24$ ($2.43$) & $98.16$ ($2.44$) & $94.12$ ($2.37$) & $54.67\pm 0.52$ ($2.42$) \\
\hline
SGE ($t=5, \epsilon = 0.1, \delta = 0.1$) & $94.90$ ($2.74$) & $99.05$ ($2.71$) & $98.76$ ($2.76$) & $91.18$ ($2.77$) & $56.33\pm 0.52$ ($2.76$) \\
SGE ($t=5, \epsilon = 0.1, \delta = 0.05$) & $95.50$ ($2.91$) & $99.05$ ($2.93$) & $98.82$ ($2.92$) & $91.18$ ($2.94$) & $54.00\pm 0.73$ ($2.93$) \\
SGE ($t=5, \epsilon = 0.05, \delta = 0.1$) & $97.90$ ($3.29$) & $99.43$ ($3.31$) & $\mathbf{99.12}$ ($3.32$) & $94.12$ ($3.34$) & $60.33\pm 0.45$ ($3.27$) \\
SGE ($t=5, \epsilon = 0.05, \delta = 0.05$) & $\mathbf{98.86}$ ($3.43$) & $\mathbf{99.62}$ ($3.39$) & $98.92$ ($3.41$) & $\mathbf{97.06}$ ($3.46$) & $\mathbf{62.33\pm 0.14}$ ($3.42$) \\
\hline
\end{tabular}}
\end{center}
\end{table}

Similarly to the previous experiments, we use the histogram intersection kernel~\cite{Barla2003} on top of SGE and we plug it into SVM for learning and graph classification. In order to measure the accuracy of our method (reported in \tab{tab:expt-graph-class2}), we use the available splits of COIL, GREC and AIDS into training, validation and test sets; for MAO, we consider instead the leave-one-out error split. Note that these splits correspond to the ones used by most of the related state-of-the-art methods. These related methods also include dissimilarity embedding (DE) with a prototype set of cardinality $100$ and node attribute statistics (NAS) based on fuzzy $k$-means and soft edge assignment. \tab{tab:expt-graph-class2} shows the performance of our proposed stochastic graphlet embedding on these datasets for different graphlet orders (and pairs of $\epsilon$, $\delta$) and its comparison against the related work. Similarly to the previous section, we globally observe an influencing positive impact of high-order graphlets on performances. We also observe a gain in performances as $M$ (the number of samples) increases. These results clearly show that our proposed method outperforms the related state-of-the-art on COIL and MAO while on GREC and AIDS, it performs comparably and utterly well.

\subsection{AMA Dental Forms}
\begin{figure}[!htbp]
\centering
\resizebox{\columnwidth}{!}{
\begin{tabular}{cc}
\includegraphics[width=0.48\textwidth]{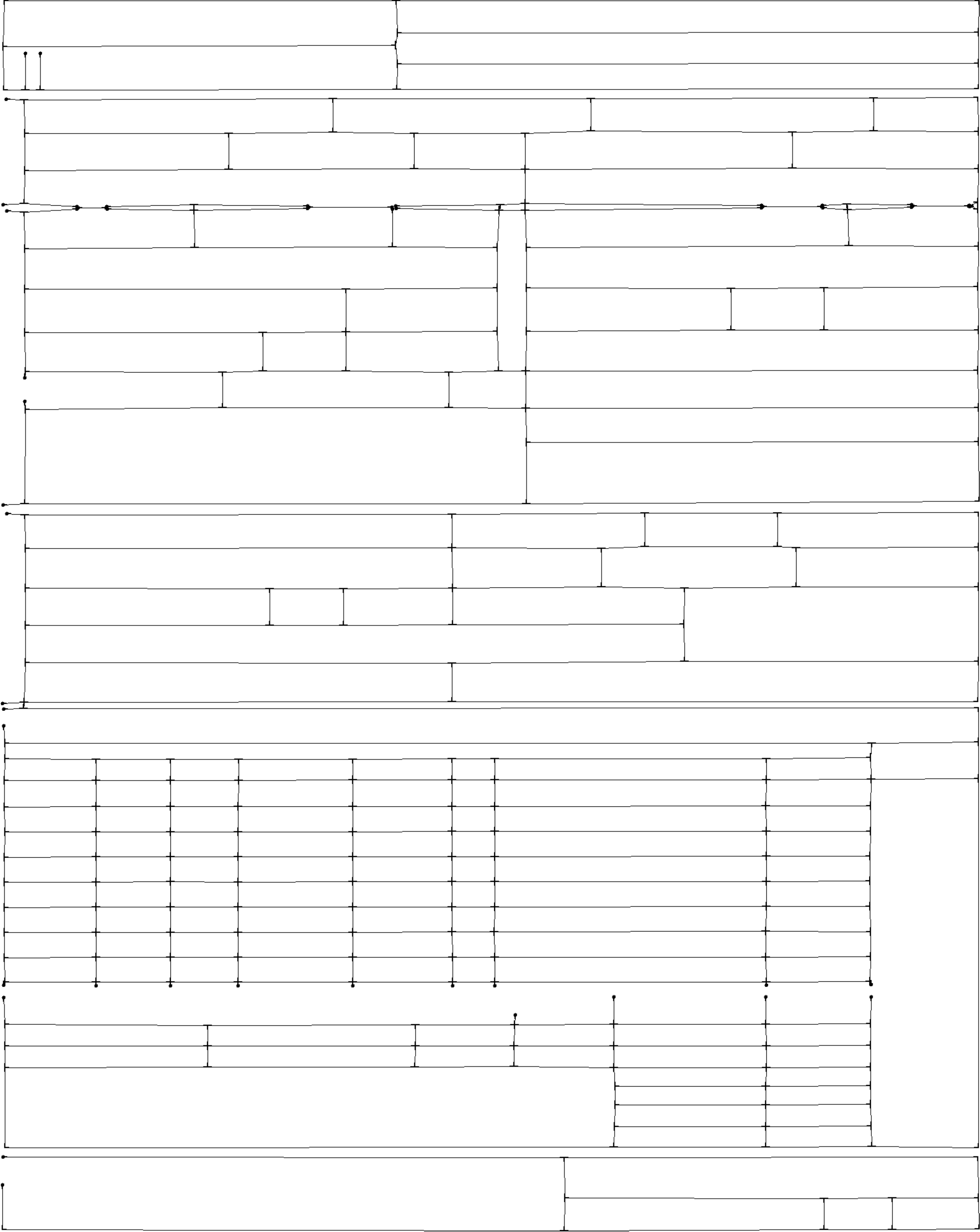} & \includegraphics[width=0.48\textwidth]{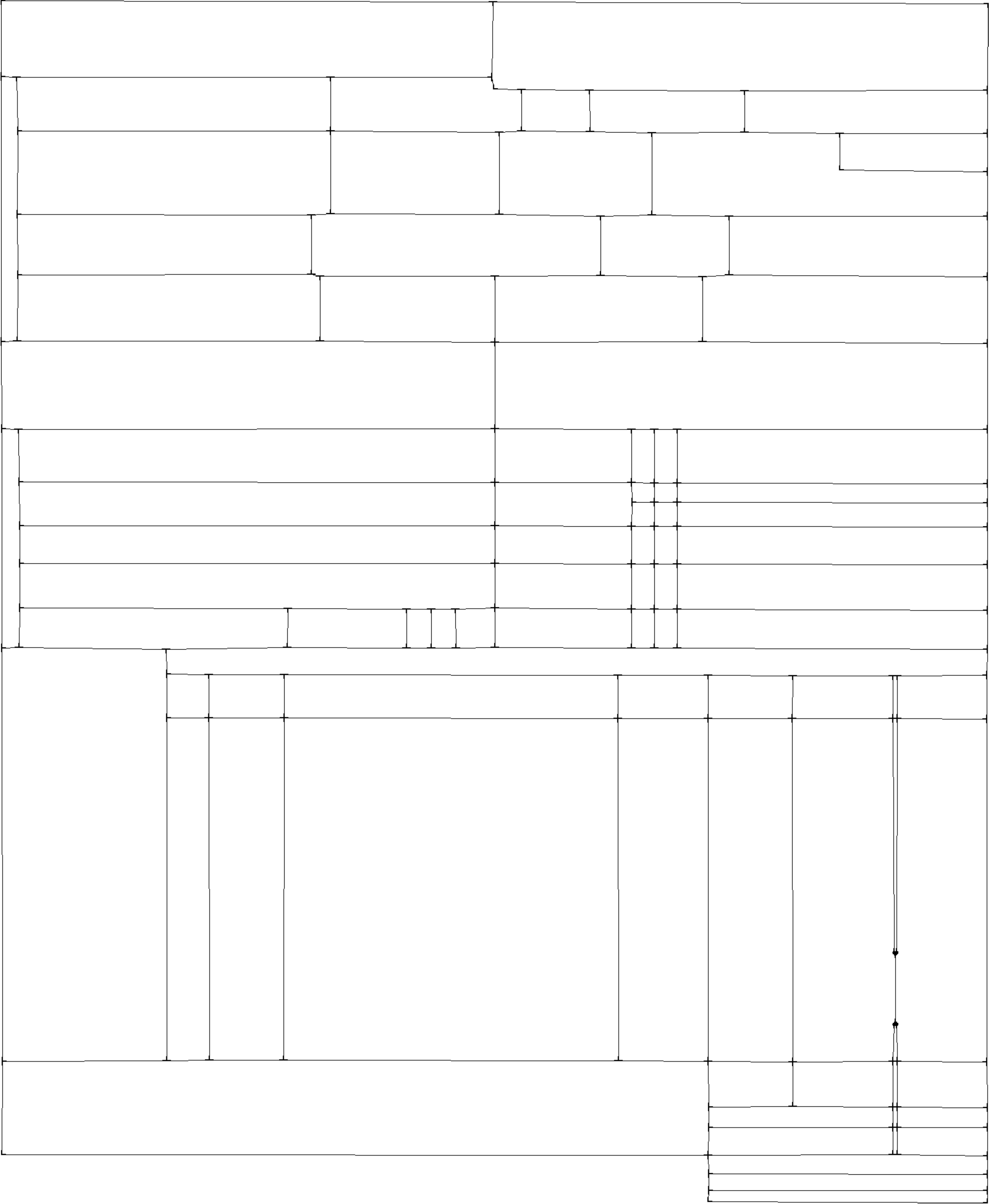}\\
\multicolumn{2}{c}{} \\
\Large{(a) FDent013} & \Large{(b) FDent097}\\
\multicolumn{2}{c}{} \\
\includegraphics[width=0.48\textwidth]{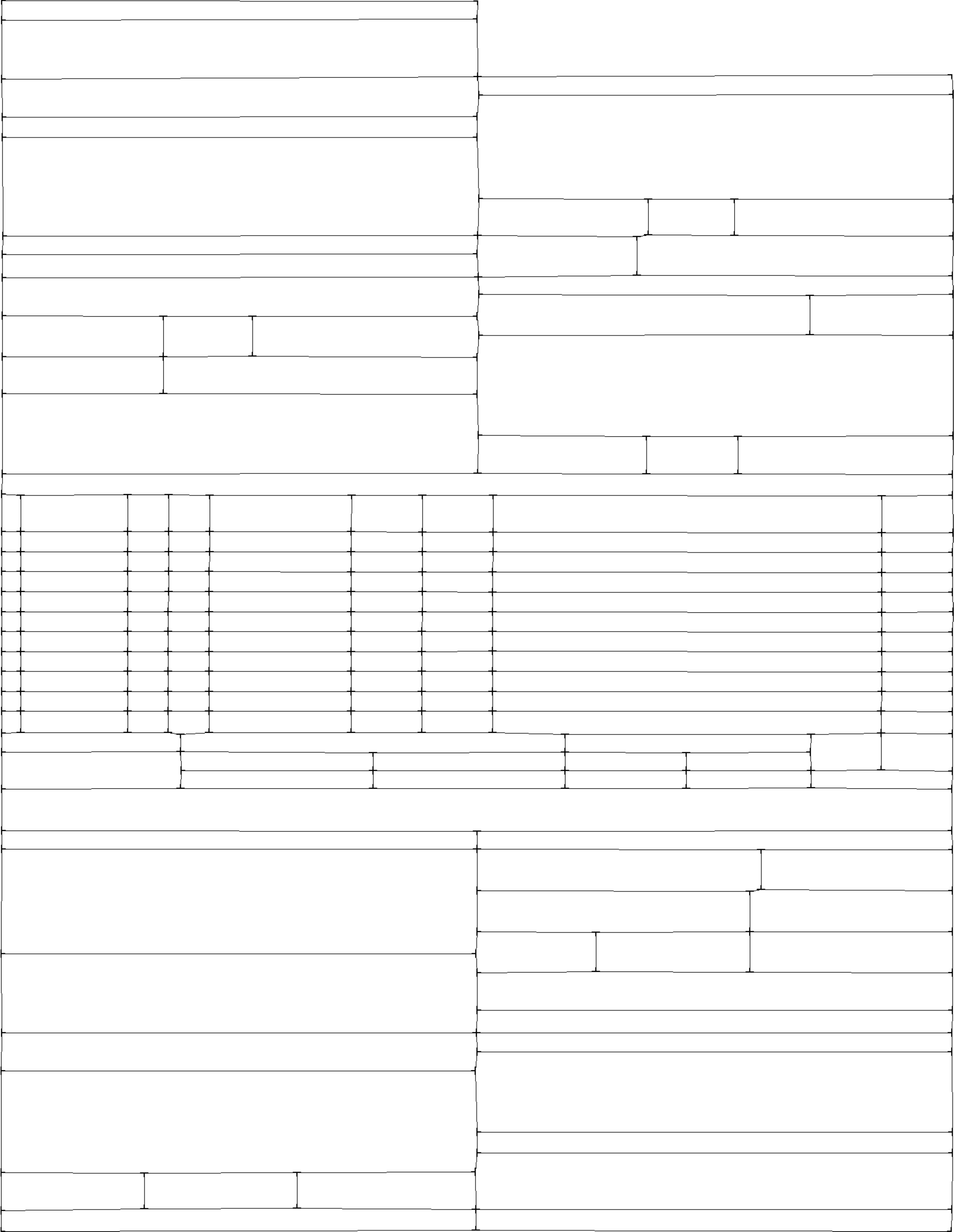} & \includegraphics[width=0.48\textwidth]{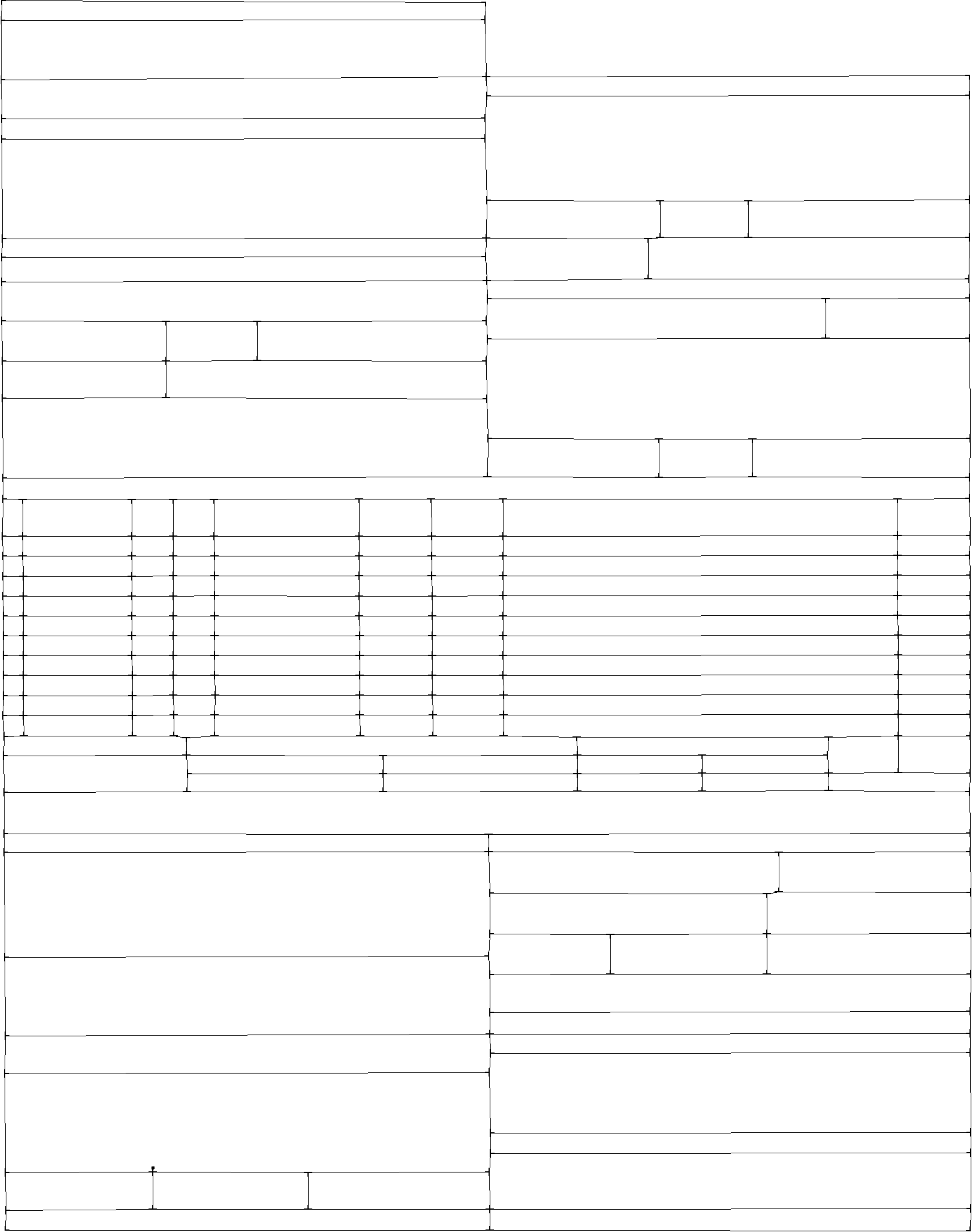}\\
\multicolumn{2}{c}{} \\
\Large{(c) FDent102} & \Large{(d) 100721104848}\\
\end{tabular}}
\caption{Examples of American Medical Association (AMA) dental claim forms documents. Among the above `FDent013', `FDent097' and `FDent102' are the three different categories, which are obtained by digitizing and removing the textual parts from the respective blank form templates and `100721104848' is a dental claim form encountered in a production document processing application, which is obtained by digitizing and removing the textual parts from it. This particular form belongs to the same class as of `FDent102'. (Best viewed in pdf).}
\label{fig:example-AMA}
\end{figure}

Inspired by the same protocol as~\cite{Saund2013}, we apply our method to form document indexing and retrieval on the publicly available benchmark\footnote{See \url{www2.parc.com/isl/groups/pda/data/DentalFormsLineArtDataSet.zip}} used in ~\cite{Saund2013}; the latter is closely related to our framework. Indeed, it also seeks to describe data by measuring the distribution of their subgraphs. Therefore we consider this benchmark and the related work in~\cite{Saund2013} in order to evaluate and compare the performance of our method. The main goal of this benchmark is to index and retrieve form documents that have sparse and inconsistent textual content (due to the variability in filling the fields of these documents). These forms usually contain networks of rectilinear rule lines serving as region separators, data field locators, and field group indicators (see \fig{fig:example-AMA}).
 
The dataset used for this experiment is basically a collection of $6247$ American Medical Association (AMA) dental claim forms encountered in a production document processing application. This dataset also includes $208$ blank forms which serve as ground-truth categories, so the task is to assign each of these forms to one of the $208$ categories. In these forms the rectilinear lines intersect each other in well defined ways that form junction and also free end terminator, which essentially serve as the graph nodes and their connections as the graph edges. There are only $13$ node labels depending on the junction type (refer to~\cite{Saund2013} for more details) and only two edge labels: vertical and horizontal.

We follow the same protocol, as~\cite{Saund2013}, in order to evaluate and compare the performances of our method. This protocol consists in comparing the ranking of category model matches to the document image graphs between the classifier output and the ground-truth. Let $r_{g,c}$ be the ranking assigned by a classifier to the model with the top ranking in the ground-truth and let $r_{c,g}$ be the ranking in the ground-truth of the model assigned top ranking by the classifier. Then, the performance of our method is measured by 
\begin{equation}
\rho = \frac{1}{2}\Big(\frac{1}{r_{c,g}}+\frac{1}{r_{g,c}}\Big), 
\end{equation}
here a maximum score $\rho=1$ is given only when the top ranking categories assigned by the classifier and the ground-truth agree. Some credit is also given when the top ranking category (of the ground truth or classifier output) score highly in the complement rankings. For more details on this performance measure, we refer to~\cite{Saund2013}.

\begin{table}[!htbp]
\begin{center}
\caption{Performance measure $\rho$ obtained by our method (SGE) for retrieving the AMA dental forms documents into $208$ model categories and comparison with the method proposed by Saund~\cite{Saund2013}. It shows the results varying the size of graphlets and their combination. \emph{hist. int. sim.} refers to feature vector comparison using histogram intersection similarity whereas \emph{cosine sim.} refers to feature vector comparison using cosine similarity. \emph{CMD comp.} refers to feature vector comparison using the CMD distance~\cite{Saund2013}. \emph{cos comp.} refers to feature vector comparison using the cosine distance. \emph{Extv. G.L. Level} refers to the size of subgraph in terms of number of nodes. {The average processing time for generating the embedding of a given graph is indicated within the parenthesis after each performance measure.}}
\label{tab:expt-doc-class}
\resizebox{\columnwidth}{!}{
\begin{tabular}{|c|l|c||l|c||c|c|c|}
\hline
\multicolumn{5}{|c||}{SGE} & \multicolumn{3}{c|}{Saund~\cite{Saund2013}}\\\hline
Distance or & & Perf. & & Perf. & & Extv. & Perf. \\
Similarity & Graphlets & Measure & Graphlets & Measure & Test & G.L. & Measure \\
Measure & & $\rho$ & & $\rho$ & Condition & Level & $\rho$ \\\hline
hist. int. sim. & $t=0$ & $0.291$ ($0.24$) & $-$ & $-$ & $-$ & $-$ & $-$\\
hist. int. sim. & $t=1$ & $0.264$ ($1.02$) & $t=\lbrace 0,\dots, 1 \rbrace$ & $0.296$ ($1.15$) & $-$ & $-$ & $-$\\
hist. int. sim. & $t=2$ & $0.336$ ($1.21$) & $t=\lbrace 0,\dots, 2 \rbrace$ & $0.337$ ($1.37$) & $-$ & $-$ & $-$\\
hist. int. sim. & $t=3$ & $0.382$ ($1.43$) & $t=\lbrace 0,\dots, 3 \rbrace$ & $0.390$ ($1.61$) & $-$ & $-$ & $-$\\
hist. int. sim. & $t=4$ & $0.388$ ($2.42$) & $t=\lbrace 0,\dots, 4 \rbrace$ & $0.416$ ($2.71$) & CMD comp. & $\lbrace 1, \dots, 2 \rbrace$ & $0.411$\\
hist. int. sim. & $t=5$ & $0.393$ ($3.43$) & $t=\lbrace 0,\dots, 5 \rbrace$ & $0.435$ ($3.67$) & CMD comp. & $\lbrace 1, \dots, 3 \rbrace$ & $0.467$\\
hist. int. sim. & $t=6$ & $0.452$ ($3.87$) & $t=\lbrace 0,\dots, 6 \rbrace$ & $0.486$ ($4.15$) & CMD comp. & $\lbrace 1, \dots, 4 \rbrace$ & $0.507$\\
hist. int. sim. & $t=7$ & $0.489$ ($6.22$) & $t=\lbrace 0,\dots, 7 \rbrace$ & $\mathbf{0.536}$ ($6.45$) & CMD comp. & $\lbrace 1, \dots, 5 \rbrace$ & $0.524$\\
\hline
cosine sim. & $t=0$ & $0.289$ ($0.23$) & $-$ & $-$ & $-$ & $-$ & $-$\\
cosine sim. & $t=1$ & $0.217$ ($1.04$) & $t=\lbrace 0,\dots, 1 \rbrace$ & $0.293$ ($1.17$) & $-$ & $-$ & $-$\\
cosine sim. & $t=2$ & $0.276$ ($1.24$) & $t=\lbrace 0,\dots, 2 \rbrace$ & $0.304$ ($1.41$) & $-$ & $-$ & $-$\\
cosine sim. & $t=3$ & $0.282$ ($1.41$) & $t=\lbrace 0,\dots, 3 \rbrace$ & $0.316$ ($1.64$) & $-$ & $-$ & $-$\\
cosine sim. & $t=4$ & $0.308$ ($2.46$) & $t=\lbrace 0,\dots, 4 \rbrace$ & $0.328$ ($2.49$) & cosine comp. & $\lbrace 1, \dots, 2 \rbrace$ & $0.341$\\
cosine sim. & $t=5$ & $0.312$ ($3.51$) & $t=\lbrace 0,\dots, 5 \rbrace$ & $0.336$ ($3.53$) & cosine comp. & $\lbrace 1, \dots, 3 \rbrace$ & $0.353$\\
cosine sim. & $t=6$ & $0.323$ ($3.97$) & $t=\lbrace 0,\dots, 6 \rbrace$ & $0.361$ ($3.98$) & cosine comp. & $\lbrace 1, \dots, 4 \rbrace$ & $0.371$\\
cosine sim. & $t=7$ & $0.341$ ($6.27$) & $t=\lbrace 0,\dots, 7 \rbrace$ & $\mathbf{0.382}$ ($6.31$) & cosine comp. & $\lbrace 1, \dots, 5 \rbrace$ & $0.377$\\
\hline
\end{tabular}}
\end{center}
\end{table}

We apply our stochastic graphlet embedding both to the form documents and also to the templates (with $\epsilon=0.05$ and $\delta=0.05$). We consider two different functions that measure the similarity between each pair of document and template embeddings; \viz~histogram intersection{~\cite{Barla2003}} (a.k.a \emph{Common-Minus-Difference}) and \emph{cosine} as also achieved in~\cite{Saund2013}. \tab{tab:expt-doc-class} shows these measures obtained by our stochastic graphlet embedding using graphlets with different fixed orders taken separately and combined; again, $t=0$ corresponds to singleton graphlets \ie~only nodes. As observed previously, high order graphlets have more influencing positive impact on performances. Furthermore, mixing graphlets with different orders is highly beneficial and makes it possible to overtake the related work~\cite{Saund2013}.

\subsection{MNIST Database}
\begin{figure}[!htbp]
\centering
\resizebox{\columnwidth}{!}{
\begin{tabular}{cccccccc}
\includegraphics[width=0.05\textwidth]{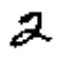}&
\includegraphics[width=0.05\textwidth]{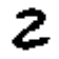}& 
\includegraphics[width=0.05\textwidth]{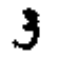}&
\includegraphics[width=0.05\textwidth]{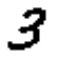}&
\includegraphics[width=0.05\textwidth]{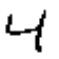}&
\includegraphics[width=0.05\textwidth]{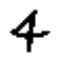}&
\includegraphics[width=0.05\textwidth]{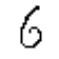}&
\includegraphics[width=0.05\textwidth]{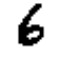}
\end{tabular}}
\caption{Sample of image pairs belonging to the same class taken from MNIST.}
\label{fig:mnist}
\end{figure}

In this section, we show the impact of our proposed stochastic graphlet embedding on the performance of handwritten digit classification. We consider the well known MNIST database\footnote{Available at \url{http://yann.lecun.com/exdb/mnist}} (see example in~\fig{fig:mnist}) which consists in $60000$ training and $10000$ test images belonging to $10$ different digit categories. In this task, the goal is to assign each test sample to one of the $10$ categories; in these experiments, we are again interested in showing significant and progressive impact -- of combining increasing order graphlets -- on performances. 
We model each binary digit with its skeleton graph; nodes in this graph correspond to pixels and edges connect these pixels to their 8 respective immediate neighbors (see~\cite{Dutta2018SGEsupp} for graph representation of digits). In order to label nodes, we consider the general shape context descriptor~\cite{Belongie2002} on nodes and cluster them using k-means algorithm (with $k=20$); the latter assigns each node a discrete label in $[1,20]$. Considering the resulting graphs (with labeled nodes) on the handwritten digits, we use our stochastic graphlet embedding in order to obtain the distributions of high-order graphlets (with $\epsilon=0.05$ and $\delta=0.05$), and we evaluate the histogram intersection {kernel~\cite{Barla2003}} (on these distributions) to achieve SVM training and classification; first, we use LIBSVM to train a ``one-vs-all'' SVM classifier for each digit category, and then we assign a given test digit to the category with the largest SVM score. \tab{tab:mnist} shows the classification accuracy obtained by our stochastic graphlet embedding, using graphlets with increasing orders; as shown in~\cite{Dutta2018SGEsupp}, we consider a kernel for each order. As already observed on the other datasets, the classification performances steadily improve as graphlet orders increase.

\begin{table}[!htbp]
\begin{center}
\caption{Accuracies (in \%) obtained by our method with a combination of different graphlet orders (values of $t$) on the MNIST dataset. {The average processing time for generating the embedding of a given graph is indicated within the parenthesis after each accuracy value.}}
\label{tab:mnist}
\resizebox{\columnwidth}{!}{
\begin{tabular}{|ccccccc|}
\hline
$t$ & $\lbrace 1,2\rbrace$ & $\lbrace 1,\ldots,3\rbrace$ & $\lbrace 1,\ldots,4\rbrace$ & $\lbrace 1,\ldots,5\rbrace$ & $\lbrace 1,\ldots,6\rbrace$ & $\lbrace 1,\ldots,7\rbrace$\\\hline
Acc. & $93.75$ ($1.37$) & $95.08$ ($1.65$) & $96.15$ ($2.45$) & $97.32$ ($3.51$) & $98.67$ ($3.95$) & $99.20$ ($6.27$)\\\hline
\end{tabular}}
\end{center}
\end{table}
 
\section{Conclusion}
\label{sec:concl}

In this paper, we introduce a novel high-order stochastic graphlet embedding for graph-based pattern recognition. Our method is based on a stochastic depth-first search strategy that samples connected and increasing orders subgraphs (a.k.a graphlets) from input graphs. By its design, this sampling is able to handle large (unlimited) order graphlets where nodes (in these graphlets) correspond to local information and edges capture interactions between these nodes. Our proposed method is also able to measure the distribution of the sampled isomorphic graphlets, effectively and efficiently, using hashing and without addressing the GI-complete graph isomorphism {\it nor} the NP-complete subgraph isomorphism; indeed, we use {\it efficient} hash functions to assign graphlets to isomorphic subsets with a very low probability of collision. Under the regime of large graphlet sampling, the proposed method produces empirical graphlet distributions that converge to the actual ones. Extensive experiments show the effectiveness and the positive impact of high-order graphlets on the performances of pattern recognition using various challenging databases.

As a future work, one may improve the estimates of graphlet distributions by designing other hash functions (while reducing further their probability of collision) and by eliminating the residual effect of colliding graphlets in these distributions. One may also extend the proposed framework to graphs with other attributes in order to further enlarge the application field of our method.

\section*{Acknowledgement}
{This work was partially supported by a grant from the research agency ANR (Agence Nationale de la Recherche) under the MLVIS project (Machine Learning for Visual Annotation in Social-media: ANR-11-BS02-0017) and the European Union\textquotesingle s Horizon 2020 research and innovation program under the Marie Sk\l{}odowska-Curie grant agreement No. 665919 (H2020-MSCA-COFUND-2014:665919:CVPR:01).}
\ifCLASSOPTIONcaptionsoff
 \newpage
\fi



{
\bibliographystyle{IEEEtranS}
\bstctlcite{IEEEexample:BSTcontrol}
\bibliography{IEEEabrv,./bibliography/bibliography.bib}
}
\begin{IEEEbiography}[{\includegraphics[width=1in,height=1.25in,clip,keepaspectratio]{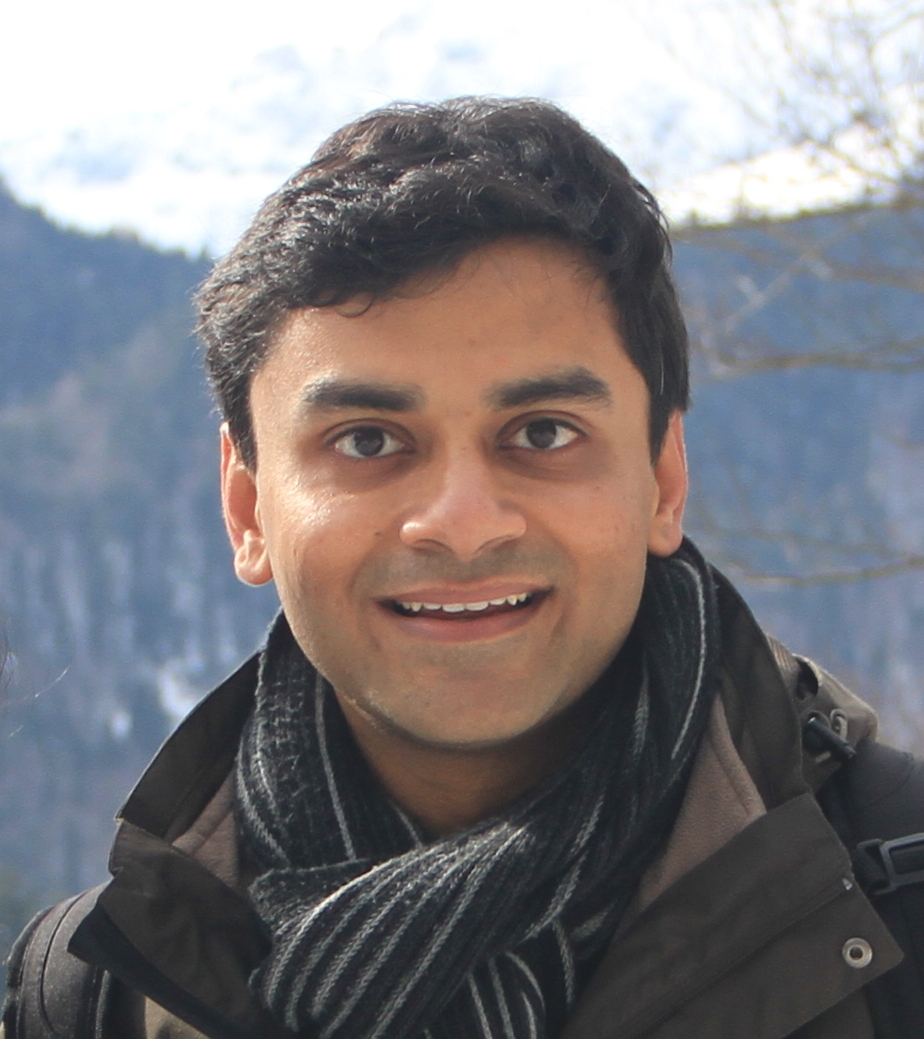}}]{Anjan Dutta}is a Marie-Curie postdoctoral fellow under the P-SPHERE project at the Computer Vision Center of Barcelona. He received PhD in computer science from the Autonomous University of Barcelona (UAB) in the year 2014. His doctoral thesis was awarded with Cum Laude qualification (highest grade). Additionally, he is a recipient of the Extraordinary PhD Thesis Award for the year 2013-14. Before his PhD, he obtained MS in computer vision and artificial intelligence also from the UAB, MCA in computer applications from the West Bengal University of Technology and BS in mathematics (honors) from the University of Calcutta respectively in the year 2010, 2009 and 2006. After completing his PhD, he worked as a postdoctoral researcher at few institutes including T\'{e}l\'{e}com ParisTech, Paris; Indian Statistical Institute, Kolkata. His recent research interests have revolved around graph-based algorithms, graph neural network, multi-modal embedding and deep learning.
\end{IEEEbiography}
\begin{IEEEbiography}[{\includegraphics[width=1in,height=1.25in,clip,keepaspectratio]{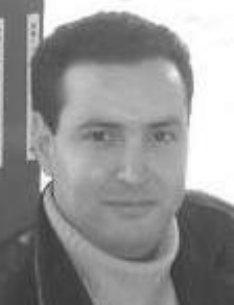}}]{Hichem Sahbi}received his MSc degree in theoretical computer science from the University of Paris Sud, Orsay, France, and his PhD in computer vision and machine learning from INRIA/Versailles University, France, in 1999 and 2003, respectively. From 2003 to 2006 he was a research associate first at the Fraunhofer Institute in Darmstadt, Germany, and then at the Machine Intelligence Laboratory at Cambridge University, UK. From 2006 to 2007, he was a senior research associate at the Ecole des Ponts ParisTech, Paris, France. Since 2007, he has been a CNRS researcher at Telecom ParisTech and then at UPMC, Sorbonne University, Paris. His research interests include statistical machine learning, kernel and graph-based inference, computer vision, and image retrieval.
\end{IEEEbiography}

\end{document}


%
\title{Supplemental Material: Stochastic Graphlet Embedding}
%
%
%
%

\author{Anjan Dutta,~\IEEEmembership{Member,~IEEE,} 
        and~Hichem Sahbi,~\IEEEmembership{Member,~IEEE,}

}

\maketitle



%
\section{Additional Experimental Results}
This document is the supplemental material of~\cite{Dutta2018SGE}. The main goal of this documentation is to provide additional experimental results or justifications related to the proposed Stochastic Graphlet Embedding (SGE) method presented in~\cite{Dutta2018SGE}.

\subsection{Finding $t$, $\epsilon$ and $\delta$ through $n$-fold cross validation}
In the original manuscript, we have shown results by varying the parameters $t\in\lbrace1,\ldots,7\rbrace$, $\epsilon\in\lbrace0.1, 0.05\rbrace$ and $\delta\in\lbrace0.1, 0.05\rbrace$. However, these parameter values can be selected based on $n$-fold cross validation performed on a validation set. \tab{tab:param-cross-valid} shows selected values of $t$, $\epsilon$ and $\delta$ for different datasets based on $10$-fold cross validation and the corresponding classification accuracy on the test sets. The validation and test sets are obtained by equally dividing the respective test sets used in the original manuscript (Table VI); note that classification accuracies (obtained by selecting the parameters through cross validation) agree with those already shown in the paper.

\begin{table}[!ht]
\begin{center}
\caption{Determination of optimal $t$, $\epsilon$ and $\delta$ for different datasets based on a $10-$fold cross validation on the validation set.}
\label{tab:param-cross-valid}
\begin{tabular}{|l|c|c|c|c|}
\hline
\multirow{2}{*}{Dataset} & \multicolumn{3}{c|}{Optimal parameters} & \multirow{2}{*}{Accuracy on test set} \\
\cline{2-4}
 & $t$ & $\epsilon$ & $\delta$ & \\
\hline
MUTAG   & $6$ & $0.05$ & $0.05$ & $89.71$ \\
PTC     & $4$ & $0.05$ & $0.05$ & $63.68$ \\
ENZYMES & $7$ & $0.05$ & $0.05$ & $40.81$ \\
D \& D  & $7$ & $0.05$ & $0.05$ & $76.15$ \\
NCI1    & $7$ & $0.05$ & $0.10$ & $82.87$ \\
NCI109  & $7$ & $0.05$ & $0.10$ & $82.41$ \\
\hline
\end{tabular}
\end{center}
\end{table}

\begin{figure}
\begin{center}
\includegraphics[width=8.5cm, height=8.5cm]{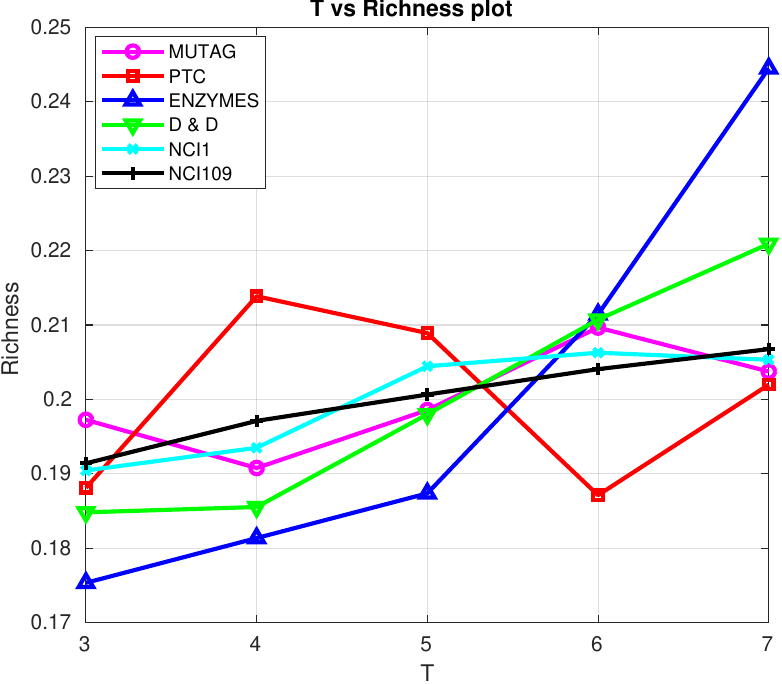}
\caption{Richness vs. T plot for MUTAG, PTC, ENZYMES, D \& D, NCI1 and NCI109 datasets.}
\label{fig:richness}
\end{center}
\end{figure}

\noindent In order to better understand the behavior of our model w.r.t $t$ and its impact on the classification accuracy (particularly on the PTC dataset), we consider the following "richness" measure for each $t$
\begin{equation*}
\text{Richness}=\frac{\sum_c ||\mu_c - \mu||_2^2}{\sum_c \sum_i \delta_{ic} ||x_i-\mu_c||_2^2/N_c}
\end{equation*}
where $\mu$ is the average of all data, $\mu_c$ is average of data belonging to class $c$, $\delta_{ic}$ is the indicator function denoting whether the $i^\text{th}$ example belongs to the class $c$, and $N_c=\sum_i\delta_{ic}$. This measure basically reflects the separability of various classes. \fig{fig:richness} shows the richness  for different $T$ (upper bound of $t$); this measure reaches its peak when $T=4$ and this is in accordance with the highest discrimination power, of our graphlet-based representation, on the PTC dataset.

\begin{table*}[!t]
\begin{center}
\caption{Classification accuracies (in \%) obtained by SGE on MUTAG, PTC, ENZYMES, D\&D, NCI1 and NCI109 datasets by uniformly sampling graphlets with larger $M$ ($5\times$) and $T$.}
\label{tab:expt-graph-class-large-M-T}
\resizebox{\textwidth}{!}{
\begin{tabular}{|l|c|c|c|c|c|c|}
\hline
Parameters & \ \ \ \ MUTAG \ \ \ \ & \ \ \ \ PTC \ \ \ \ & \ \ \ \ ENZYMES \ \ \ \ & \ \ \ \ D \& D \ \ \ \ & \ \ \ \ NCI1 \ \ \ \ & \ \ \ \ NCI109 \ \ \ \ \\
\hline
$t=3, \epsilon = 0.1, \delta = 0.1$   & $84.44 \pm 0.71$ ($0.37$) & $55.59 \pm 0.10$ ($0.39$) & $27.83 \pm 0.48$ ($0.36$) & $64.23 \pm 0.14$ ($0.35$) & $76.15 \pm 0.57$ ($0.37$) & $74.90 \pm 0.31$ ($0.38$) \\
$t=3, \epsilon = 0.1, \delta = 0.05$  & $84.21 \pm 0.47$ ($0.43$) & $56.18 \pm 0.57$ ($0.44$) & $28.34 \pm 0.67$ ($0.45$) & $63.39 \pm 0.37$ ($0.44$) & $76.15 \pm 0.84$ ($0.44$) & $74.66 \pm 0.59$ ($0.44$) \\
$t=3, \epsilon = 0.05, \delta = 0.1$  & $82.22 \pm 0.17$ ($1.05$) & $57.06 \pm 0.51$ ($1.03$) & $26.00 \pm 0.31$ ($1.04$) & $65.76 \pm 0.47$ ($1.05$) & $75.86 \pm 0.24$ ($1.03$) & $72.56 \pm 0.54$ ($1.04$) \\
$t=3, \epsilon = 0.05, \delta = 0.05$ & $85.56 \pm 0.74$ ($1.34$) & $58.40 \pm 0.67$ ($1.35$) & $29.50 \pm 0.32$ ($1.36$) & $68.90 \pm 0.99$ ($1.35$) & $77.71 \pm 0.24$ ($1.33$) & $76.21 \pm 0.82$ ($1.36$) \\
\hline
$t=4, \epsilon = 0.1, \delta = 0.1$   & $85.00 \pm 0.41$ ($1.67$) & $60.00 \pm 0.27$ ($1.65$) & $29.17 \pm 0.92$ ($1.69$) & $68.64 \pm 0.23$ ($1.71$) & $77.37 \pm 0.61$ ($1.66$) & $76.55 \pm 0.22$ ($1.66$) \\
$t=4, \epsilon = 0.1, \delta = 0.05$  & $84.44 \pm 0.41$ ($1.83$) & $61.18 \pm 0.39$ ($1.81$) & $30.50 \pm 0.26$ ($1.86$) & $68.94 \pm 0.88$ ($1.84$) & $77.37 \pm 0.65$ ($1.84$) & $76.21 \pm 0.41$ ($1.85$) \\
$t=4, \epsilon = 0.05, \delta = 0.1$  & $85.56 \pm 0.31$ ($2.15$) & $62.06 \pm 0.17$ ($2.17$) & $28.67 \pm 0.85$ ($2.18$) & $70.08 \pm 0.59$ ($2.19$) & $76.15 \pm 0.91$ ($2.16$) & $78.48 \pm 0.60$ ($2.18$) \\
$t=4, \epsilon = 0.05, \delta = 0.05$ & $86.67 \pm 0.89$ ($2.27$) & $64.12 \pm 0.23$ ($2.28$) & $31.17 \pm 0.72$ ($2.29$) & $72.63 \pm 0.24$ ($2.26$) & $78.49 \pm 0.67$ ($2.26$) & $78.48 \pm 0.80$ ($2.25$) \\
\hline
$t=5, \epsilon = 0.1, \delta = 0.1$   & $86.67 \pm 0.05$ ($2.59$) & $59.12 \pm 0.26$ ($2.58$) & $32.17 \pm 0.43$ ($2.60$) & $72.54 \pm 0.60$ ($2.61$) & $79.51 \pm 0.49$ ($2.62$) & $78.82 \pm 0.33$ ($2.61$) \\
$t=5, \epsilon = 0.1, \delta = 0.05$  & $87.78 \pm 0.05$ ($2.73$) & $61.18 \pm 0.23$ ($2.75$) & $35.17 \pm 0.46$ ($2.72$) & $72.54 \pm 0.84$ ($2.75$) & $81.70 \pm 0.67$ ($2.75$) & $78.48 \pm 0.23$ ($2.74$) \\
$t=5, \epsilon = 0.05, \delta = 0.1$  & $86.11 \pm 0.52$ ($3.13$) & $63.53 \pm 0.90$ ($3.15$) & $36.67 \pm 0.27$ ($3.14$) & $70.08 \pm 0.22$ ($3.13$) & $81.13 \pm 0.13$ ($3.11$) & $76.55 \pm 0.80$ ($3.12$) \\
$t=5, \epsilon = 0.05, \delta = 0.05$ & $87.22 \pm 0.89$ ($3.31$) & $65.00 \pm 0.79$ ($3.33$) & $37.17 \pm 0.85$ ($3.32$) & $73.05 \pm 0.81$ ($3.33$) & $81.26 \pm 0.29$ ($3.34$) & $81.22 \pm 0.85$ ($3.30$) \\
\hline
$t=6, \epsilon = 0.1, \delta = 0.1$   & $87.78 \pm 0.31$ ($3.72$) & $58.53 \pm 0.06$ ($3.73$) & $36.83 \pm 0.22$ ($3.74$) & $72.80 \pm 0.90$ ($3.71$) & $81.70 \pm 0.84$ ($3.74$) & $81.25 \pm 0.29$ ($3.73$) \\
$t=6, \epsilon = 0.1, \delta = 0.05$  & $88.33 \pm 0.15$ ($3.85$) & $59.41 \pm 0.52$ ($3.84$) & $37.33 \pm 0.66$ ($3.86$) & $73.05 \pm 0.48$ ($3.85$) & $81.84 \pm 0.94$ ($3.87$) & $81.25 \pm 0.92$ ($3.83$) \\
$t=6, \epsilon = 0.05, \delta = 0.1$  & $86.11 \pm 0.70$ ($4.22$) & $58.53 \pm 0.58$ ($4.21$) & $36.67 \pm 0.28$ ($4.23$) & $72.54 \pm 0.37$ ($4.25$) & $81.75 \pm 0.88$ ($4.24$) & $81.38 \pm 0.54$ ($4.25$) \\
$t=6, \epsilon = 0.05, \delta = 0.05$ & $88.89 \pm 0.24$ ($4.36$) & $57.65 \pm 0.96$ ($4.35$) & $40.50 \pm 0.26$ ($4.34$) & $74.56 \pm 0.64$ ($4.34$) & $82.48 \pm 0.87$ ($4.35$) & $82.32 \pm 0.56$ ($4.34$) \\
\hline
$t=7, \epsilon = 0.1, \delta = 0.1$   & $89.44 \pm 0.68$ ($4.72$) & $59.12 \pm 0.99$ ($4.71$) & $40.17 \pm 0.46$ ($4.70$) & $76.15 \pm 0.66$ ($4.73$) & $82.40 \pm 0.74$ ($4.74$) & $82.62 \pm 0.80$ ($4.73$) \\
$t=7, \epsilon = 0.1, \delta = 0.05$  & $86.11 \pm 0.93$ ($4.86$) & $60.00 \pm 0.82$ ($4.83$) & $37.33 \pm 0.85$ ($4.84$) & $76.08 \pm 0.41$ ($4.85$) & $81.75 \pm 0.55$ ($4.87$) & $82.62 \pm 0.15$ ($4.83$) \\
$t=7, \epsilon = 0.05, \delta = 0.1$  & $88.33 \pm 0.37$ ($5.95$) & $62.06 \pm 0.26$ ($5.94$) & $40.00 \pm 0.50$ ($5.96$) & $73.05 \pm 0.33$ ($5.93$) & $79.51 \pm 0.91$ ($5.94$) & $80.94 \pm 0.42$ ($5.95$) \\
$t=7, \epsilon = 0.05, \delta = 0.05$ & $89.75 \pm 0.27$ ($6.31$) & $57.65 \pm 0.99$ ($6.32$) & $40.67 \pm 0.40$ ($6.33$) & $77.43 \pm 0.27$ ($6.33$) & $82.49 \pm 0.94$ ($6.33$) & $83.12 \pm 0.65$ ($6.33$) \\
\hline
$t=8, \epsilon = 0.1, \delta = 0.1$   & $88.33 \pm 0.47$ ($6.82$) & $59.41 \pm 0.79$ ($6.83$) & $40.00 \pm 0.06$ ($6.83$) & $74.56 \pm 0.12$ ($6.82$) & $81.26 \pm 0.34$ ($6.83$) & $79.74 \pm 0.24$ ($6.85$) \\
$t=8, \epsilon = 0.1, \delta = 0.05$  & $88.89 \pm 0.15$ ($7.14$) & $61.18 \pm 0.59$ ($7.15$) & $37.33 \pm 0.73$ ($7.13$) & $74.56 \pm 0.71$ ($7.16$) & $81.70 \pm 0.54$ ($7.13$) & $80.94 \pm 0.43$ ($7.12$) \\
$t=8, \epsilon = 0.05, \delta = 0.1$  & $87.78 \pm 0.86$ ($8.47$) & $60.00 \pm 0.17$ ($8.45$) & $40.00 \pm 0.14$ ($8.48$) & $73.05 \pm 0.61$ ($8.46$) & $82.03 \pm 0.13$ ($8.45$) & $81.25 \pm 0.71$ ($8.45$) \\
$t=8, \epsilon = 0.05, \delta = 0.05$ & $89.44 \pm 0.61$ ($8.75$) & $61.47 \pm 0.96$ ($8.76$) & $40.67 \pm 0.64$ ($8.74$) & $76.08 \pm 0.74$ ($8.73$) & $82.10 \pm 0.76$ ($8.75$) & $82.62 \pm 0.78$ ($8.76$) \\
\hline
$t=9, \epsilon = 0.1, \delta = 0.1$   & $87.78 \pm 0.30$ ($8.91$) & $59.12 \pm 0.55$ ($8.93$) & $37.33 \pm 0.02$ ($8.94$) & $73.05 \pm 0.92$ ($8.90$) & $81.13 \pm 0.87$ ($8.93$) & $79.89 \pm 0.41$ ($8.92$) \\
$t=9, \epsilon = 0.1, \delta = 0.05$  & $86.67 \pm 0.30$ ($9.27$) & $61.08 \pm 0.60$ ($9.25$) & $36.67 \pm 0.80$ ($9.28$) & $73.44 \pm 0.26$ ($9.29$) & $81.70 \pm 0.45$ ($9.30$) & $79.74 \pm 0.12$ ($9.25$) \\
$t=9, \epsilon = 0.05, \delta = 0.1$  & $87.22 \pm 0.55$ ($10.54$) & $60.00 \pm 0.35$ ($10.53$) & $38.50 \pm 0.15$ ($10.55$) & $75.39 \pm 0.98$ ($10.56$) & $82.03 \pm 0.67$ ($10.57$) & $82.62 \pm 0.45$ ($10.52$) \\
$t=9, \epsilon = 0.05, \delta = 0.05$ & $88.33 \pm 0.11$ ($10.67$) & $62.06 \pm 0.05$ ($10.68$) & $40.00 \pm 0.73$ ($10.65$) & $76.08 \pm 0.31$ ($10.64$) & $82.49 \pm 0.45$ ($10.68$) & $83.12 \pm 0.18$ ($10.65$) \\
\hline
$t=10, \epsilon = 0.1, \delta = 0.1$   & $88.89 \pm 0.82$ ($11.24$) & $58.53 \pm 0.28$ ($11.25$) & $38.47 \pm 0.51$ ($11.26$) & $74.76 \pm 0.15$ ($11.26$) & $82.03 \pm 0.49$ ($11.27$) & $80.94 \pm 0.27$ ($11.25$) \\
$t=10, \epsilon = 0.1, \delta = 0.05$  & $87.78 \pm 0.37$ ($11.67$) & $59.41 \pm 0.13$ ($11.68$) & $38.50 \pm 0.32$ ($11.66$) & $76.08 \pm 0.21$ ($11.69$) & $82.03 \pm 0.65$ ($11.68$) & $81.22 \pm 0.54$ ($11.66$) \\
$t=10, \epsilon = 0.05, \delta = 0.1$  & $88.33 \pm 0.29$ ($13.57$) & $61.47 \pm 0.65$ ($13.55$) & $36.67 \pm 0.84$ ($13.58$) & $75.39 \pm 0.64$ ($13.55$) & $81.75 \pm 0.41$ ($13.56$) & $82.32 \pm 0.15$ ($13.58$) \\
$t=10, \epsilon = 0.05, \delta = 0.05$ & $89.75 \pm 0.27$ ($14.89$) & $61.18 \pm 0.32$ ($14.90$) & $40.17 \pm 0.32$ ($14.91$) & $77.43 \pm 0.45$ ($14.88$) & $82.48 \pm 0.54$ ($14.92$) & $82.62 \pm 0.54$ ($14.90$) \\
\hline
$t=11, \epsilon = 0.1, \delta = 0.1$   & $89.44 \pm 0.89$ ($15.43$) & $56.18 \pm 0.38$ ($15.44$) & $38.00 \pm 0.15$ ($15.45$) & $73.05 \pm 0.32$ ($15.42$) & $79.51 \pm 0.67$ ($15.43$) & $79.89 \pm 0.45$ ($15.42$) \\
$t=11, \epsilon = 0.1, \delta = 0.05$  & $88.89 \pm 0.44$ ($16.27$) & $58.53 \pm 0.75$ ($16.25$) & $38.50 \pm 0.88$ ($16.29$) & $72.54 \pm 0.67$ ($16.29$) & $81.26 \pm 0.62$ ($16.30$) & $80.94 \pm 0.45$ ($16.26$) \\
$t=11, \epsilon = 0.05, \delta = 0.1$  & $87.78 \pm 0.72$ ($18.12$) & $57.65 \pm 0.22$ ($18.14$) & $40.00 \pm 0.78$ ($18.10$) & $76.08 \pm 0.89$ ($18.15$) & $81.75 \pm 0.45$ ($18.11$) & $82.62 \pm 0.67$ ($18.13$) \\
$t=11, \epsilon = 0.05, \delta = 0.05$ & $89.75 \pm 0.72$ ($19.08$) & $55.59 \pm 0.92$ ($19.10$) & $40.50 \pm 0.20$ ($19.09$) & $76.58 \pm 0.89$ ($19.11$) & $82.10 \pm 0.65$ ($19.08$) & $83.12 \pm 0.38$ ($19.07$) \\
\hline
$t=12, \epsilon = 0.1, \delta = 0.1$   & $87.78 \pm 0.23$ ($20.10$) & $57.06 \pm 0.31$ ($20.09$) & $36.83 \pm 0.32$ ($20.12$) & $72.80 \pm 0.45$ ($20.12$) & $77.71 \pm 0.54$ ($20.11$) & $81.38 \pm 0.15$ ($20.11$) \\
$t=12, \epsilon = 0.1, \delta = 0.05$  & $88.33 \pm 0.78$ ($21.14$) & $57.65 \pm 0.78$ ($21.15$) & $37.33 \pm 0.21$ ($21.18$) & $72.54 \pm 0.25$ ($21.17$) & $77.71 \pm 0.21$ ($21.12$) & $81.22 \pm 0.61$ ($21.15$) \\
$t=12, \epsilon = 0.05, \delta = 0.1$  & $88.33 \pm 0.65$ ($22.50$) & $58.53 \pm 0.32$ ($22.51$) & $38.50 \pm 0.51$ ($22.53$) & $73.05 \pm 0.54$ ($22.53$) & $79.51 \pm 0.25$ ($22.52$) & $82.32 \pm 0.61$ ($22.51$) \\
$t=12, \epsilon = 0.05, \delta = 0.05$ & $89.44 \pm 0.52$ ($23.42$) & $61.47 \pm 0.21$ ($23.39$) & $40.17 \pm 0.54$ ($23.40$) & $76.08 \pm 0.21$ ($23.43$) & $81.13 \pm 0.63$ ($23.42$) & $82.62 \pm 0.51$ ($23.44$) \\
\hline
$t=13, \epsilon = 0.1, \delta = 0.1$   & $88.33 \pm 0.58$ ($24.28$) & $59.41 \pm 0.03$ ($24.29$) & $38.47 \pm 0.21$ ($24.30$) & $74.68 \pm 0.21$ ($24.31$) & $81.13 \pm 0.34$ ($24.26$) & $80.65 \pm 0.34$ ($24.27$) \\
$t=13, \epsilon = 0.1, \delta = 0.05$  & $87.78 \pm 0.21$ ($25.58$) & $60.29 \pm 0.21$ ($25.56$) & $37.33 \pm 0.20$ ($25.55$) & $75.39 \pm 0.73$ ($25.56$) & $81.70 \pm 0.76$ ($25.59$) & $81.25 \pm 0.29$ ($25.61$) \\
$t=13, \epsilon = 0.05, \delta = 0.1$  & $86.67 \pm 0.51$ ($27.25$) & $61.18 \pm 0.85$ ($27.27$) & $36.67 \pm 0.54$ ($27.28$) & $76.08 \pm 0.35$ ($27.26$) & $79.51 \pm 0.39$ ($27.26$) & $80.94 \pm 0.43$ ($27.27$) \\
$t=13, \epsilon = 0.05, \delta = 0.05$ & $88.89 \pm 0.54$ ($28.42$) & $61.47 \pm 0.74$ ($28.39$) & $40.67 \pm 0.78$ ($28.43$) & $76.58 \pm 0.46$ ($28.44$) & $82.49 \pm 0.27$ ($28.45$) & $82.62 \pm 0.67$ ($28.44$) \\
\hline
$t=14, \epsilon = 0.1, \delta = 0.1$   & $88.89 \pm 0.49$ ($29.74$) & $58.53 \pm 0.37$ ($29.76$) & $36.33 \pm 0.64$ ($29.75$) & $72.63 \pm 0.37$ ($29.73$) & $79.51 \pm 0.49$ ($29.75$) & $81.22 \pm 0.64$ ($29.76$) \\
$t=14, \epsilon = 0.1, \delta = 0.05$  & $88.33 \pm 0.37$ ($30.10$) & $59.41 \pm 0.46$ ($30.12$) & $37.33 \pm 0.79$ ($30.11$) & $73.05 \pm 0.67$ ($30.12$) & $81.70 \pm 0.39$ ($30.11$) & $81.25 \pm 0.37$ ($30.12$) \\
$t=14, \epsilon = 0.05, \delta = 0.1$  & $87.78 \pm 0.49$ ($31.23$) & $60.00 \pm 0.76$ ($31.22$) & $38.50 \pm 0.34$ ($31.21$) & $75.39 \pm 0.39$ ($31.24$) & $82.03 \pm 0.73$ ($31.20$) & $82.62 \pm 0.67$ ($31.21$) \\
$t=14, \epsilon = 0.05, \delta = 0.05$ & $89.44 \pm 0.56$ ($33.41$) & $62.06 \pm 0.48$ ($33.43$) & $40.00 \pm 0.67$ ($33.42$) & $74.68 \pm 0.46$ ($33.40$) & $82.10 \pm 0.16$ ($33.44$) & $83.12 \pm 0.13$ ($33.43$) \\
\hline
$t=15, \epsilon = 0.1, \delta = 0.1$   & $88.89 \pm 0.34$ ($35.21$) & $59.41 \pm 0.43$ ($35.22$) & $37.33 \pm 0.34$ ($35.25$) & $73.80 \pm 0.46$ ($35.24$) & $81.13 \pm 0.46$ ($35.25$) & $78.48 \pm 0.37$ ($35.20$) \\
$t=15, \epsilon = 0.1, \delta = 0.05$  & $88.33 \pm 0.46$ ($36.78$) & $61.18 \pm 0.67$ ($36.80$) & $38.67 \pm 0.64$ ($36.77$) & $73.05 \pm 0.34$ ($36.76$) & $82.40 \pm 0.63$ ($36.80$) & $79.74 \pm 0.64$ ($36.79$) \\
$t=15, \epsilon = 0.05, \delta = 0.1$  & $89.75 \pm 0.37$ ($38.91$) & $60.29 \pm 0.27$ ($38.93$) & $38.47 \pm 0.49$ ($38.95$) & $72.63 \pm 0.38$ ($38.93$) & $79.51 \pm 0.28$ ($38.94$) & $80.94 \pm 0.47$ ($38.95$) \\
$t=15, \epsilon = 0.05, \delta = 0.05$ & $87.78 \pm 0.64$ ($41.38$) & $61.47 \pm 0.34$ ($41.40$) & $40.67 \pm 0.37$ ($41.36$) & $75.68 \pm 0.46$ ($41.39$) & $81.26 \pm 0.29$ ($41.42$) & $81.22 \pm 0.39$ ($41.39$) \\
\hline
\end{tabular}}
\end{center}
\end{table*}

\begin{table*}[!t]
\begin{center}
\caption{Classification accuracies (in \%) obtained by SGE on MUTAG, PTC, ENZYMES, D\&D, NCI1 and NCI109 datasets, where the first node of each graphlet is sampled following different strategies: \emph{highest betweenness centrality}, \emph{highest degree} and \emph{random seeds}.}
\label{tab:expt-diff-sample-strategy}
\resizebox{\textwidth}{!}{
\begin{tabular}{|l|l|c|c|c|c|c|c|}
\hline
Sampling Strategy & Parameters & \ \ \ \ MUTAG \ \ \ \ & \ \ \ \ PTC \ \ \ \ & \ \ \ \ ENZYMES \ \ \ \ & \ \ \ \ D \& D \ \ \ \ & \ \ \ \ NCI1 \ \ \ \ & \ \ \ \ NCI109 \ \ \ \ \\
\hline
\multirow{8}{*}{Highest betweenness centrality} & $t=3$, $\epsilon=0.1$, $\delta=0.1$ & $70.00 \pm 0.89$ & $51.45 \pm 0.57$ & $22.57 \pm 0.45$ & $58.32 \pm 0.52$ & $71.67 \pm 0.52$ & $70.31 \pm 0.52$ \\
 & $t=3$, $\epsilon=0.1$, $\delta=0.05$ & $71.67 \pm 0.48$ & $52.12 \pm 0.75$ & $23.14 \pm 0.52$ & $59.53 \pm 0.42$ & $73.12 \pm 0.74$ & $71.74 \pm 0.52$ \\
 & $t=3$, $\epsilon=0.05$, $\delta=0.1$ & $73.33 \pm 0.86$ & $53.74 \pm 0.54$ & $24.74 \pm 0.63$ & $61.47 \pm 0.74$ & $74.32 \pm 0.71$ & $72.85 \pm 0.37$ \\
 & $t=3$, $\epsilon=0.05$, $\delta=0.05$ & $74.65 \pm 0.47$ & $54.12 \pm 0.87$ & $25.78 \pm 0.35$ & $62.75 \pm 0.54$ & $75.65 \pm 0.42$ & $74.15 \pm 0.43$ \\
\cline{2-8}
 & $t=7$, $\epsilon=0.1$, $\delta=0.1$ & $76.11 \pm 0.57$ & $53.41 \pm 0.25$ & $24.89 \pm 0.57$ & $62.87 \pm 0.42$ & $74.52 \pm 0.41$ & $73.52 \pm 0.34$ \\
 & $t=7$, $\epsilon=0.1$, $\delta=0.05$ & $75.56 \pm 0.38$ & $54.24 \pm 0.41$ & $25.65 \pm 0.78$ & $63.75 \pm 0.39$ & $75.34 \pm 0.53$ & $74.89 \pm 0.42$ \\
 & $t=7$, $\epsilon=0.05$, $\delta=0.1$ & $76.67 \pm 0.39$ & $55.42 \pm 0.74$ & $26.89 \pm 0.45$ & $65.24 \pm 0.51$ & $76.43 \pm 0.42$ & $73.58 \pm 0.45$ \\
 & $t=7$, $\epsilon=0.05$, $\delta=0.05$ & $77.65 \pm 0.74$ & $56.24 \pm 0.56$ & $28.12 \pm 0.74$ & $68.34 \pm 0.24$ & $78.45 \pm 0.36$ & $75.42 \pm 0.54$ \\
\hline
\multirow{8}{*}{Highest degree} & $t=3$, $\epsilon=0.1$, $\delta=0.1$ & $73.89 \pm 0.34$ & $53.82 \pm 0.42$ & $24.89 \pm 0.74$ & $63.47 \pm 0.35$ & $73.24 \pm 0.52$ & $71.25 \pm 0.42$ \\
 & $t=3$, $\epsilon=0.1$, $\delta=0.05$ & $75.00 \pm 0.71$ & $55.12 \pm 0.57$ & $26.12 \pm 0.45$ & $64.76 \pm 0.42$ & $74.35 \pm 0.43$ & $72.89 \pm 0.25$ \\
 & $t=3$, $\epsilon=0.05$, $\delta=0.1$ & $77.78 \pm 0.89$ & $55.89 \pm 0.59$ & $27.34 \pm 0.68$ & $66.34 \pm 0.53$ & $74.76 \pm 0.52$ & $73.72 \pm 0.56$ \\
 & $t=3$, $\epsilon=0.05$, $\delta=0.05$ & $78.24 \pm 0.74$ & $56.78 \pm 0.75$ & $28.74 \pm 0.42$ & $67.24 \pm 0.46$ & $76.24 \pm 0.47$ & $73.87 \pm 0.74$ \\
\cline{2-8}
 & $t=7$, $\epsilon=0.1$, $\delta=0.1$ & $78.89 \pm 0.84$ & $56.49 \pm 0.58$ & $28.87 \pm 0.25$ & $66.54 \pm 0.74$ & $75.32 \pm 0.34$ & $73.62 \pm 0.24$ \\
 & $t=7$, $\epsilon=0.1$, $\delta=0.05$ & $81.67 \pm 0.44$ & $57.32 \pm 0.74$ & $29.45 \pm 0.74$ & $67.18 \pm 0.36$ & $77.54 \pm 0.42$ & $74.42 \pm 0.46$ \\
 & $t=7$, $\epsilon=0.05$, $\delta=0.1$ & $82.22 \pm 0.72$ & $58.19 \pm 0.78$ & $30.45 \pm 0.42$ & $68.87 \pm 0.61$ & $75.76 \pm 0.82$ & $75.32 \pm 0.52$ \\
 & $t=7$, $\epsilon=0.05$, $\delta=0.05$ & $83.74 \pm 0.47$ & $59.76 \pm 0.49$ & $32.42 \pm 0.56$ & $70.24 \pm 0.72$ & $78.52 \pm 0.61$ & $76.56 \pm 0.48$ \\
\hline
\multirow{8}{*}{Random seeds} & $t=3$, $\epsilon=0.1$, $\delta=0.1$   & $71.54 \pm 0.28$ & $52.58 \pm 0.52$ & $24.45 \pm 0.42$ & $60.16 \pm 0.78$ & $72.23 \pm 0.42$ & $70.83 \pm 0.42$ \\
 & $t=3$, $\epsilon=0.1$, $\delta=0.05$  & $75.24 \pm 0.78$ & $54.42 \pm 0.46$ & $25.78 \pm 0.78$ & $61.12 \pm 0.26$ & $73.82 \pm 0.26$ & $72.34 \pm 0.42$ \\
 & $t=3$, $\epsilon=0.05$, $\delta=0.1$  & $86.08 \pm 0.57$ & $55.16 \pm 0.62$ & $27.23 \pm 0.13$ & $63.56 \pm 0.42$ & $75.84 \pm 0.42$ & $74.12 \pm 0.73$ \\
 & $t=3$, $\epsilon=0.05$, $\delta=0.05$ & $85.29 \pm 0.84$ & $56.42 \pm 0.38$ & $28.87 \pm 0.42$ & $64.32 \pm 0.47$ & $76.84 \pm 0.62$ & $75.42 \pm 0.72$ \\
\cline{2-8}
 & $t=7$, $\epsilon=0.1$, $\delta=0.1$   & $85.47 \pm 0.47$ & $56.34 \pm 0.54$ & $34.64 \pm 0.14$ & $71.32 \pm 0.42$ & $80.42 \pm 0.27$ & $80.48 \pm 0.83$ \\
 & $t=7$, $\epsilon=0.1$, $\delta=0.05$  & $84.31 \pm 0.15$ & $58.45 \pm 0.42$ & $35.24 \pm 0.46$ & $72.85 \pm 0.27$ & $81.24 \pm 0.42$ & $81.54 \pm 0.42$ \\
 & $t=7$, $\epsilon=0.05$, $\delta=0.1$  & $86.21 \pm 0.27$ & $60.42 \pm 0.75$ & $36.64 \pm 0.17$ & $75.42 \pm 0.42$ & $82.48 \pm 0.47$ & $82.83 \pm 0.72$ \\
 & $t=7$, $\epsilon=0.05$, $\delta=0.05$ & $87.73 \pm 0.78$ & $62.76 \pm 0.24$ & $38.45 \pm 0.32$ & $76.08 \pm 0.42$ & $82.49 \pm 0.94$ & $83.21 \pm 0.16$ \\
\hline
\end{tabular}
}
\end{center}
\end{table*}

\subsection{Experiments with large $M$ and $T$}

In the original manuscript, we experimented different values of $T\in\lbrace 1,\ldots, 7\rbrace$ and $M$ (according to \textbf{Theorem 1}). Besides, we have also performed other experiments with larger values of $T$ and $M$; in these new experiments, $T \in \{1,\dots,15\}$ and $M$ is set to values $5$ times bigger than the ones suggested by \textbf{Theorem 1}. \tab{tab:expt-graph-class-large-M-T} shows the classification accuracies on MUTAG, PTC, ENZYMES, D \& D, NCI1, and NCI109 datasets. From these results, it can be seen that large values of $M$ make it possible to achieve better accuracies with smaller graphlet structures. For instance, in case of MUTAG dataset, sampling $M$ graphlets (following \textbf{Theorem 1} with $t=3, \epsilon=0.1, \delta=0.1$) makes it possible to achieve a classification accuracy of $71.67 \pm 0.86$ (see Table VI in the manuscript), whereas increasing $M$ five times makes performance reaching an upper limit of $84.44 \pm 0.71$ (see \tab{tab:expt-graph-class-large-M-T}). A similar behavior is observed on the other datasets as well; nevertheless, this gain is obtained to the detriment of an increase in the computational efficiency.

\begin{table*}[!t]
\begin{center}
\caption{Classification accuracies (in \%) obtained by SGE on MUTAG, PTC, ENZYMES, D\&D, NCI1 and NCI109 datasets by considering the combination of histograms of graphlets with different number of edges.}
\label{tab:expt-graph-class1}
\resizebox{\textwidth}{!}{
\begin{tabular}{|l|c|c|c|c|c|c|}
\hline
Kernel & \ \ \ \ MUTAG \ \ \ \ & \ \ \ \ PTC \ \ \ \ & \ \ \ \ ENZYMES \ \ \ \ & \ \ \ \ D \& D \ \ \ \ & \ \ \ \ NCI1 \ \ \ \ & \ \ \ \ NCI109 \ \ \ \ \\
\hline
SGE ($t=\lbrace 3,\dots,4 \rbrace, \epsilon = 0.1, \delta = 0.1$) & $77.22\pm 0.47$ & $58.53\pm 0.79$ & $28.50\pm 0.06$ & $63.42\pm 0.12$ & $77.45\pm 0.34$ & $78.25\pm 0.24$ \\
SGE ($t=\lbrace 3,\dots,4 \rbrace, \epsilon = 0.1, \delta = 0.05$) & $83.89\pm 0.15$ & $63.24\pm 0.59$ & $27.33\pm 0.73$ & $64.27\pm 0.71$ & $77.65\pm 0.54$ & $78.67\pm 0.43$ \\
SGE ($t=\lbrace 3,\dots,4 \rbrace, \epsilon = 0.05, \delta = 0.1$) & $88.89\pm 0.86$ & $58.24\pm 0.17$ & $31.50\pm 0.14$ & $66.58\pm 0.61$ & $78.76\pm 0.13$ & $79.34\pm 0.71$ \\
SGE ($t=\lbrace 3,\dots,4 \rbrace, \epsilon = 0.05, \delta = 0.05$) & $87.78\pm 0.61$ & $60.00\pm 0.96$ & $33.17\pm 0.64$ & $66.66\pm 0.74$ & $78.87\pm 0.76$ & $79.98\pm 0.78$ \\
\hline
SGE ($t=\lbrace 3,\dots,5 \rbrace, \epsilon = 0.1, \delta = 0.1$) & $85.00\pm 0.30$ & $57.94\pm 0.55$ & $30.67\pm 0.02$ & $66.75\pm 0.92$ & $78.25\pm 0.87$ & $79.56\pm 0.41$ \\
SGE ($t=\lbrace 3,\dots,5 \rbrace, \epsilon = 0.1, \delta = 0.05$) & $87.22\pm 0.30$ & $57.94\pm 0.60$ & $31.50\pm 0.80$ & $67.66\pm 0.26$ & $78.92\pm 0.45$ & $79.78\pm 0.12$ \\
SGE ($t=\lbrace 3,\dots,5 \rbrace, \epsilon = 0.05, \delta = 0.1$) & $86.11\pm 0.55$ & $55.88\pm 0.35$ & $34.50\pm 0.15$ & $68.63\pm 0.98$ & $79.81\pm 0.67$ & $80.23\pm 0.45$ \\
SGE ($t=\lbrace 3,\dots,5 \rbrace, \epsilon = 0.05, \delta = 0.05$) & $84.44\pm 0.11$ & $56.76\pm 0.05$ & $34.50\pm 0.73$ & $69.56\pm 0.31$ & $80.12\pm 0.45$ & $79.93\pm 0.18$ \\
\hline
SGE ($t=\lbrace 3,\dots,6 \rbrace, \epsilon = 0.1, \delta = 0.1$) & $87.22\pm 0.89$ & $58.82\pm 0.38$ & $31.17\pm 0.15$ & $70.15\pm 0.32$ & $79.48\pm 0.67$ & $80.12\pm 0.45$ \\
SGE ($t=\lbrace 3,\dots,6 \rbrace, \epsilon = 0.1, \delta = 0.05$) & $87.22\pm 0.44$ & $57.35\pm 0.75$ & $33.17\pm 0.88$ & $70.81\pm 0.67$ & $80.12\pm 0.62$ & $80.78\pm 0.45$ \\
SGE ($t=\lbrace 3,\dots,6 \rbrace, \epsilon = 0.05, \delta = 0.1$) & $86.11\pm 0.72$ & $57.35\pm 0.22$ & $35.50\pm 0.78$ & $72.27\pm 0.89$ & $81.14\pm 0.45$ & $81.76\pm 0.67$ \\
SGE ($t=\lbrace 3,\dots,6 \rbrace, \epsilon = 0.05, \delta = 0.05$) & $86.11\pm 0.72$ & $56.18\pm 0.92$ & $35.33\pm 0.20$ & $73.10\pm 0.89$ & $81.98\pm 0.65$ & $81.76\pm 0.38$ \\
\hline
SGE ($t=\lbrace 3,\dots,7 \rbrace, \epsilon = 0.1, \delta = 0.1$) & $88.33\pm 0.15$ & $58.24\pm 0.15$ & $36.67\pm 0.57$ & $73.44\pm 0.12$ & $81.89\pm 0.70$ & $81.38\pm 0.14$ \\
SGE ($t=\lbrace 3,\dots,7 \rbrace, \epsilon = 0.1, \delta = 0.05$) & $87.78\pm 0.51$ & $57.65\pm 0.91$ & $38.50\pm 0.33$ & $74.59\pm 0.46$ & $82.10\pm 0.70$ & $82.67\pm 0.19$ \\
SGE ($t=\lbrace 3,\dots,7 \rbrace, \epsilon = 0.05, \delta = 0.1$) & $87.78\pm 0.11$ & $59.71\pm 0.36$ & $40.17\pm 0.11$ & $75.39\pm 0.94$ & $82.49\pm 0.96$ & $82.92\pm 0.42$ \\
SGE ($t=\lbrace 3,\dots,7 \rbrace, \epsilon = 0.05, \delta = 0.05$) & $86.67\pm 0.68$ & $59.41\pm 0.17$ & $39.67\pm 0.75$ & $\mathbf{77.31\pm 0.66}$ & $\mathbf{82.61\pm 0.40}$ & $\mathbf{83.12\pm 0.57}$ \\
\hline
\end{tabular}}
\end{center}
\end{table*}

\begin{table*}[!t]
\begin{center}
\caption{Classification accuracies (in \%) obtained by SGE on COIL, GREC, AIDS, MAO and ENZYMES (labeled) datasets by considering the combination of histograms of graphlets with different number of edges.}
\label{tab:expt-graph-class2}
\resizebox{\textwidth}{!}{
\begin{tabular}{|l|c|c|c|c|c|}
\hline
Method & \ \ \ \ COIL \ \ \ \ & \ \ \ \ GREC \ \ \ \ & \ \ \ \ AIDS \ \ \ \ & \ \ \ \ MAO \ \ \ \ & \ \ \ \ ENZYMES (labeled) \ \ \ \ \\
\hline
SGE ($t=\lbrace1,\ldots,2\rbrace, \epsilon = 0.1, \delta = 0.1$) & $92.10$ & $93.78$ & $95.09$ & $86.76$ & $33.43\pm 0.52$ \\
SGE ($t=\lbrace1,\ldots,2\rbrace, \epsilon = 0.1, \delta = 0.05$) & $93.20$ & $93.25$ & $94.87$ & $86.76$ & $34.54\pm 0.34$ \\
SGE ($t=\lbrace1,\ldots,2\rbrace, \epsilon = 0.05, \delta = 0.1$) & $94.30$ & $94.25$ & $95.14$ & $88.23$ & $35.22\pm 0.97$ \\
SGE ($t=\lbrace1,\ldots,2\rbrace, \epsilon = 0.05, \delta = 0.05$) & $95.10$ & $96.17$ & $95.56$ & $88.23$ & $36.24\pm 0.89$ \\
\hline
SGE ($t=\lbrace1,\ldots,3\rbrace, \epsilon = 0.1, \delta = 0.1$) & $92.40$ & $95.13$ & $95.54$ & $89.71$ & $39.84\pm 0.89$ \\
SGE ($t=\lbrace1,\ldots,3\rbrace, \epsilon = 0.1, \delta = 0.05$) & $94.10$ & $96.47$ & $96.14$ & $89.71$ & $40.56\pm 0.56$ \\
SGE ($t=\lbrace1,\ldots,3\rbrace, \epsilon = 0.05, \delta = 0.1$) & $94.70$ & $96.98$ & $97.34$ & $91.18$ & $44.32\pm 0.26$ \\
SGE ($t=\lbrace1,\ldots,3\rbrace, \epsilon = 0.05, \delta = 0.05$) & $95.20$ & $97.37$ & $97.12$ & $92.65$ & $45.53\pm 0.14$ \\
\hline
SGE ($t=\lbrace1,\ldots,4\rbrace, \epsilon = 0.1, \delta = 0.1$) & $94.10$ & $97.29$ & $97.12$ & $91.18$ & $45.32\pm 0.67$ \\
SGE ($t=\lbrace1,\ldots,4\rbrace, \epsilon = 0.1, \delta = 0.05$) & $95.20$ & $97.77$ & $97.59$ & $91.18$ & $46.12\pm 0.52$ \\
SGE ($t=\lbrace1,\ldots,4\rbrace, \epsilon = 0.05, \delta = 0.1$) & $96.10$ & $98.13$ & $97.37$ & $92.65$ & $50.15\pm 0.25$ \\
SGE ($t=\lbrace1,\ldots,4\rbrace, \epsilon = 0.05, \delta = 0.05$) & $97.20$ & $99.07$ & $97.76$ & $94.12$ & $50.98\pm 0.81$ \\
\hline
SGE ($t=\lbrace1,\ldots,5\rbrace, \epsilon = 0.1, \delta = 0.1$) & $95.30$ & $97.89$ & $98.17$ & $94.12$ & $50.67\pm 0.73$ \\
SGE ($t=\lbrace1,\ldots,5\rbrace, \epsilon = 0.1, \delta = 0.05$) & $96.20$ & $98.17$ & $98.54$ & $94.12$ & $51.67\pm 0.62$ \\
SGE ($t=\lbrace1,\ldots,5\rbrace, \epsilon = 0.05, \delta = 0.1$) & $98.20$ & $98.89$ & $98.65$ & $97.06$ & $57.67\pm 0.70$ \\
SGE ($t=\lbrace1,\ldots,5\rbrace, \epsilon = 0.05, \delta = 0.05$) & $\mathbf{98.90}$ & $99.43$ & $98.76$ & $\mathbf{98.53}$ & $58.33\pm 0.31$ \\
\hline
\end{tabular}}
\end{center}
\end{table*}

\subsection{Experiments with different strategies for sampling the initial node}

In these experiments, we consider different strategies in order to sample the initial node. In the original manuscript, the initial node was sampled uniformly from all the nodes of a given input graph. However, there are many possible ways of sampling the initial node; we have come out with three different strategies: (1) highest betweenness centrality, (2) highest node degree, and (3) random seed, where strategies ``highest betweenness centrality'' and ``highest node degree'' refer to the way of selecting the initial node according to the highest betweenness centrality and node degree respectively, and ``random seed'' stands for randomly initializing the initial node. We have used these three strategies to select the first node of each graphlet sampling. \tab{tab:expt-diff-sample-strategy} shows the results of classifying graphs, where the graph embedding is obtained by our proposed SGE and the sampling of the first node is done with the three strategies mentioned above. Among these strategies, the ones considering the highest degree and the highest betweenness centrality (for selecting the initial node) have performed worse than the original algorithm. 
Indeed, seeding nodes with these two strategies makes it possible to explore only a small subset of graphlets from the input graphs, resulting into biased distributions in the estimated graphlet histograms. In contrast, random node seeding allows us to explore/cover a larger subset of graphlets in the input graphs and provides us with a better estimate of the underlying graphlet distributions, as also corroborated through experiments (see again \tab{tab:expt-diff-sample-strategy}).

\subsection{Experiments combining graphlet histograms of different $T$}

We have done another set of experiments by combining the histograms of graphlets with different $T$ (upper bound of $t$). \tab{tab:expt-graph-class1} contains the results obtained by SGE on the MUTAG, PTC, ENZYMES, D\&D, NCI1 and NCI109 datasets, where we have combined the histograms of graphlets with different number of edges. We observe that on some of the datasets, histograms of fixed graphlet orders achieve the highest classification  accuracy (Table VI of the main manuscript), whereas, on  some other datasets, histograms of combined graphlet orders achieve the best performance. \tab{tab:expt-graph-class2} contains the results obtained by combining the histograms of graphlets with different number of edges on the COIL, GREC, AIDS, MAO and ENZYMES (labeled) datasets. In these datasets as well, we have observed the same phenomena as the previous ones.

\begin{figure}[!ht]
\begin{center}
\includegraphics[width=0.9\columnwidth]{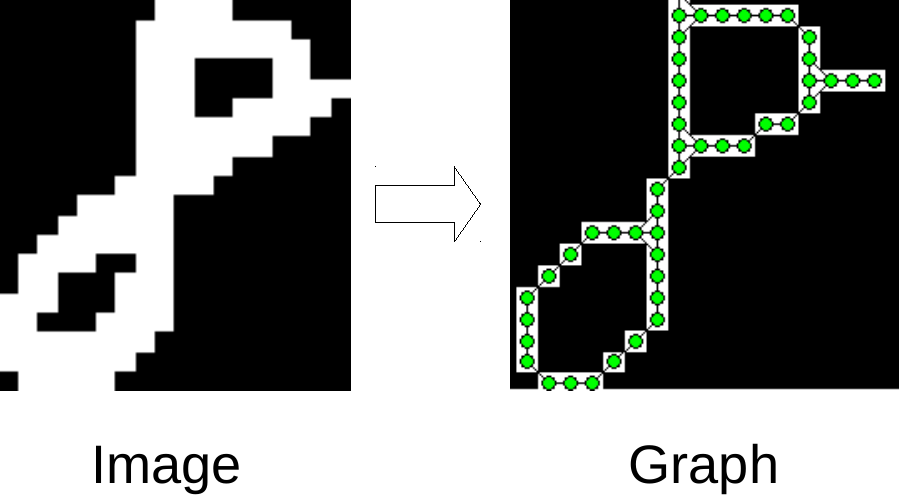}
\caption{Graph representation of digits from MNIST dataset.}
\label{fig:mnist-graph}
\end{center}
\end{figure}

\subsection{Graph representation of digits from MNIST dataset}
For obtaining the graph representation of the digits from MNIST dataset, we skeletonize the binary digit images and consider each pixel on the skeleton as a node, where all the adjacent (considering $8$ neighbours) pixels on the skeleton are connected and each connection constitutes an edge (see~\fig{fig:mnist-graph}). For labeling the graph nodes, we consider the general rotation variant shape context descriptors~\cite{Belongie2002} on each of the nodes and cluster them using k-means with $k=20$, which assign each node a label in $[1,20]$.






\ifCLASSOPTIONcaptionsoff
  \newpage
\fi



{
\bibliographystyle{IEEEtranS}
\bstctlcite{IEEEexample:BSTcontrol}
\bibliography{IEEEabrv,./bibliography/bibliography.bib}
}






